\pdfoutput=1
\documentclass[nohyperref]{article}
\usepackage[numbers]{natbib}
\usepackage{microtype}
\usepackage{graphicx}
\usepackage{subfigure}
\usepackage{booktabs} 
\usepackage{tabularray}
\usepackage{bbm}

\newcommand{\norm}[1]{\left\lVert#1\right\rVert}
\usepackage{hyperref}

\usepackage[accepted]{mystyle}

\usepackage{amsmath}
\usepackage{amssymb}
\usepackage{mathtools}
\usepackage{amsthm}
\allowdisplaybreaks
\usepackage[capitalize,noabbrev]{cleveref}

\theoremstyle{plain}
\newtheorem{theorem}{Theorem}[section]
\newtheorem{proposition}[theorem]{Proposition}
\newtheorem{lemma}[theorem]{Lemma}
\newtheorem{corollary}[theorem]{Corollary}
\theoremstyle{definition}

\theoremstyle{remark}

\usepackage[textsize=tiny]{todonotes}

\icmltitlerunning{Revisiting the Noise Model of Stochastic Gradient Descent}

\begin{document}

\twocolumn[
\icmltitle{Revisiting the Noise Model of Stochastic Gradient Descent}



\icmlsetsymbol{equal}{*}

\begin{icmlauthorlist}
\icmlauthor{Barak Battash}{yyy}
\icmlauthor{Ofir Lindenbaum}{yyy}\\
\normalsize{$^{1}$Faculty of Engineering, Bar Ilan University}
\end{icmlauthorlist}

\icmlkeywords{Machine Learning, ICML}

\vskip 0.3in
]



\printAffiliationsAndNotice{\icmlEqualContribution} 

\begin{abstract}
The stochastic gradient noise (SGN) is a significant factor in the success of stochastic gradient descent (SGD). 
Following the central limit theorem, SGN was initially modeled as Gaussian, and lately, it has been suggested that stochastic gradient noise is better characterized using $S\alpha S$ Lévy distribution.
This claim was allegedly refuted and rebounded to the previously suggested Gaussian noise model.
            This paper presents solid, detailed empirical evidence that SGN is heavy-tailed and better depicted by the $S\alpha S$ distribution. Furthermore, we argue that different parameters in a deep neural network (DNN) hold distinct SGN characteristics throughout training. To more accurately approximate the dynamics of SGD near a local minimum, we construct a novel framework in $\mathbb{R}^N$, based on Lévy-driven stochastic differential equation (SDE), where one-dimensional Lévy processes model each parameter in the DNN. Next, we show that SGN jump intensity (frequency and amplitude) depends on the learning rate decay mechanism (LRdecay); furthermore, we demonstrate empirically that the LRdecay effect may stem from the reduction of the SGN and not the decrease in the step size. Based on our analysis, we examine the mean escape time, trapping probability, and more properties of DNNs near local minima. Finally, we prove that the training process will likely exit from the basin in the direction of parameters with heavier tail SGN. We will share our code for reproducibility. 
\end{abstract}

\section{Introduction}
\label{sec:introduction}
 The tremendous success of deep learning \citep{bengio2009learning,hinton2012deep,lecun2015deep} can be partly attributed to implicit properties of the optimization tools, in particular, the popular SGD \citep{robbins1951stochastic,bottou1991stochastic} scheme.
Despite its simplicity, i.e., being a noisy first-order optimization method, SGD empirically outperforms gradient descent (GD) and second-order methods.
The stochastic gradient noise of SGD can improve the model's generalization by escaping from sharp basins and settling in wide minima \citep{ziyin2021minibatch,smith2020generalization}.
SGD noise stems from
the stochasticity in the mini-batch sampling operation, whose formation and amplitude are affected
by the DNN architecture and data distribution.
The main hurdle in improving deep learning stems from the lack of theory behind specific processes and modules frequently used; better understanding will help break current barriers in the field. Hence, studying the properties of SGD should be of the highest priority.

Analyzing the behavior of SGD optimization for non-convex cost functions is ongoing research \citep{chaudhari2018stochastic,zhou2019toward,draxler2018essentially,nguyen2017loss,he2019asymmetric,li2017visualizing,smith2021origin,ziyin2021minibatch,2019arXiv190801878Y}.  
 The problem of analyzing SGD noise has recently received much attention. Studies mainly examine the distribution and nature of the noise, with its ability to escape local minima and generalize better \citep{2017arXiv170507562H,he2019control,wu2019multiplicative,haochen2020shape,zhou2019toward,keskar2016large}.
 
SGD is based on an iterative update rule; where the $k$-th step of the iterative update rule is formulated as:
\begin{align}
\MoveEqLeft[7]
\label{eq:sgd_update_rule_cont}
    w_{k} = w_{k-1}-\frac{\eta}{B}\sum_{\ell\in \Omega_t}\nabla U^{(\ell)}(w_{k-1}) \\{}&\nonumber= 
    w_{k-1}-\eta\nabla U(w_{k-1})+\epsilon \zeta_k,
\end{align}
where $w$ denotes the weights (parameters) of the DNN, $\nabla U(w)$ is the gradient of the objective function, $B$ is the batch size, $\Omega_k \subset \{1,..,D\}$, is the randomly selected mini-batch. Thus $|\Omega_k|=B$, $D   $ is the number of data points in the dataset, $\zeta_k$ is the SGD noise, which is formulated as $\zeta_k = \nabla U(w_k) -\frac{1}{B}\sum_{\ell\in \Omega_k}\nabla U^{(\ell)}(w_k)$, i.e., the difference between the gradient produced by GD and SGD, finally $\epsilon=\eta^{\frac{\alpha-1}{\alpha}}$, and $\eta$ is the learning rate.

While gradient flow is a popular apparatus for understanding GD dynamics, continuous-time SDE is typically used to investigate the SGD optimization process. By modeling SGD using an SDE, we can examine the evolution of the dynamic system in the continuous time domain~
\citep{2018arXiv180300195Z,2020arXiv200613719M,xie2020diffusion,chaudhari2018stochastic,2017arXiv170507562H,pmlr-v32-satoa14}.
\newline
The results of empiric experimentation have produced a lively discussion on how SGN distributes. Specifically, the majority of previous works ~\citep{2018arXiv180300195Z,2016arXiv160202666M,wu2020noisy,ziyin2021minibatch} argue that the noise is Gaussian, i.e., $u_t \sim \mathcal{N}(0,\lambda(w_t))$, 
where $\lambda(w_t)$ is the noise covariance matrix and formulated as follows:
\begin{align}
\MoveEqLeft[0]\label{eq:cov_noise}
    \lambda(W_t)=\\{}&\nonumber 
    \frac{1}{B} \left[\frac{1}{D}\sum_{j=1}^{D} \nabla U^{(j)}(W_t) \nabla U^{(j)}(W_t)^T -\nabla U(W_t) \nabla U(W_t)^T \right].
\end{align}
Recently, \cite{2018arXiv180300195Z} showed the importance of modeling the SGN as an anisotropic noise. Precisely, they show that an anisotropic noise model improves the approximation of the dynamics of SGD. 
In \cite{simsekli2019tail},
the authors argue that SGN obeys $\mathcal{S\alpha S}$ Lévy distribution
due to SGN's heavy-tailed nature. $\mathcal{S\alpha S}$ Lévy process is described by a single parameter $\alpha_i$, also named “stability parameter,“ and holds a unique property, large discontinuous jumps. Therefore, 
Lévy-driven SDE does
not depend on the height of the potential; on the contrary, it directly depends on the horizontal distance
to the domain's boundary; this implies that the process can escape from narrow minima – no matter how
deep they are and will stay longer in wide minima.

In this work, we claim that the noise of distinct parameters in the DNN distributes differently, and we further argue that it is crucial to incorporate this discrepancy into the SGN model. Hence, we model the training process as Lévy-driven stochastic differential equations (SDEs) in $\mathbb{R}^N$, where each parameter $i$ distributes with a unique $\alpha_i$; this formulation helps us investigate the properties and influence of each parameter on the training process.

Another critical aspect of NN optimization is the learning rate. \citet{bengio2012practical} have argued that the learning rate is “the single most important hyper-parameter” in training DNNs;
we yearn to understand what is the interplay between the learning rate decay (LRdecay) and properties of the SGN. Therefore, we examine the effect of the learning rate scheduler on the training process.
 We argue that decreasing the learning rate improves the properties of the optimization due to attenuation of the noise and not merely reduction of the step size; we brace the above claim using theoretical and experimental evidence.
\newline
\newline
Our contributions can be summarized as follows:
\begin{itemize}
   \item Demonstrate empirically that the SGN of each parameter in a deep neural network is better characterized by $S\alpha S$ distribution.
    \item Provide experimental evidence which strongly indicates that different parametric distributions characterize the noise of distinct parameters.    
    \item Propose a novel dynamical system in $\mathbb{R}^N$ consisting of $N$ one-dimensional Lévy processes with $\alpha_i$-stable components.
    \item Using our framework, we present an approximation of the mean escape time, the probability of escaping the local minima using a specific parameter, and additional properties of the training process near the local minima. 
    \item We prove that parameters with low $\alpha_i$ are associated with a high probability of aiding the training process to exit from the local minima, and show empirical evidence. 
\end{itemize}

\section{Related Work}
\label{sec:related_work}
Stochastic optimization has been demonstrated effective for several applications, including generative modeling \cite{li2020variational}, support recovery \cite{lindenbaum2021randomly,lindenbaum2022refined}, clustering, and many more. 
The study of stochastic dynamics of systems with small random perturbations is a well-established field, first by modeling as Gaussian perturbations \citep{freidlin2012random,kramers1940brownian}, then replaced by Lévy noise with discontinuous trajectories \citep{imkeller2006first,imkeller2010first,imkeller2008metastable,burghoff2015spectral}. Characterizing the noise as L\'evy perturbations has attracted interest in the context of extreme events modeling, such as in climate \citep{ditlevsen1999observation}, physics \citep{brockmann2002levy} and finance \citep{scalas2000fractional}. 

\textbf{Remark} We note that a Symmetric $\alpha$ stable distribution ($S\alpha S$ or Lévy $S\alpha S$) is a heavy-tailed distribution, parameterized by $\alpha$ - the stability parameter, where a smaller $\alpha$ leads to heavier tail (i.e., extreme events are more frequent and with more amplitude), and vice versa.

Modeling SGD using differential equations is a deep-rooted method.
\citet{Li2015DynamicsOS} used an
SDE to approximate SGD and focused on momentum and adaptive parameter tuning
schemes to study the dynamical properties of stochastic optimization. \citet{Mandt2015ContinuousTimeLO} employed a similar procedure 
to derive an SDE approximation for the SGD to study the influence of the value of the learning rate.
\citet{2015arXiv151106251L} showed that an SDE could approximate SGD in a first-order weak approximation. 
The early works in the field have approximated SGD by Langevin dynamic with isotropic diffusion coefficients \citep{sato2014approximation,raginsky2017non, zhang2017hitting}. Later more accurate modeling suggested
\citep{mandt2017stochastic,2018arXiv180300195Z,2021arXiv210509557M} using an anisotropic noise covariance matrix.
Lately, it has been argued \citep{simsekli2019tail} that SGN is better characterized by $S\alpha S$ noise, presenting experimental and theoretical justifications. This model was allegedly refuted by \citep{xie2020diffusion}, claiming that the experiments performed by \citep{simsekli2019tail} are inaccurate since the noise calculation was done across parameters and not across mini-batches. Lévy driven SDEs
Euler approximation literature is sparser than for the Brownian motion SDEs; however, it is still intensely investigated; for more details about the convergence of Euler approximation for Lévy discretization, see  \citep{2010arXiv1009.4728M,protter1997euler,burghoff2015spectral}.

Learning rate decay is an essential technique in training DNNs, investigated first for gradient descent (GD) by \citep{lecun2012efficient}. \citet{kleinberg2018alternative} showed that SGD is equivalent to the convolution of the loss
surface, with the learning rate serving as the conceptual kernel size of the convolution. Hence spurious local minima can be smoothed out; and, the decay of the learning rate later helps the network converge around the
local minimum. \citet{2019arXiv190801878Y} suggested that learning rate decay improves the ability to learn complex separation patterns.

\section{Framework}
\label{sec:framework}
In our analysis, we consider a DNN with $\mathbf{\bar{L}}$ layers and a total of $N$ weights (parameters),
the domain $\mathcal{G}$ is the local environment of a minimum. 
Our framework considers an $N$-dimensional dynamic system, representing the update rule of SGD as a Lévy-driven stochastic differential equation. In contrast to previous works \citep{zhou2020towards,simsekli2019tail}, our framework does not assume that SGN distributes the same for every parameter $l$ in the DNN.
Thus, the SGN of each parameter is characterized by a different $\alpha$. The governing SDE that depicts the SGDs dynamic inside the domain $\mathcal{G}$ at time $t$ is as follows:
\begin{equation}
\label{eq:W_eq}
    W_t = w- \int_{0}^{t}\nabla U(W_p)\,dp + \sum_{l=1}^{N} s_t^{\frac{\alpha_l-1}{\alpha_l}} \epsilon\mathbf{1}^T\lambda_l(t) r_l L_t^l, 
\end{equation}
where $W_t$ is the process that depicts the evolution of DNN weights at time $t$. $L_t^l\in \mathbb{R}$ is a mean-zero $S\alpha S$ Lévy processes with a stable parameter $\alpha_l$. $\lambda_l(t)\in \mathbb{R}^{N}$
is the $l$-th row of the noise covariance matrix, $ \mathbf{1}\in \mathbb{R}^{N}$
is a vector of ones, and its purpose is to sum the $l$-th row of the noise covariance matrix.
$r_l\in \mathbb{R}^{N}$
is a unit vector and we demand $|\langle r_i,r_j \rangle| \neq 1,$ for $i \neq j$, we will use $r_i$ as a one-hot vector.
$s_t$ represents the learning rate scheduler, and $w$ are the initial weights.

\textbf{Remark} $L_t^l$ can be decomposed into a small jump part  $\xi_t^l$, and an independent part with large jumps  $\psi_t^l$, i.e. $L_l=\xi_t^l+\psi_t^l$, more information on $S\alpha S$ process appears in \ref{sec:sas_background}.

Let $\sigma_\mathcal{G} =\inf\{t\geq0:W_t\notin \mathcal{G}\}$ depict the first exit time from $\mathcal{G}$. $\tau_k^l$ denotes the time of the $k$-th largest jump of parameter $l$, which is driven by the process $\psi^l$ , where we define $\tau_0 = 0$. The interval between large jumps is denoted as:  $S_k^l=\tau_k^l-\tau_{k-1}^l$ and is exponentially distributed with mean $\beta_l(t)^{-1}$, while $\tau_k^l$ is gamma distributed $Gamma(k,\beta_l(t))$; where $\beta_l(t)$ is the intensity of the jump and will be defined in Sec~\ref{sec:jump_intensity}.
We define the arrival time of the $k$-th jump of all parameters combined as $ \tau_k^*$, for $k \geq 1$ we can write
\begin{equation}
    \tau_k^* \triangleq \bigwedge_{\tau_j^l> \tau_{k-1}^*} \tau_j^l, 
\end{equation}
following that $S_k^*=\tau_k^*-\tau_{k-1}^*$. Jump heights are notated as: $J_k^l=\psi_{\tau_k}^l-\psi_{\tau_{k^-}}^l$.
We will define $\alpha_\nu$ as the average $\alpha$ over the entire DNN; this will help us describe the global properties of our network.
\newline
 Let us define a measure of horizontal distance from the domain boundary using $d_l^+$ and $d_l^-$; we present a rigorous formulation of our assumptions in Sec.~\ref{sec:assumptions}.
\newline
We define two additional processes to better understand the dynamics inside the basin (between the large jumps).
\paragraph{The deterministic process}denoted as $Y_t$ is affected by the drift alone, without any perturbations. This process starts within the domain and does not escape the domain as time proceeds. The drift forces this process towards the stable point $W^*$ as $t\rightarrow \infty$, i.e., the local minimum of the basin; furthermore, the process converges to the stable point exponentially fast and is defined for $t>0$, and $w\in\mathcal{G}$ by:
\begin{equation}
\label{eq:determe}
    Y_t = w-  \int_{0}^{t}\nabla U(Y_s)\,ds. 
\end{equation}
The following Lemma shows how fast $Y_t$ converges to the local minima from any starting point $w$ inside the domain.
\begin{lemma}
\label{sec:lemma1}
$\forall w\in \mathcal{G}$ , $\tilde{U} = U(w)-U(W^*)$,  the process $Y_t$ converges to the minimum $W^*$ exponentially fast:
\begin{align}
\MoveEqLeft[3]
\norm{Y_t-W^*}^2\leq \frac{2\tilde{U}}{\mu}e^{-2\mu t}.
\end{align}
The complete proof appears in Appendix \ref{sec:det_conv_proof}
\end{lemma}

\paragraph{The small jumps process} $Z_t$ is composed of the deterministic process $Y_t$ and a stochastic process
 with infinite small jumps denoted as $\xi_t$  (see more details in \ref{sec:sas_background}). $Z_t$ describes the system's dynamic in the intervals between the large jumps; hence we add an index $k$ that represents the index of the jump, for instance $Z_{t,k}$ represent the time $t$ between the jump $k$ to jump jump $k+1$.
 Due to strong Markov property, $\xi_{t+\tau}^l-\xi_{\tau}^l, t\geq 0$ is also a L\'evy process with the same law as $\xi^l$.
 Hence, for $t\geq 0$ and $k\geq 0$:
\begin{equation}
\xi_{t,k}^l = \xi_{t+\tau_{k-1}}^l-\xi_{\tau_{k-1}}^l.
\end{equation}
The full small jumps process for $\forall t\in[0,S_{k}]$ is defined as:
\begin{equation}
Z_{t,k} = w + \int_0^t\nabla U(y_s)ds+ \sum_{l=1}^Ns_t^{\frac{\alpha_l-1}{\alpha_l}}\epsilon \mathbf{1}^T\lambda_l(t)r_l \xi_{t,k}^l.
\end{equation}
In the following proposition, we estimate the deviation in the $l$-th parameter between the SDE solution driven by the process of the small jumps $Z_{t,k}^l$, and the deterministic trajectory. \begin{proposition}
\label{prop:prop1}
Let $T_\epsilon> 0$ exponentially distributed with parameter $\beta_l$, $\forall w \in \mathcal{G}$,  and $\bar{\theta}_l \triangleq -\rho(1-\alpha_l)+2-2\theta_l$, s.t.  
 $\theta_l\in(0
 ,\frac{2-\alpha_l}{4})$, the following holds:
\begin{equation}
P\left(\sup_{t \in [0,T_\epsilon]}|Z_{t,k}^l(w)-Y_{t,k}^l(w)| \geq c \bar{\epsilon}^{\theta_l}\right) \leq  C_{\theta_l} \bar{\epsilon}^{\bar{\theta}_l}.
\end{equation}
\end{proposition}
Where $C_{\theta_l}>0$ and $c > 0$ are constants, let us remind the reader that: $\bar{\epsilon}_l = s_t^{\frac{\alpha_l-1}{\alpha_l}}\epsilon_l$.
Precisely, proposition \ref{prop:prop1} describes the distance between the deterministic process  $Y_{t,k}$ and the process of small jumps $Z_{t,k}$ at time $t$ that occurs in the interval after the jump $k$ and before jump $k+1$. It indicates that between large jumps, the processes are close to each other with high probability. The complete proof appears in Appendix \ref{sec:proofprop1}.

 Let us present additional notations: $H()$ and $\nabla U$ are the Hessian and the gradient of the objective function. To denote different mini-batches, we use subscript $d$. That is, $H_d()$ and $\nabla U_d(W^*)$ are the Hessian and gradient of the $d$ mini-batch. To represent different parameters, as before we will use subscript $l$, for example $\nabla u_{d,l}$, is the gradient of the $l$-th parameter after a forward pass over mini-batch $d$. Furthermore, $h_{l,j}$ represents the $l$-th row and $j$-th column of $H(W^*)$, which is the Hessian after a forward pass over the entire dataset $D$, i.e., the Hessian when performing standard gradient descent.
 Next, we turn our attention to another property of the process of the small jumps $Z^l_{t,k}$. This will help us understand the noise covariance matrix. Using stochastic asymptotic expansion, we can approximate $Z_{t,k}^l$ using the deterministic process and a first-order approximation of $Z_{t,k}^l$.
\begin{lemma}
\label{lemma:z_product}
For a general scheduler $s_t$ , $\rho \in (0,1)$, $\forall w_l,w_j\in\mathcal{G}$, starting point after a big jump at time $\tau_k^*+p$ where $p\rightarrow 0$, and $ A_{lj}(t) \triangleq \bar{\epsilon}_lw_je^{-h_{jj}t}    \mu_\xi^l(2t +  \frac{1}{h_{ll}}(1-e^{-h_{ll}t}))$, for $t\in[0,S_{k}^*)$ the following fulfills:
\begin{align}
\MoveEqLeft[3]
    \mathop{\mathbb{E}}[Z_{t,k}^lZ_{t,k}^j]
     =
     w_lw_je^{-(h_{ll}+h_{jj})t}+A_{jl}(t)
     +A_{lj}(t) + \mathcal{O}(\epsilon^2) .
\end{align}
\end{lemma}
Where $\mu_\xi^i=2t\left[\frac{\bar{\epsilon}^{-\rho(1-\alpha_l)}-1}{1-\alpha_l}\right]$, $\bar{\epsilon}_l=s_t^{\frac{\alpha_l-1}{\alpha_l}}\epsilon_l$.
Lemma \ref{lemma:z_product} depicts the dynamics between two parameters in the intervals between the large jumps; this helps us to accurately express the covariance matrix of the noise; the complete derivation of this result appears in Appendix \ref{sec:proof_lemma_z_prod}.

\subsection{Noise covariance matrix}
The covariance of the noise matrix holds a vital role in modeling the training process; in this subsection, we aim to achieve an expression of the noise covariance matrix based on the stochastic processes we presented in the previous subsection. We can achieve the following approximation using stochastic Taylor expansion near the basin $W^*$. 
\begin{proposition}
Let us define $\tilde{u}_{l}= \sum_{j=1}^N\nabla  u_l\nabla u_j$,  $\tilde{h}_{l,m,p,j} \vcentcolon= \frac{1}{B}\sum_{b=1}^B h_{b,l,m}h_{b,p,j}$, $h_{l,m,p,j} \vcentcolon= h_{l,m}h_{p,j}$ and $\bar{h}_{l,m,p,j} \vcentcolon= \tilde{h}_{l,m,p,j} -h_{l,m,p,j} $, then for any $t\in [0,S_{k}^*)$, the sum of the $l$-th row of the covariance matrix:
\label{prop:lambda_t_raw_lemma}
\begin{align}
\label{eq:lambda_t_raw}
\MoveEqLeft[3]
\mathbf{1}^T\lambda_{l}^k(W_t)= 
\frac{1}{BD}
        \sum_{j=1}^N\bar{u}_{lj}+  \\{}&\nonumber
   \frac{1}{B}\sum_{j,m,p=1}^N\bar{h}_{l,m,p,j}(
     w_mw_pe^{-(h_{mm}+h_{pp})t}+ \MoveEqLeft[3] \\{}&\nonumber     A_{mp}(t)
     +A_{pm}(t))+\mathcal{O}(\bar{\epsilon}^2),
\end{align}
\end{proposition}
where $A_{mp}(t)$ and $A_{pm}(t)$ are defined in lemma~\ref{lemma:z_product}. We note that $h_{l,m,p,j}$ and $\tilde{h}_{l,m,p,j}$ represent the interaction of two terms in the Hessian matrix when performing GD and SGD respectively, and $\bar{h}_{l,m,p,j}$ is the difference between them. The proof of the proposition appears in Appendix 
\ref{sec:noise_cov_calc}.
\begin{figure*}[t]
\vspace{0 in}
\begin{tabular}{cc}
\hspace{-0.1in}
 \includegraphics[width=.45\linewidth]{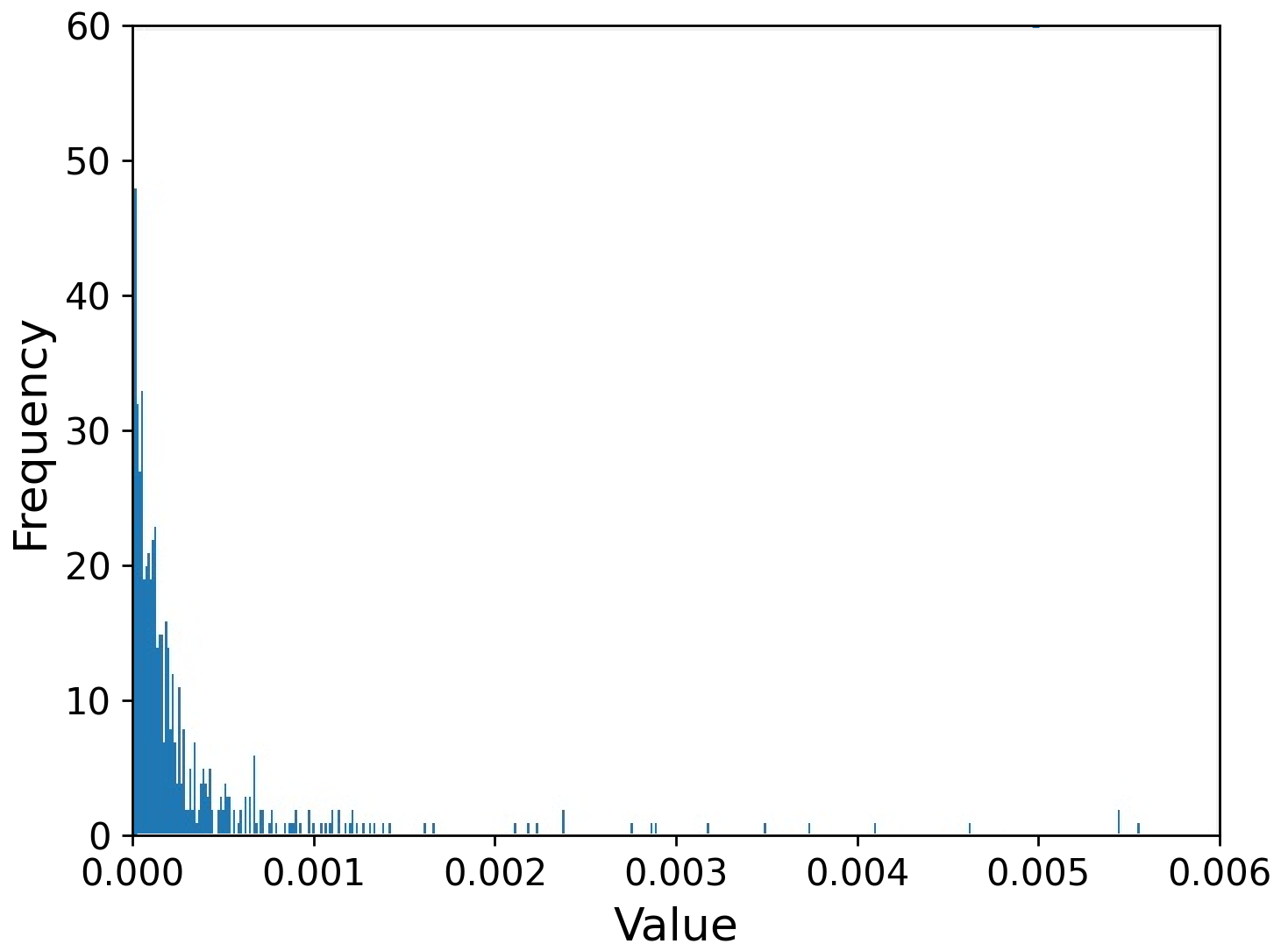}
\includegraphics[width=.45\linewidth]{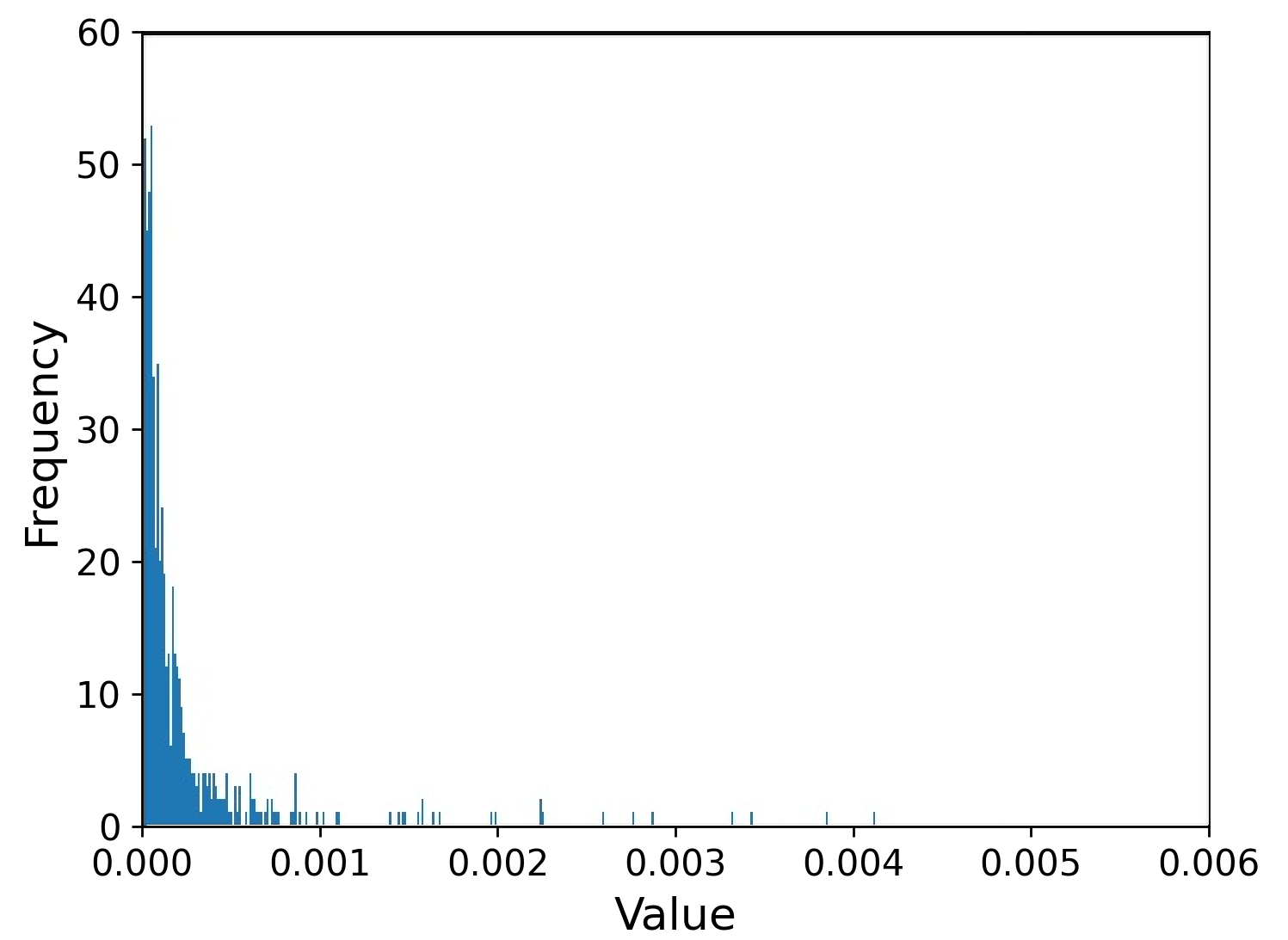}\\
\end{tabular}
 \vspace{-0.1 in}
\caption{Histograms of the stochastic gradient noise for a single parameter in  ResNet34 for :(left) layer number 1, (right) layer number 2. The plots qualitatively shows that SGN is far from Normal distribution, and presents heavy tail nature. }
 \vspace{-0.1 in}
\label{fig:grad_histo}
 \end{figure*}

\subsection{Jump Intensity}
\label{sec:jump_intensity}
Let us denote $\beta_l(t)$ as the jump intensity of the compound Poisson process $\xi_l$. $\beta_l(t)$ simultaneous responsible for scaling of the jump frequency and size. Jumps are distributed according 
to the law $\beta_l(t)^{-1}\nu_\eta$, and the jump intensity is formulated as:
\begin{equation}
    \beta_l(t) = \nu_{\eta_l}(\mathbb{R})= \int_{ \mathbb{R} \mathbin{/} [-O,O]}\nu_l(dy)=\frac{2}{\alpha_i}s_t^
    {\rho(\alpha_l-1)}\epsilon^{\rho\alpha_l}_l,
\end{equation}
\newline
where the integration boundary is $O\triangleq \epsilon^{-\rho}s_t^{-\rho\frac{\alpha_l-1}{\alpha_l}}$, which is time-dependent, due to the learning rate scheduler, which decreases the size and frequency of the large jumps, thus the jump intensity is not stationary. Hence, changing the learning rate during training enables us to increase and decrease the jumps frequency and amplitude. The entire DNN jump intensity as $\beta_S(t )\triangleq \sum_{l=1}^N\beta_l(t)$.

The probability of escaping the local minima in the first jump, in a single parameter perspective, is expressed by:
\begin{equation}
    P(s_t \epsilon\mathbf{1}^T\lambda_l(t)J^l_1\notin[d_l^-,d_l^+])=\frac{m_l(t)\Phi_ls_t^
    {\alpha_l-1}}{\beta_l(t)},
\end{equation}
where $m_l(t)=\frac{\mathbf{1}^T\lambda_l(t)\epsilon_l^{\alpha_l}}{\alpha_l}$, and $\Phi_l=(-d_l^-)^{-\alpha_l}+(d_l^+)^{-\alpha_l}$.

\begin{table}
 \vspace{-0.15 in}
\label{tab:cinic10_fit}.
\centering
\begin{tabular}{ccccccc}
 Model & Gauss &  $S\alpha S $ Const $\alpha$  & $S\alpha S $  \\
  \hline
 ResNet18& $1.39\pm0.41  $& $1.55\pm0.71  $ &$\textbf{0.65}\pm0.27 $\\
 \hline
 ResNet34& $1.58\pm 0.73 $  &$2.31 \pm 1.16$ &$ \textbf{1.15}\pm 0.74 $\\
 \hline
 ResNet50& $1.42\pm 0.73$  & $1.47\pm 0.98$    &$\textbf{0.99}\pm 0.61$\\
 \hline
\end{tabular}
 \vspace{-0.15 in}
\caption{The fitting error between SGN and $S\alpha S$/Gaussian distribution. Averaged over $150$ randomly sampled parameters, three different CNNs trained on the CINIC10 data with a batch size of 400. Sum of Squares Error (SSE) is used to evaluate the fitting error of each distribution, . "Gauss" represents the Gaussian distribution. Our results demonstrate that $S\alpha S$ better depicts SGN. Values in the table were multiplied by 10 to simplify the exposition} 
 \vspace{-0.2 in}
\label{tab:mean_fit_err}
\end{table}
\section{Theorems}
\label{sec:theorems}
In the following section, we provide a theoretical analysis of SGD dynamics during the training of DNNs. Our analysis is based on two empirical pieces of evidence demonstrated in this work; the first is that SGN is indeed heavy-tailed. The second is that each parameter in the DNN's training process has a different stability parameter $\alpha$ drastically affects the noise properties.

Our work will assume that the training process can exit from the domain only at times that coincide with large jumps. This assumption is based on a few realizations; first, the deterministic process $Y_t$ initialized in any point $w\in \mathcal{G}_\delta$, will converge to the local minima of the domain by the positively invariance of the process, see assumptions in ~Appendix \ref{item:positively_invariant}. Second, $Y_t$ converges to the minimum much faster than the average temporal gap between the large jumps; third, using lemma~\ref{sec:lemma1} we conclude that the small jumps are less likely to help the process escape from the local minimum. Next, we will show evidence for the second realization mentioned above, the relaxation time $T_R^l$ is the time for the deterministic process $Y_t^l$, starting from any arbitrary $w\in \mathcal{G}$, to reach an $\bar{\epsilon}_l^\zeta$-neighbourhood of the attractor. For some $C_1>0$, the relaxation time is 
\begin{align}
\MoveEqLeft[3]
   T_R^l =  \max \left\{ \int_{d_l^-}^{-\bar{\epsilon}_l^\zeta}\frac{dy}{-U'(y)_l},\int_{\bar{\epsilon}_l^\zeta}^{d_l^+}
   \frac{dy}{U'(y)_l} \right\}\leq C_1|ln\bar{\epsilon}_l|.
\end{align}

Now, let us calculate the expectation of  $S_k^*=\tau_k^*-\tau_{k-1}^*$, i.e. the interval between the large jumps:
\begin{equation}
   \mathop{\mathbb{E}}[S_k^l]  =\mathop{\mathbb{E}}[\tau_k^l-\tau_{k-1}^l]=\beta_l^{-1}=\frac{\alpha_l}{2}\bar{\epsilon_l}^{-\rho\alpha_l}.
\end{equation}
Since $\bar{\epsilon}\in (0,1)$, usually even $\bar{\epsilon} \ll 1$, it is easy to notice that $\mathop{\mathbb{E}}[S_k^l] \gg T_R$, thus we can approximate that the process $W_t$ is near the neighborhood of the basin, right before the large jumps. This means that it is highly improbable
 that two large jumps will occur before the training process returns to a neighborhood of the local minima.
Using the information described above, we analyze the escaping time for the exponential scheduler and for the multi-step scheduler; expanding our framework for more LRdecay schemes is straightforward. Let us define a constant that will be used for the remaining of the paper: 
$A_{l,\nu} \triangleq (1-\bar{m}_\nu\bar{\beta}_\nu^{-1} \Phi_\nu) (1-\bar{\beta}_l\bar{\beta}_S^{-1})$, for the next theorem we denote:
 $C_{l,\nu,p}\triangleq \frac{2+(\gamma-1)(\alpha_l-1+\rho(\alpha_l-\alpha_\nu))}{{1+(\gamma-1)(\alpha_l-1)}} $, where $C_{l,\nu,p}$ depends on $\alpha_l$, $\gamma$, and on the difference $\alpha_l-\alpha_\nu$.
The following theorem describes the approximated mean escape time for the exponential scheduler:
\begin{theorem}
\label{th:mean_trans_exp}
Given $C_{l,\nu,p}$ and $A_{l,\nu}$, let $s_t$ be an exponential scheduler $s_t=t^{\gamma-1}$, the mean transition time from the domain $\mathcal{G}$: 
\begin{align}
\MoveEqLeft[1]
\mathop{\mathbb{E}}[\sigma_{\mathcal{G}}]\leq
{}\nonumber  
\sum_{l=0}^N A_{l,\nu}^{-1}\frac{\beta_l(\bar{m}_l \Phi_l)^{1-C_{l,\nu,p}}}{\beta_S(1+(\gamma-1)(\alpha_l-1))}
\Gamma\left(C_{l,\nu,p}\right).
 \end{align}
 \end{theorem}
 \noindent
Where $\Gamma$ is the gamma function,  $\bar{m}_l=\frac{\bar{\lambda}_l^{\alpha_l}\epsilon_l^{\alpha_l}}{\alpha_l}$ and $\bar{\beta}_l=\frac{2\epsilon_l^{\rho\alpha_l}}{\alpha_l}$ is the time independent jump intensity.
For the full proof, see Appendix \ref{sec:proofmean_trans}.

It can be observed from Thm.~\ref{th:mean_trans_exp} that as $\gamma$ decreases, i.e., faster learning rate decay, the mean transition time increases. Interestingly, when $\alpha_l \rightarrow 2$ (nearly Gaussian) and $\gamma \rightarrow 0$, the mean escape time goes to infinity, which means that the training process is trapped inside the basin.
\begin{corollary}
Using Thm.~\ref{th:mean_trans_exp}, if the cooling rate is negligible, i.e $\gamma \rightarrow 1$, the mean transition time:
\begin{align}
\MoveEqLeft[3]
\label{eq:col_escape_eq}
\mathop{\mathbb{E}}[\sigma_{\mathcal{G}}] \leq \sum_{l=0}^N A_{l,\nu}^{-1}\frac{1}{\beta_S1^T\bar{\lambda}_l\epsilon^{\alpha_l(1-\rho)} \Phi_l}.
\end{align}
\end{corollary}


 The framework presented in this work enables us to understand in which direction $r_i$  the training process is more probable to exit the basin $\mathcal{G}$, i.e., which parameter is more liable to help the process escape; this is a crucial feature for understanding the training process. The following theorems will be presented for the exponential scheduler but can be expanded for any scheduler.
 \begin{theorem}
 \label{th:exit_dir_exp}
Let $s_t$ be an exponential scheduler $s_t=t^{\gamma-1}$, $C_l\triangleq \frac{(\gamma-1)(\alpha_l-1+\rho(2\alpha_l-\alpha_\nu-\alpha_l))+2}{(\gamma-1)(\alpha_l-1)+1}$, for $\delta \in (0,\delta_0)$, the probability of the training process to exit the basin through the $l$-th parameter is as follows:
\begin{align}
\MoveEqLeft[1]
P(W_\sigma \in \Omega_i^+(\delta))\leq
 \sum_{l=0}^N 
A_{l,\nu}^{-1}
\frac{\bar{m}_i \Phi_i}{\bar{\beta}_i}(d_i^+)^{-\alpha_i}
\\{}&\nonumber
\frac{\beta_l^2(\bar{m}_l\Phi_l)^{-C_l}}{\beta_S((\gamma-1)(\alpha_l-1)+1)}
\Gamma\left(C_l\right).
 \end{align}
 \end{theorem}
\noindent
Let us focus on the terms that describes the $i$-th parameter:
\begin{align}
\MoveEqLeft[3]
P(W_\sigma \in \Omega_i^+(\delta)) \leq
   \frac{\bar{m}_i}{\bar{\beta}_i}(d_i^+)^{-\alpha_i}
   \sum_{l=0}^N \tilde{C}_l,
 \end{align}
where $\tilde{C}_l$ encapsulate all the terms that do not depend on $i$.
When considering SGN as Lévy noise, we can see that the training process needs only polynomial time to escape a basin.
The following result helps us to assess the escaping ratio of two parameters. 
\begin{corollary}
The ratio of probabilities for exiting the local minima from two different DNN parameters is:
\begin{align}
\MoveEqLeft[3]
\frac{P(W_\sigma \in \Omega_l^+(\delta))}{P(W_\sigma \in \Omega_j^+(\delta)) } \leq
   \frac{1^T\lambda_l^{\alpha_l}}{1^T\lambda_j^{\alpha_j}}\epsilon^{(\alpha_l-\alpha_j)(1-\rho)} \frac{(d_l^+)^{-\alpha_l}}{(d_j^+)^{-\alpha_j}}.
 \end{align}
\end{corollary}
Let us remind the reader that $(d_i^+)$ is a function of the horizontal distance from the domain's edge. Therefore,
the first conclusion is that the higher $(d_l^+)$ is, the probability of exiting from the $l$-th direction decreases. However, the dominant term is $\epsilon^{(\alpha_l-\alpha_j)(1-\rho)}$, combining both factors, parameters with lower $\alpha$ will have more chance of being in the escape path.
It can also be seen from the definition of $\beta_l$ that parameters with lower $\alpha$
jump earlier and contribute more significant jump intensities. We can conclude by writing:
\begin{align}
\MoveEqLeft[3]
\frac{P(W_\sigma \in \Omega_l^+(\delta))}{P(W_\sigma \in \Omega_j^+(\delta)) } \propto
   \epsilon^{\Delta_{l,j}},
 \end{align}
 where $\Delta_{l,j}=\alpha_l-\alpha_j$.
\newline
\newline
The next theorem evaluates the probability of exiting the basin after time $u$.
\begin{theorem}
\label{th:ecape_u}Let $s_t = t^{\gamma-1}$, where $\gamma$ is the cooling rate; let us denote two constants that express the effect of the scheduler: $\gamma_l\triangleq 1+(\gamma-1)(\alpha_l-1)$
and $\kappa \triangleq \frac{1+(\gamma-1)(\alpha_l-1+\rho(\alpha_l-\alpha_\nu))}{\gamma_l}$,
for $u>0$:
\begin{align}
\MoveEqLeft[3]P(\sigma>u) \le
\sum_{l=0}^N  A_{l,\nu}^{-1}\frac{\bar{\beta}_l\bar{m}_l \Phi_l}{\bar{\beta}_S\gamma_l(\bar{m}_l\Phi_l)^{\kappa}}   \Gamma\left(\kappa,
\bar{m}_l\Phi_lu^{\gamma_l}\right)\;\;\;.
 \end{align}
 \end{theorem}
 
 \begin{figure*}
 \vspace{0 in}
\begin{tabular}{ccc}
 \includegraphics[width=.3225\linewidth]{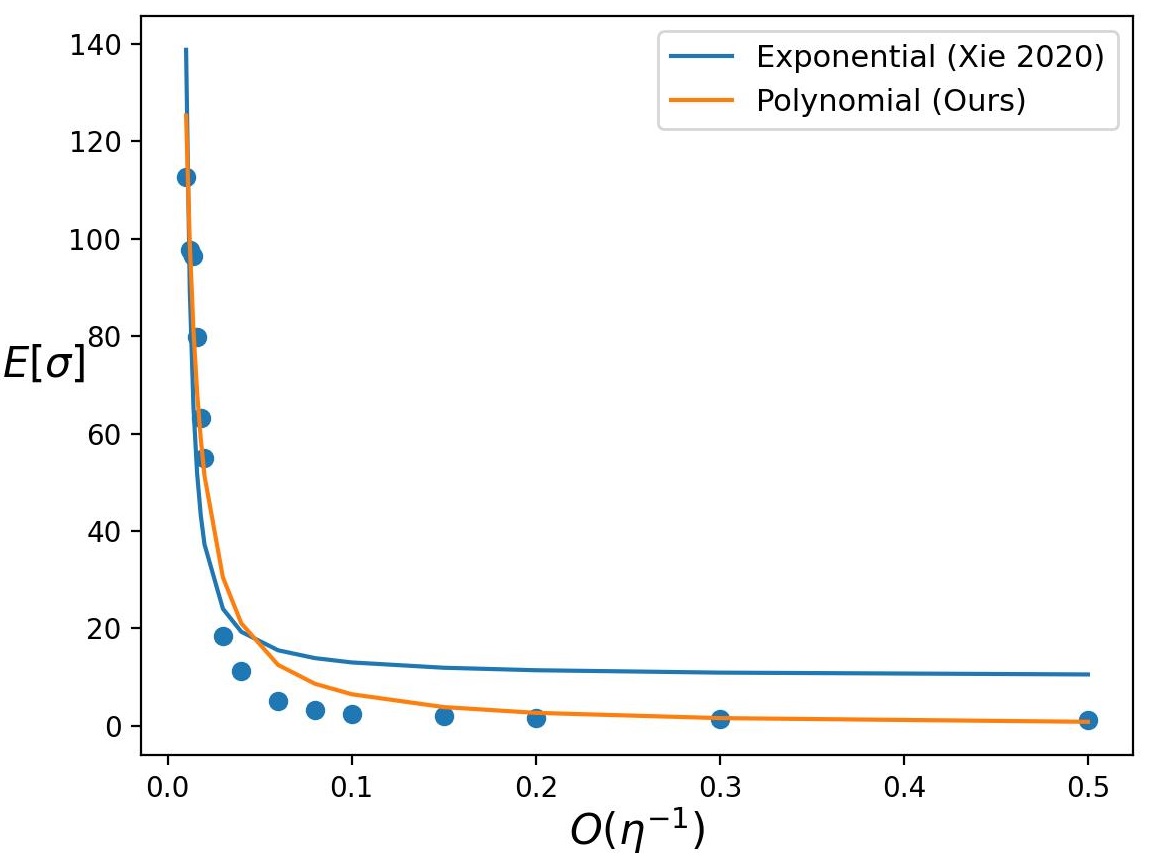}
 \includegraphics[width=.3225\linewidth]{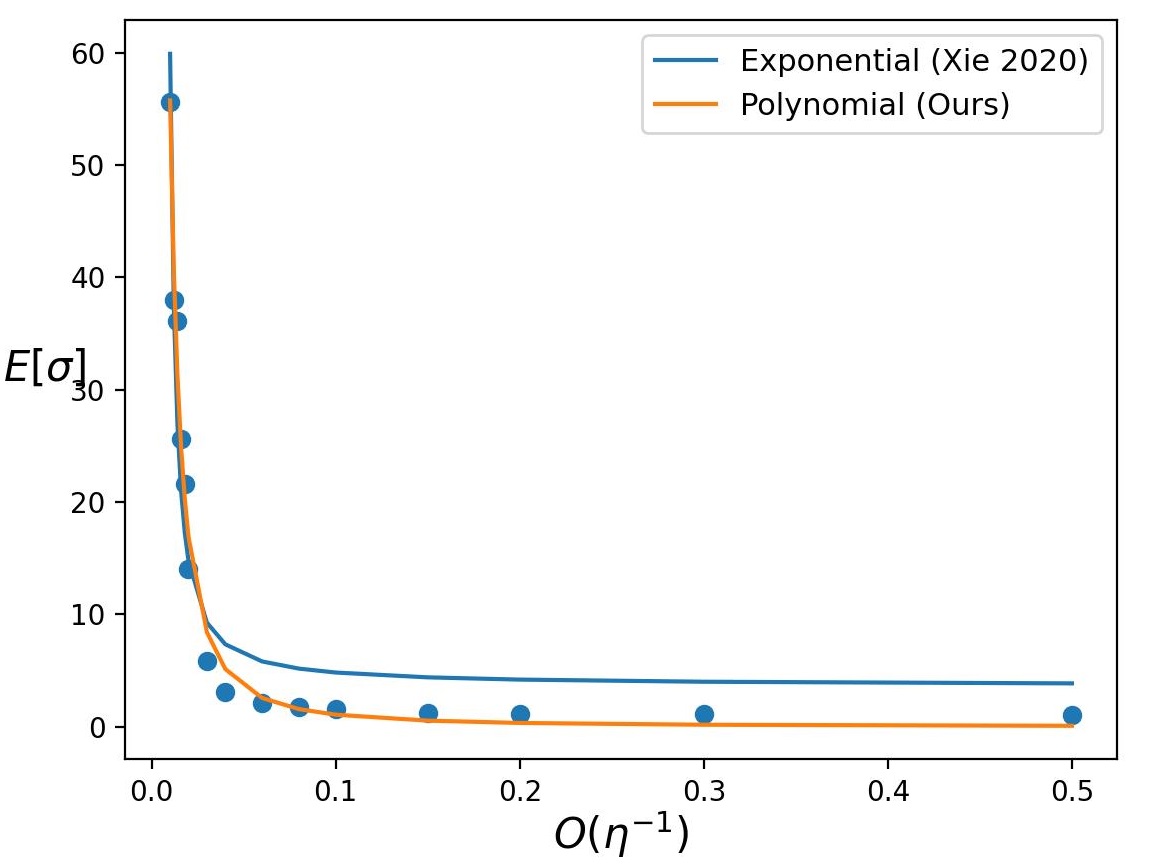}
 \includegraphics[width=.3225\linewidth]{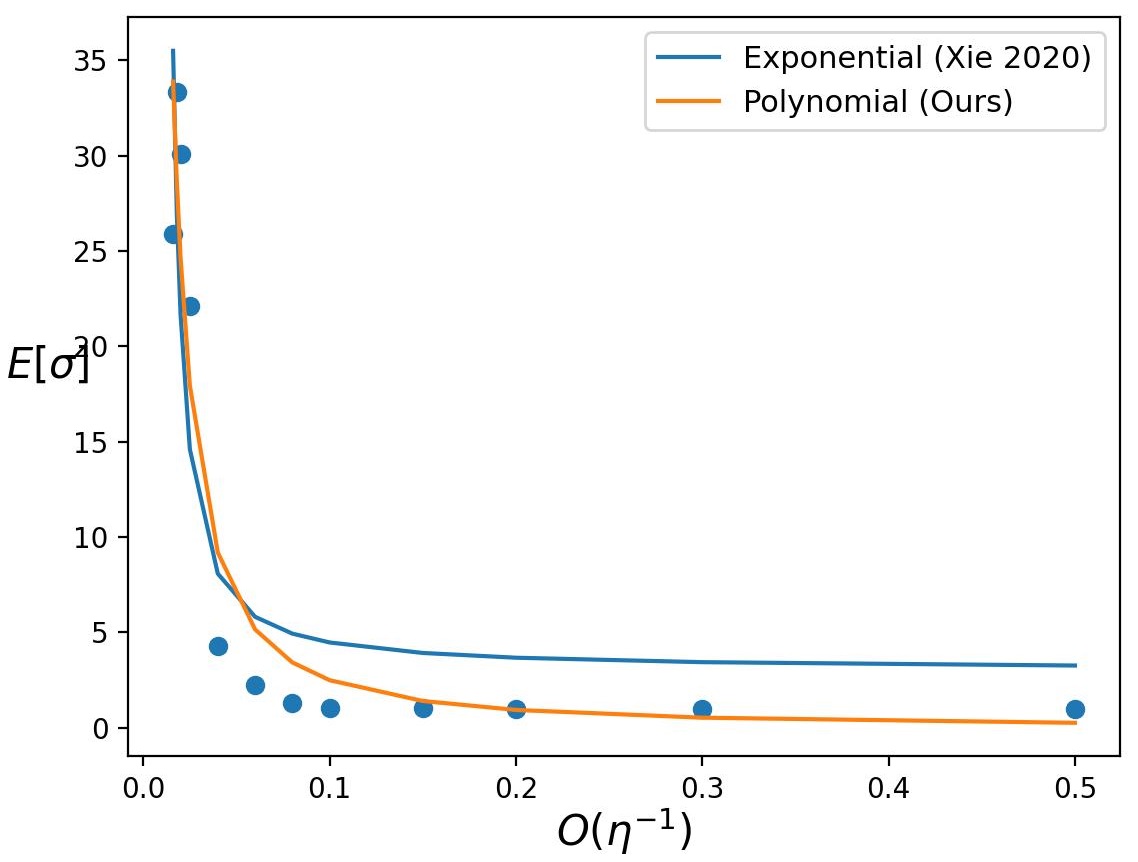} 
 \vspace{0 in}
\end{tabular}
\caption{The mean escape time of SGD on Breastw (left), Cardio (middle), and Satellite (right) datasets. The plots show the fitting base on two  methods: ours and \cite{xie2020diffusion} using a batch size of 32. Each dot represents the mean escape time for a sweep of learning rates. For each learning rate, the dot is an average of over $100$ random seeds. We observe that the empiric results are better explained by our theory for all three datasets examined.}
 \vspace{0 in}
\label{fig:escapetime}
\end{figure*}
\noindent
We now show in a corollary, that the probability of exiting a basin after $u$ iterations decays exponentially with respect to $u$, $\bar{m}_l$, and $\Phi_l$.
\begin{corollary}
\label{cor:escape_u}
Using Thm.~\ref{th:ecape_u},  for $\gamma \rightarrow 1$:
\begin{align}
\MoveEqLeft[3]P(\sigma>u) \leq
\sum_{l=0}^N  A_{l,\nu}^{-1}\frac{\bar{\beta}_l}{\bar{\beta}_S} e^{-\bar{m}_l\Phi_lu}\;\;\;.
 \end{align}
\end{corollary}
 The value $\Phi_l$ describes the horizontal width of the basin, and $\bar{m}_l$ is a function of the learning rate and the noise covariance matrix. Our proof appears in Appendix \ref{ssec:proof_u}.

\section{Experiments}
This section presents the core experimental results supporting our analysis; additional experiments can be found in the Appendix. All the experiments were conducted using SGD without momentum and weight decay. 
\begin{table}
\begin{tabular}{cccccc}
 Model &BS& Gauss &    $S\alpha S $ Const $\alpha$  &  $S\alpha S $ \\
 \hline
 Bert &8 & $2.15 \pm0.64 $&$	1.98 \pm 0.88$ &$	\textbf{0.71} \pm 0.33$ \\
 \hline
 Bert &32&$0.37 \pm 0.33$&$	0.36 \pm 0.19$ &$\textbf{0.18} \pm0.12$\\
 \hline
\end{tabular}
 \vspace{0 in}
\caption{The fitting errors. The errors were computed by averaging $150$ randomly sampled parameters from BERT~\cite{2018arXiv181004805D} base model trained on the Cola dataset. Sum of Squares Error (SSE) is used to evaluate the fitting error of each distribution. Gauss represents the Gaussian distribution.}
 \vspace{0 in}
\label{tab:bert_tab}
\end{table}
\paragraph{Stochastic gradient noise distribution} We empirically show that SGN is better characterized using the $S\alpha S$ Lévy distribution. Unlike previous works \citep{simsekli2019tail,zhou2020towards,xie2020diffusion} we use numeric results to demonstrate the heavy-tailed nature of SGN. Our methodology follows \cite{xie2020diffusion}, calculating the noise of each parameter separately using multiple mini-batches; as opposed to \cite{simsekli2019tail} that calculated the noise of multiple parameters on one mini-batch and averages over all parameters and batches to characterize the distribution of SGN. In \cite{xie2020diffusion}, the authors estimate SGN on a DNN with randomly initialized weights; we, on the other hand, estimate the properties of SGN based on a pre-trained DNN. Expressly, since we want to estimate the escape time, we reason that a pre-trained DNN would better characterize this property. 

We examine the SGN of three ResNet variants and a Bert-base architecture. The ResNets were examined using CINIC10 dataset ~\cite{abs-1810-03505}. Bert's SGN was examined using CoLA~\cite{warstadt2018neural} dataset. The complete technical details appear in appendix \ref{sec:sgn_tech_details}.
We show qualitative and quantitative evidence for SGN's heavy tail nature. The qualitative results in Fig.~\ref{fig:grad_histo} depicts  the histogram of SGN norm, which shows
 the heavy tail nature of the SGN, more visualization for NNs trained on CIFAR100  \cite{Krizhevsky09learningmultiple}
 dataset are available in Sec.~\ref{sec:more_visual_histo}.
Furthermore Fig.~\ref{fig:mean_and_std} shows the $\alpha_i$ values of randomly sampled parameters. In this figure, if the noise of the sampled parameter were Gaussian, we would expect all the blobs to concentrate around $\alpha=2$ (since at this value $S\alpha S$ boils down to a Gaussian distribution). 

The quantitative results depict the fitting error of the empirical distribution of SGN with three 
 distributions: (1) Gaussian \cite{xie2020diffusion}, (2) $S\alpha S$ with constant $\alpha$ \cite{simsekli2019tail}, and (3) $S\alpha S$ with multiple $\alpha_i$ values (ours). The fitting errors for ResNets on  CINIC10~\cite{abs-1810-03505} are shown in Tab.~\ref{tab:cinic10_fit},  and for Bert see Tab.~\ref{tab:bert_tab}, the results shows strong evidence that SGN is best explained by $S\alpha S$ distribution
 .


\begin{figure*}
 \vspace{0 in}
\begin{tabular}{cc}
 \includegraphics[width=.5\linewidth]{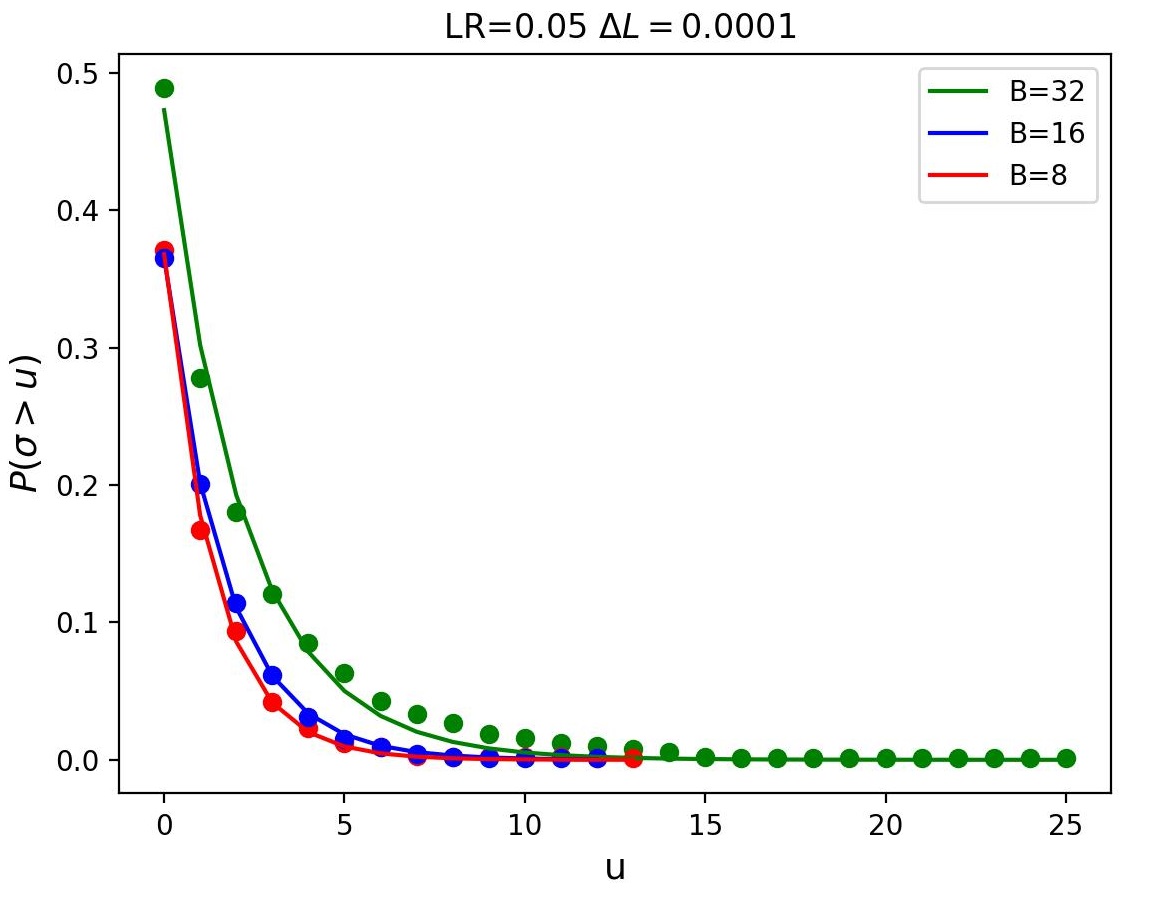}
 \includegraphics[width=.5\linewidth]{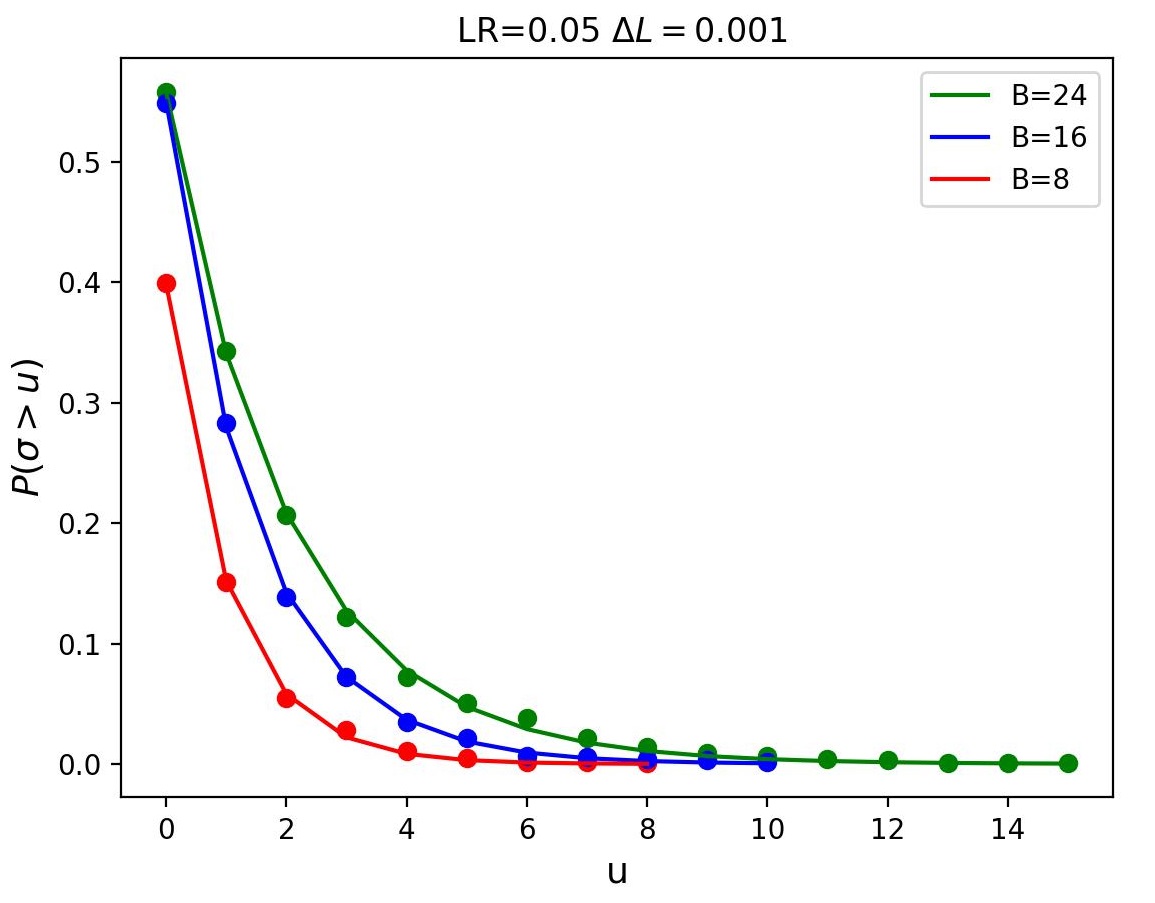}
\end{tabular}
 \vspace{-0.15 in}
\caption{The x-axis represents the number of iterations. The y-axis represents the probability of exiting the basin; We train the same model, 1000 runs, and in each run, we keep the iteration of escaping the basin. The left plot shows results on the Cardio dataset with different mini-batch sizes, and the right plot shows the same on the Speech dataset. The exponential decay predicted by our theorem (lines) coincides with the empirical results (dots). }
 \vspace{-0.15 in}
\label{fig:escape_u}
\end{figure*}


\paragraph{Different parameters hold different noise distributions? } This experiment shows that distinct DNN parameters lead to different SGN during training.
We randomly sampled 100 parameters from five different DNNs. Then, we calculated the SGN and estimated $\alpha_i$ for each parameter; Fig.~\ref {fig:mean_and_std} depicts the results for the five DNNs. We observe that different parameters have noise that distributes differently during training. We can further notice that the variance is stretched on large segments of $\alpha_i$ values. This implies that building a framework that considers the DNN as one homogeneous system is insufficient; each parameter in the DNN has its characteristics, and we should consider this when modeling the noise. Models were trained as detailed in Appendix \ref{sec:sgn_tech_details}.

\paragraph{Mean escape time}
The following experiment validates Theorem~\ref{th:mean_trans_exp}. We trained a three-layer neural network with Relu activation on "BreastW", "Satellite" and "Cardio" datasets \cite{Dua:2019}.
We first train the model using SGD with multiple learning rates and batch size of 256 until reaching a local minimum (see discussion Appendix \ref{sec:select_minima}).   
After reaching the critical point, we decrease the mini-batch size and try to escape the critical minimum,
Fig~\ref{fig:escapetime} shows the escape time using different learning rates. The escape time is measured by the number of iterations, averaged over $100$ seeds. We fit empirical results to two theories, ours and \cite{xie2020diffusion}, both theories fitted with the same amount of free parameters. 
The results in Fig~\ref{fig:escapetime} show the mean escape time using batch size of 32, one can observe that the empiric results are better explained by our theory on all three datasets examined. Our method shows limitation when using small batch sizes, as depicted in Appendix~\ref{sec:escapetime_moreexp}, our theory overshoot when predicting mean escape time for Satellite dataset, competitive on Cardio and better on BreastW dataset. 
\paragraph{Probability of escaping after time u.}
The following experiment validates Thm.~\ref{th:ecape_u}. We trained a three-layer neural network with Relu activation on Speech, Cardio and dataset \cite{Dua:2019} using SGD with a learning rate of 0.05 and batch size of 128 until convergence to a local minima.
We measure the time to escape the local minimum on 1000 seeds and plot the probability distribution to exit as a function of time in Fig.~\ref{fig:escape_u}. This results demonstrates that our theoretical results concide with the empiric evidence.

\paragraph{Learning rate decay} The heavy tail behaviour of SGN may prevent from the training process to converge to a critical point due to the large jump process, hence reducing the frequency and size of the large jumps may be crucial for good convergence. This paragraph aims to demonstrate that the LRdecay's effectiveness may be due to the attenuation of SGN. We show two experiments, first we trained ResNet110 \cite{2015arXiv151203385H} on CIFAR100 \cite{Krizhevsky09learningmultiple}, on epoch $280$ the learning rate is decreased by a factor of $10$. Fig.~\ref{fig:lrdecplots} shows that the learning rate decay results in a lower noise amplitude and less variance. In the second experiment, a ResNet20 \cite{2015arXiv151203385H} is trained in three different variations for 90 epochs; the first variation had LRdecay at epochs 30 and 60, the second had a batch-size increase at epochs 30 and 60, the third was trained with the same learning rate and batch size for the entire training process, the results show almost identical results on the first two cases, (i.e., LRdecay and batch increase) reaching a top-1 score of 66.7 and 66.4 on the validation set. In contrast, the third led to worse performances reaching a top-1 score of 53.
 \cite{smith2017don} performed similar experiments to show the similarity between decreasing the learning rate and increasing the batch size; however, their purpose was to suggest a method for improving training speed without degrading the results.

LRdecay decreases the step size and the noise amplitude; on the other hand, increasing the batch size only decreases the noise amplitude. Combining the results of the two experiments above, we may carefully deduce that the main effect of LRdecay is reducing the fluctuation in the gradient update phase and not decreasing the step size (step size is the movement of the deterministic process towards the minus of the gradient). SGN amplitude reduction enables the training process to get easier localization in the current promising domain.

\paragraph{Escaping Axis} In this section, we demonstrate that the optimization process is more probable to escape from the axis with lower $\alpha_i$. We use a 2D Ackleys function; the escape process starts at the global minimum $\vec{0}$. We apply Gradient Descent with added $S\alpha S$ noise ($S\alpha S(\alpha_{x_1})$,$S\alpha S(\alpha_{x_2})$), where $\alpha_{1}=\alpha_{2}- \Delta$, learning rate of $1e-4$, with no momentum or weight decay. Once the optimization process passes some predefined radius, we check which axis is larger. Fig~\ref{fig:eacape_axis} shows how probable it is to exit from $x_1$ based on 1000 different seeds. This result implies that as the $\delta$ between the $\alpha_i$s increases, the axis with the smaller value of $\alpha$ has more probability of being the axis through which the optimization process can escape.
\section{Conclusions}
Our experiments corroborate that the $S\alpha S$ better characterized SGN qualitatively and quantitatively. Furthermore, we show that distinct parameters are better characterized by different distribution parameters, $\alpha_i$.
Based on the mentioned experiments, we constructed a framework in $\mathbb{R}^N$ consisting of $N$ one-dimensional Lévy processes with $\alpha_i$-stable components. This framework enables us to better understand the nature of DNN training with SGD, such as the escaping properties from different local minima, a learning rate scheduler, and other parameters' effects in the DNN. We also presented experiments that support the claim that a significant feature of LR schedulers comes from reducing the fluctuations of the SGN. Finally, we show that parameters in the DNN that hold noise that distributes with low $\alpha_i$ have a unique role in the training process, helping the training process escape local minima.

\paragraph{Limitations and Future Research} The presented framework is valid once the training process is near a local minimum; how the training acts in other states, for example, at the beginning of the training, is not intended to be solved in this work. 
Further, how $\alpha$ evolves in time is still unclear and demands future research. It is also unclear why different parameters holds different SGN distributions, and what are the roles of each parameter in the optimization process. 

\bibliography{arxiv}
\bibliographystyle{plainnat}

\newpage
\appendix
\onecolumn
\section{Additional details}
Here we provide additional information required to reproduce our results and for completeness of our exposition.
\subsection{Notations}
\begin{table}[h]
    \centering
\begin{tblr}{
      colspec={|X|X[3]|}, row{1} = {c}, hlines,
    }
    Symbol & Description  \\
 t & Train iteration\\
 $S\alpha S$ & Symmetric $\alpha$ stable \\
 U& Potential/ loss function\\
 $W_t$ & The process that depicts DNN weights time evolution . \\
  $Y_t$ & The deterministic process. \\
 $Z_t$ & The small jumps process\\
 $L_t^l$ &  Mean-zero $S\alpha S$ Lévy processes in 1d- represent the SGN of the l-th parameter  \\
 $\psi_t$ & Large jump part of $L_t$ \\
 $\xi_t$ & Small jump part of $L_t$ \\
 $\eta$  & Learning rate\\
  B & Batch size\\
  $\Omega$ & Batch sample ($|\Omega|=B$) \\
  D  & Number of samples in training datasets\\
  $s_t$  & LR scheduler at time t \\
  $\gamma$  & Cooling rate \\
  $\alpha$& Stability parameter of $S\alpha S$ dist.\\
  $\lambda$ & Noise covariance matrix\\
  $\tau_k^l$& The time of the k-th large jump of parameter $l$\\     $S_k^l$& The difference between the (k-1)-th large jump and k-th large jump of parameter $l$\\ 
  $\beta_l$ & The jump intensity of the compound Poisson process $\xi_l$\\
  J&Large jumps height \\
  \end{tblr}
\end{table}

\subsection{Technical details}
\label{sec:sgn_tech_details}
We trained several CNNs on CINIC dataset ~\cite{abs-1810-03505} and the BERT base model on CoLA~\cite{warstadt2018neural} dataset. All models are trained until reaching
convergence.
Using the pre-trained weights, we sample $100$ random parameters; for each parameter, we estimate the noise by computing the gradients of all of the mini-batches in the dataset without updating the weights.
Then, we fitted the empiric stochastic gradient noise to multiple distributions; Sum of square error (SSE) is used to evaluate the quality of our fit.
 We trained four ResNet variants Resnet18/34/50, those models were trained using SGD optimizer, learning rate of 0.01, and a batch size of 128 (\ref{sec:select_minima}). We used a multistep learning rate scheduler on epochs 200 and 400 to accelerate the convergence. 
 We examine the SGD noise of BERT model, which was fine-tuned on CoLA~\cite{warstadt2018neural} dataset using Adam optimizer with a learning rate of 2e-05 and batch size of 32 for 20 epochs. This is the standard Bert fine-tuning procedure. The results are shown in Tab.~\ref{tab:bert_tab}. 
Visual examples for the heavy-tailed nature of SGN can be seen in Fig.~\ref{fig:grad_histo}, and additional results are presented in Sec.~\ref{sec:more_visual_histo}. These results corroborates our claim of the heavy-tailed nature of SGN, even for different DNN architectures (CNN and Transformer based models) and input domains (text and images).
\subsection{$S\alpha S$ background}
\label{sec:sas_background}
A L\'evy process is random with independent and stationary
increments, continuous in probability, and possesses right-continuous paths with left
limits. Except for special cases, its probability density does not generally have a closed-form formula. Hence the process is characterized by the L\'evy–Hincin formula. In this paper, the noise is assumed to be best fitted by symmetric $\alpha$ stable L\'evy distribution, also known as L\'evy flights (LF), and mainly parameterized using a stability parameter $\alpha$, hence the characteristic function: 
\begin{align}
\MoveEqLeft[1]
    \mathop{\mathbb{E}}[e^{iwL_t^l}]= 
    exp\{-t{ \int_{\mathbb{R} \mathbin{/}\{0\}}[e^{iwy}-1-iwy}\mathbb{I}\{|y|\leq 1\}]\frac{dy}{|y|^{1+\alpha_l}}\},
\end{align}
where $\mathbb{I}\{B\}$ denotes the indicator function of a set
with the corresponding generating triplet $(0,\nu_l,0)$ and the  L\'evy measure $ \nu_l(dy) =|y|^{-1-\alpha_l},y \neq 0 ,\alpha_l \in (0,2)$.In this work we assume $\alpha_l \in (0.5,2)$. Unlike Brownian motion which almost surely holds 
continuous path, Lévy motion might obtain large discontinuous jumps.
  Using Lévy-
Itô-decomposition of $L^l$ can be decomposed into a small jump part  $\xi_t^l$, and an independent part with large jumps $\psi_t^l$, i.e., $L_i=\xi_t^l+\psi_t^i$. \newline
The process $\xi_t^l$ has an infinite L\'evy measure with support:$\{y| 0 < \norm{y}\leq \epsilon_l^{-\rho}\}$,$\forall \rho\in (0,1)$, and makes infinitely
many jumps on any time interval. The absolute value of $\xi_t^l$ jumps is bounded by  $\epsilon^{-\rho}$.\newline
$\psi_t^i$ is a compound Poisson process with finite L\'evy measure, and is responsible on the big jumps, more details about $\psi^i_t$ in Sec.~\ref{sec:jump_intensity}
\subsection{Selecting minimum point}
\label{sec:select_minima}
 In order to find local minimum, we measure the loss of the entire data, i.e. loss when running GD; if the loss does not change more then $\epsilon$ for more then 100 iterations, we exit the training process and select the checkpoint as a minimum point. Since we do not know the domain boundary of the current minimum, we measure the number of iterations until the training process passes a predefined loss delta ($\Delta L$) from the current local minimum.
\subsection{Assumption on the Potential near critical points}
\label{sec:disscuss_on_assumption}
We assume that the potential $U(W_t)$ is $\mu$-strongly convex and can be approximated by a second order Taylor approximation near critical points that will be noted as $W^*$:
\begin{equation}
    U(W) = U(W^*)+\nabla U(W^*)(W-W^*)+\frac{1}{2}(W-W^*)^TH(W^*)(W-W^*)
\end{equation}
This does not mean that $U(W)$ fulfills any of the assumptions above in general. 
\section{Proofs} 
\everypar{\looseness=-1}
\subsection{Proof of Theorem~\ref{th:mean_trans_exp}}
The first equality is true under the assumption that the process can exit the basin only when large jumps occur.
\label{sec:proofmean_trans}
\begin{align}
\MoveEqLeft[3]
 \mathop{\mathbb{E}}[\sigma_{\mathcal{G}}]= \sum_{k=1}^{\infty}\mathop{\mathbb{E}}[\tau_k^*]\mathbbm{I}\{\sigma_{\mathcal{G}}=\tau_k^*\}]\\{}&\nonumber 
\nonumber
 =\sum_{k=1}^{\infty}\mathop{\mathbb{E}}[\tau_k^*\mathbbm{I}\{\sum_{l=0}^Ns_t\epsilon\mathbf{1}^T\lambda_l(t)J_1^l\mathbbm{I}\{\tau_1^l=\tau_1^*\} \in \mathcal{G},\sum_{l=0}^Ns_t\epsilon\mathbf{1}^T\lambda_l(t)J_2^l\mathbbm{I}\{\tau_2^l=\tau_2^*\} \in \mathcal{G},..,\sum_{l=0}^Ns_t\epsilon\mathbf{1}^T\lambda_l(t)J_k^l\mathbbm{I}\{\tau_k^l=\tau_k^*\} \notin \mathcal{G}\}]
 \\{}&\nonumber \nonumber
\sum_{k=1}^{\infty}\mathop{\mathbb{E}}[\tau_k^*\mathbbm{I}\{J_1^* \in \mathcal{G},J_2^* \in \mathcal{G},..,J_k^* \notin \mathcal{G}\}] \leq
\sum_{k=1}^{\infty}\mathop{\mathbb{E}}[\tau_k^*(1- \mathbbm{I}\{J^{*}_k \notin \mathcal{G}\})^{k-1} \mathbbm{I}\{J^{*}_k \notin \mathcal{G}\}]
  \\{}&\nonumber \nonumber=
  \sum_{k=1}^{\infty} \sum_{l=1}^{N}\mathop{\mathbb{E}}[\tau_k^l(1- \mathbbm{I}\{J^{l}_k \notin \mathcal{G}\})^{k-1} \mathbbm{I}\{J^{l}_k \notin \mathcal{G}\}\mathbbm{I}\{\tau_k^l=\tau_k^*\}]
  \\{}&\nonumber \nonumber \leq
 \sum_{k=1}^{\infty} \sum_{l=1}^{N}\sum_{w=1}^{k}\mathop{\mathbb{E}}[\tau_w^l(1- \mathbbm{I}\{s_t^{\frac{\alpha_l-1}{\alpha_l}} \epsilon\mathbf{1}^T\lambda_l(t)J^{l}_w \notin \mathcal{G}\})^{w-1}(1- \mathbbm{I}\{s_t^{\frac{\alpha_\nu-1}{\alpha_\nu}} \epsilon\mathbf{1}^T\lambda_\nu(t)J^{m}_w \notin \mathcal{G}\})^{k-w}
 \\{}&\nonumber \nonumber
 \mathbbm{I}\{s_t^{\frac{\alpha_l-1}{\alpha_l}} \epsilon\mathbf{1}^T\lambda_i(t)J^{l}_w \notin \mathcal{G}\}\mathbbm{I}\{\tau_{w}^l=\tau_k^*\}]\;\;\;.
 \label{eq:first_main_proof}
\end{align}
$\mathbbm{I}\{\tau_{w}^l=\tau_k^*\}$ incorporates the probability that the $k$-th jump occurred by the $l$-th parameter, and the chance that within a total of $k$ jumps the parameter $l$, will respect the $w$-th jump:
\begin{align}
\MoveEqLeft[3]
\mathbbm{I}\{\tau_{w}^l=\tau_k^*\} = \frac{\beta_l(t)}{\beta_S(t)}{k-1\choose w-1}\left(\frac{\beta_l(t)}{\beta_S(t)}\right)^{w-1}\left(1-\frac{\beta_l(t)}{\beta_S(t)}\right)^{k-w}\\{}&\nonumber
 \frac{\beta_l(t)}{\beta_S(t)}\frac{(k-1)!}{(w-1)!(k-w)!}\left(\frac{\beta_l(t)}{\beta_S(t)}\right)^{w-1}\left(1-\frac{\beta_l(t)}{\beta_S(t)}\right)^{k-w}\;\;\;
\label{eq:Itw}
\end{align}
We will estimate the average probability of the DNN to escape the basin i.e. the general expression: $[1-\frac{s_t^{\alpha_\nu-1}m_\nu(t)}{\beta_\nu(t)} \Phi_\nu]^{k-w}$, by using $\alpha_{\nu}$ as the average $\alpha$ value of the network.
\begin{align}
\MoveEqLeft[1]
\sum_{k=1}^\infty  \sum_{l=0}^N \sum_{w=1}^{k} \int_0^\infty \frac{\beta_l(t)}{\beta_S(t)}\frac{(k-1)!}{(w-1)!(k-w)!}\left(\frac{\beta_l(t)}{\beta_S(t)}\right)^{w-1}\left(1-\frac{\beta_l(t)}{\beta_S(t)}\right)^{k-w}\beta_l(t)t\\{}&\nonumber \nonumber e^{-\beta_l(t)t}\frac{(\beta_l(t)t)^{w-1}}{(w-1)!} [1-\frac{s_t^{\alpha_l-1}m_l(t)}{\beta_l(t)} \Phi_l]^{w-1}
[1-\frac{s_t^{\alpha_\nu-1}m_\nu(t)}{\beta_\nu(t)} \Phi_\nu]^{k-w}\frac{s_t^{\alpha_l-1}m_l(t)}{\beta_l(t)} \Phi_ldt
\\{}&\nonumber \nonumber= 
\sum_{k=1}^\infty  \sum_{l=0}^N \int_0^\infty \frac{\beta_l(t)}{\beta_S(t)}t e^{-\beta_l(t)t} s_t^{\alpha_l-1}m_l(t) \Phi_l
\\{}&\nonumber \nonumber
\sum_{w=1}^{k}\frac{[\beta_l(t)t-s_t^{\alpha_l-1}m_l(t) \Phi_lt]^{w-1}}{(w-1)!}\frac{(k-1)!}{(w-1)!(k-w)!}\left(\frac{\beta_l(t)}{\beta_S(t)}\right)^{w-1}\left[\left(1-\frac{s_t^{\alpha_\nu-1}m_\nu(t)}{\beta_\nu(t)} \Phi_\nu\right)\left(1-\frac{\beta_l(t)}{\beta_S(t)}\right)\right]^{k-w}dt
\\{}&\nonumber \nonumber=
\sum_{k=1}^\infty  \sum_{l=0}^N \int_0^\infty \frac{\beta_l(t)}{\beta_S(t)}t e^{-\beta_l(t)t} s_t^{\alpha_l-1}m_l(t) \Phi_l
\\{}&\nonumber \nonumber
\left[\left(1-\frac{s_t^{\alpha_\nu-1}m_\nu(t)}{\beta_\nu(t)} \Phi_\nu\right)\left(1-\frac{\beta_l(t)}{\beta_S(t)}\right)\right]^{k-1}L_{k-1}\left(\frac{\frac{\beta_l(t)}{\beta_S(t)}(s_t^{\alpha_l-1}m_l(t) \Phi_lt-\beta_l(t)t)}{\left[\left(1-\frac{s_t^{\alpha_\nu-1}m_\nu(t)}{\beta_\nu(t)} \Phi_\nu\right)\left(1-\frac{\beta_l(t)}{\beta_S(t)}\right)\right]}\right)
\\{}&\nonumber \nonumber=
 \sum_{l=0}^N \int_0^\infty \frac{\beta_l(t)}{\beta_S(t)}t e^{-\beta_l(t)t} s_t^{\alpha_l-1}m_l(t) \Phi_l
 \\{}&\nonumber \nonumber
\left[\left(1-\frac{s_t^{\alpha_\nu-1}m_\nu(t)}{\beta_\nu(t)} \Phi_\nu\right)\left(1-\frac{\beta_l(t)}{\beta_S(t)}\right)\right]^{-1}
e^{-\frac{\frac{\beta_l(t)}{\beta_S(t)}(s_t^{\alpha_l-1}m_l(t) \Phi_lt - \beta_l(t)t)}{\left[1-\left(1-\frac{s_t^{\alpha_\nu-1}m_\nu(t)}{\beta_\nu(t)} \Phi_\nu\right)\left(1-\frac{\beta_l(t)}{\beta_S(t)}\right)\right]}}
dt 
\\{}&\nonumber \nonumber=
\sum_{l=0}^N \int_0^\infty \frac{\beta_l(t)}{\beta_S(t)}t e^{-\beta_l(t)t} s_t^{\alpha_l-1}m_l(t) \Phi_l
\\{}&\nonumber \nonumber
\left[\left(1-\frac{s_t^{\alpha_\nu-1}m_\nu(t)}{\beta_\nu(t)} \Phi_\nu\right)\left(1-\frac{\beta_l(t)}{\beta_S(t)}\right)\right]^{-1}
e^{-\frac{\frac{\beta_l(t)}{\beta_S(t)}(s_t^{\alpha_l-1}m_l(t) \Phi_lt - \beta_l(t)t)}{\left[\frac{s_t^{\alpha_\nu-1}m_\nu(t)}{\beta_\nu(t)}\Phi_\nu + \frac{\beta_l(t)}{\beta_S(t)} -\frac{s_t^{\alpha_\nu-1}m_\nu(t)\Phi_\nu}{\beta_\nu(t)}\frac{\beta_l(t)}{\beta_S(t)}\right]}}
dt 
 \\{}&\nonumber \nonumber
 \leq
\sum_{l=0}^N \int_0^\infty \frac{\beta_l(t)}{\beta_S(t)}t e^{-\beta_l(t)t} s_t^{\alpha_l-1}m_l(t) \Phi_l
\\{}&\nonumber \nonumber
\left[\left(1-\frac{\bar{m}_\nu}{\bar{\beta}_\nu} \Phi_\nu\right)\left(1-\frac{\bar{\beta}_l}{\bar{\beta}_S}\right)\right]^{-1}
e^{-(s_t^{\alpha_l-1}m_l(t) \Phi_lt - \beta_l(t)t) }
dt 
\\{}&\nonumber \nonumber\leq
\sum_{l=0}^N 
\left[\left(1-\frac{\bar{m}_\nu}{\bar{\beta}_\nu} \Phi_\nu\right)\left(1-\frac{\bar{\beta}_l}{\bar{\beta}_S}\right)\right]^{-1}
\int_0^\infty \frac{\beta_l}{\beta_S}t s_t^{\alpha_l-1+\rho(\alpha_l-\alpha_\nu)}\bar{m}_l \Phi_l
e^{-s_t^{\alpha_l-1}\bar{m}_l \Phi_lt  }
dt 
\\{}&\nonumber \nonumber=
\sum_{l=0}^N 
\left[\left(1-\frac{\bar{m}_\nu}{\bar{\beta}_\nu} \Phi_\nu\right)\left(1-\frac{\bar{\beta}_l}{\bar{\beta}_S}\right)\right]^{-1}
\int_0^\infty \frac{\beta_l}{\beta_S} t^{1+(\gamma-1)(\alpha_l-1+\rho(\alpha_l-\alpha_\nu))}\bar{m}_l \Phi_l
e^{-t^{1+(\gamma-1)\rho(\alpha_l-1)}\bar{m}_l \Phi_l}
dt 
\\{}&\nonumber \nonumber=
\sum_{l=0}^N \left[\left(1-\frac{\bar{m}_\nu}{\bar{\beta}_\nu} \Phi_\nu\right)\left(1-\frac{\bar{\beta}_l}{\bar{\beta}_S}\right)\right]^{-1}\frac{\beta_l\bar{m}_l \Phi_l}{\beta_S(1+(\gamma-1)\rho(\alpha_l-1))}
 \\{}&\nonumber \nonumber
 \left[ (\bar{m}_l \Phi_l)^{-\frac{2+(\gamma-1)(\alpha_l-1+\rho(\alpha_l-\alpha_\nu))}{{1+(\gamma-1)\rho(\alpha_l-1)}}}\Gamma\left(\frac{2+(\gamma-1)(\alpha_l-1+\rho(\alpha_l-\alpha_\nu))}{{1+(\gamma-1)\rho(\alpha_l-1)}}\right)
\right]
dt 
\\{}&\nonumber \nonumber=
\sum_{l=0}^N A_{l,\nu}^{-1}\frac{\beta_l\bar{m}_l \Phi_l}{\beta_S(1+(\gamma-1)\rho(\alpha_l-1))}
(\bar{m}_l \Phi_l)^{-C_{l,\nu,p}}\Gamma\left(C_{l,\nu,p}\right)
dt \;\;\;.
\end{align}
Where $A_{l,\nu} \triangleq \left[\left(1-\frac{\bar{m}_\nu}{\bar{\beta}_\nu} \Phi_\nu\right)\left(1-\frac{\bar{\beta}_l}{\bar{\beta}_S}\right)\right]$,$C_{l,\nu,p}\triangleq \frac{2+(\gamma-1)(\alpha_l-1+\rho(\alpha_l-\alpha_\nu))}{{1+(\gamma-1)\rho(\alpha_l-1)}} $.
Further to ease the calculation assumed that the time dependency: $\frac{\beta_l(t)}{\beta_S(t)}=\frac{\bar{\beta}_l}{\bar{\beta}_S}s^{\rho(\alpha_l-\alpha_{\nu})}$.
If the cooling rate is negligible, i.e. $\gamma \rightarrow 1$, the mean transition time:
 \begin{align}
\MoveEqLeft[3]
\mathop{\mathbb{E}}[\sigma_{\mathcal{G}}] \leq \sum_{l=0}^N A_{l,\nu}^{-1}\frac{1}{\beta_S1^T\lambda_l\epsilon^{\alpha_l(1-\rho)} \Phi_l}.
 \end{align}
\newline
\subsection{Proof of Theorem~\ref{th:exit_dir_exp}}
 
\begin{align}
\MoveEqLeft[3]
P(W_\sigma \in \Omega_i^+(\delta)) = \sum_{k=1}^{\infty}\prod_{j=1}^{k-1}P(J_j^*\in \mathcal{G})P(J_k^*\in
\Omega_i^+)\\{}&\nonumber =\sum_{k=1}^{\infty}\prod_{j=1}^{k-1}P(J_j^*\in \mathcal{G})P(J_k^* > d_i^+)
\\{}&\nonumber \nonumber=
\sum_{k=1}^{\infty}(1-P(J_k^*\notin \mathcal{G}))^{k-1}P(J_k^* \geq d_i^+)
\\{}&\nonumber \nonumber
\leq
\sum_{k=1}^{\infty}\sum_{l=1}^{N}
\sum_{w=1}^{k-1}(1-P(s_t^{\frac{\alpha_l-1}{\alpha_l}}\epsilon\mathbf{1}^t\lambda_l(t)J_w^l\notin \mathcal{G}))^{w-1}(1-P(s_t^{\frac{\alpha_\nu-1}{\alpha_\nu}}\epsilon\mathbf{1}^t\lambda_\nu(t)J_w^v\notin \mathcal{G}))^{k-w}P(J_w^l \geq d_i^+)P(\tau_w^l=\tau_k^*)
\\{}&\nonumber \nonumber 
=
\sum_{k=1}^{\infty}\sum_{l=1}^{N}
\sum_{w=1}^{k-1}\int_0^\infty  \frac{\beta_l(t)}{\beta_S(t)}\frac{(k-1)!}{(w-1)!(k-w)!}\left(\frac{\beta_l(t)}{\beta_S(t)}\right)^{w-1}\left(1-\frac{\beta_l(t)}{\beta_S(t)}\right)^{k-w}\beta_l(t)\\{}&\nonumber \nonumber e^{-\beta_l(t)t}\frac{(\beta_l(t)t)^{w-1}}{(w-1)!} [1-\frac{s_t^{\alpha_l-1}m_l(t)}{\beta_l(t)} \Phi_l]^{w-1}[1-\frac{s_t^{\alpha_\nu-1}m_\nu(t)}{\beta_\nu(t)} \Phi_\nu]^{k-w}\frac{s_t^{\alpha_i-1}m_i(t)}{\beta_i(t)} (d_i^+)^{-\alpha_i}
\\{}&\nonumber \nonumber
=
\sum_{k=1}^{\infty}\sum_{l=1}^{N}
\int_0^\infty  \frac{\beta_l(t)}{\beta_S(t)} \beta_l(t)e^{-\beta_l(t)t} \frac{s_t^{\alpha_i-1}m_i(t)}{\beta_i(t)} (d_i^+)^{-\alpha_i}
\\{}&\nonumber \nonumber
\sum_{w=1}^{k-1}[1-\frac{s_t^{\alpha_l-1}m_l(t)}{\beta_l(t)} \Phi_l]^{w-1}\frac{(k-1)!}{(w-1)!(k-w)!}\frac{(\beta_l(t)t)^{w-1}}{(w-1)!}\left((1-\frac{s_t^{\alpha_\nu-1}m_\nu(t)}{\beta_\nu(t)} \Phi_\nu)\left(1-\frac{\beta_l(t)}{\beta_S(t)}\right)\right)^{k-w}\left(\frac{\beta_l(t)}{\beta_S(t)}\right)^{w-1}dt
\\{}&\nonumber \nonumber
=
\sum_{k=1}^\infty  \sum_{l=0}^N \int_0^\infty \frac{\beta_l(t)}{\beta_S(t)}\beta_l(t)e^{-\beta_l(t)t} \frac{s_t^{\alpha_i-1}m_i(t)}{\beta_i(t)} (d_i^+)^{-\alpha_i}
\\{}&\nonumber \nonumber
\left[\left(1-\frac{s_t^{\alpha_\nu-1}m_\nu(t)}{\beta_\nu(t)} \Phi_\nu\right)\left(1-\frac{\beta_l(t)}{\beta_S(t)}\right)\right]^{k-1}L_{k-1}\left(\frac{\frac{\beta_l(t)}{\beta_S(t)}(s_t^{\alpha_l-1}m_l(t) \Phi_lt-\beta_l(t)t)}{\left[\left(1-\frac{s_t^{\alpha_\nu-1}m_\nu(t)}{\beta_\nu(t)} \Phi_\nu\right)\left(1-\frac{\beta_l(t)}{\beta_S(t)}\right)\right]}\right)dt
\\{}&\nonumber \nonumber=
  \sum_{l=0}^N \int_0^\infty \frac{\beta_l(t)}{\beta_S(t)}\beta_l(t)e^{-\beta_l(t)t} \frac{s_t^{\alpha_i-1}m_i(t)}{\beta_i(t)} (d_i^+)^{-\alpha_i}
\\{}&\nonumber \nonumber
\left[\left(1-\frac{s_t^{\alpha_\nu-1}m_\nu(t)}{\beta_\nu(t)} \Phi_\nu\right)\left(1-\frac{\beta_l(t)}{\beta_S(t)}\right)\right]^{-1}
e^{-\frac{\frac{\beta_l(t)}{\beta_S(t)}(s_t^{\alpha_l-1}m_l(t) \Phi_lt - \beta_l(t)t)}{\left[\frac{s_t^{\alpha_\nu-1}m_\nu(t)\Phi_\nu}{\beta_\nu(t)} + \frac{\beta_l(t)}{\beta_S(t)} -\frac{s_t^{\alpha_\nu-1}m_\nu(t)\Phi_\nu}{\beta_\nu(t)}\frac{\beta_l(t)}{\beta_S(t)}\right]}}
dt
\\{}&\nonumber \nonumber
 \leq
  \sum_{l=0}^N \int_0^\infty \frac{\beta_l(t)}{\beta_S(t)}\beta_l(t)e^{-\beta_l(t)t} \frac{s_t^{\alpha_i-1}m_i(t)}{\beta_i(t)} (d_i^+)^{-\alpha_i}
  \left[\left(1-\frac{s_t^{\alpha_\nu-1}m_\nu(t)}{\beta_\nu(t)} \Phi_\nu\right)\left(1-\frac{\beta_l(t)}{\beta_S(t)}\right)\right]^{-1}\\{}&\nonumber \nonumber
e^{-(s_t^{\alpha_l-1}m_l(t) \Phi_lt - \beta_l(t)t) }
dt 
\\{}&\nonumber \nonumber
 =
\sum_{l=0}^N 
\left[\left(1-\frac{\bar{m}_\nu}{\bar{\beta}_\nu} \Phi_\nu\right)\left(1-\frac{\bar{\beta}_l}{\bar{\beta}_S}\right)\right]^{-1}
\frac{\bar{m}_i \Phi_i}{\bar{\beta}_i}(d_i^+)^{-\alpha_i}\frac{\beta_l^2}{\beta_S} \int_0^\infty  t^{(\gamma-1)(\alpha_i-1+\rho(2\alpha_l-\alpha_\nu-\alpha_i))+1}
e^{-t^{(\gamma-1)\rho(\alpha_l-1)+1}\bar{m}_l \Phi_l  }
dt 
\\{}&\nonumber \nonumber
 =
\sum_{l=0}^N 
\left[\left(1-\frac{\bar{m}_\nu}{\bar{\beta}_\nu} \Phi_\nu\right)\left(1-\frac{\bar{\beta}_l}{\bar{\beta}_S}\right)\right]^{-1}
\frac{\bar{m}_i \Phi_i}{\bar{\beta}_i}(d_i^+)^{-\alpha_i}\frac{\beta_l^2(\bar{m}_l\Phi_l)^{-\frac{(\gamma-1)(\alpha_i-1+\rho(2\alpha_l-\alpha_\nu-\alpha_i))+2}{(\gamma-1)\rho(\alpha_l-1)+1}}}{\beta_S((\gamma-1)\rho(\alpha_l-1)+1)}
\\{}&\nonumber \nonumber \Gamma\left(\frac{(\gamma-1)(\alpha_i-1+\rho(2\alpha_l-\alpha_\nu-\alpha_i))+2}{(\gamma-1)\rho(\alpha_l-1)+1}\right)\;\;\;.
\end{align}
Notating: $C_l\triangleq \frac{(\gamma-1)(\alpha_i-1+\rho(2\alpha_l-\alpha_\nu-\alpha_i))+2}{(\gamma-1)\rho(\alpha_l-1)+1}$,$A_{l,\nu}\triangleq \left[\left(1-\frac{\bar{m}_\nu}{\bar{\beta}_\nu} \Phi_\nu\right)\left(1-\frac{\bar{\beta}_l}{\bar{\beta}_S}\right)\right] $
\begin{align}
\MoveEqLeft[3]
\sum_{l=0}^N 
A_{l,\nu}^{-1}
\frac{\bar{m}_i \Phi_i}{\bar{\beta}_i}(d_i^+)^{-\alpha_i}\frac{\beta_l^2(\bar{m}_l\Phi_l)^{-C_l}}{\beta_S((\gamma-1)\rho(\alpha_l-1)+1)}
\Gamma\left(C_l\right)
\end{align}

When $\gamma \rightarrow 1$:
\begin{align}
\MoveEqLeft[3]
\sum_{l=0}^N 
A_{l,\nu}^{-1}
\frac{\bar{m}_i \Phi_i}{\bar{\beta}_i}(d_i^+)^{-\alpha_i}\frac{\beta_l^2}{\beta_S(\bar{m}_l\Phi_l)^2}\;\;\; .
\end{align}

\subsection{Proof of Proposition \ref{prop:prop1} }
\label{sec:proofprop1}
$\forall k \in \mathbb{N}$, let  $S_k\geq 0$, $w\in \mathcal{G}$,  $C_E<1$ , the following event can be defined:
\begin{align}
\MoveEqLeft[3]
    \mathbf{E}_{t,k}^i = \left\{\sup_{t \in [0,S_k]}|\epsilon \xi_{t,k}^i|  < C_E\right\}  \;\;\;.
\end{align}
There exist $\bar{\epsilon}_0$, s.t $\forall \bar{\epsilon} \leq \bar{\epsilon}_0$, the following is true:
\begin{align}
\MoveEqLeft[3]
\left\{\sup_{t \in [0,S_k]}|Z_{t,k}^i(w)-Y_{t,k}^i(w)| \geq c \bar{\epsilon}^{\theta}\right\} = \left\{\sup_{t \in [0,S_k]}|\bar{\epsilon}X_{t,k}^i(w)+R_{t,k}^i(w)| \geq c \bar{\epsilon}^{\theta}\right\} \\{}&\nonumber \nonumber
\subseteq \left\{\sup_{t \in [0,S_k]}|\bar{\epsilon}X_{t,k}^i(w)| \geq \frac{c}{2} \bar{\epsilon}^{\theta}\right\}\cup \left\{|R_{t,k}^i(w)| \geq \frac{c}{2} \bar{\epsilon}^{\theta}\right\}
\\{}&\nonumber \nonumber
\subseteq \left\{\sup_{t \in [0,S_k]}|\bar{\epsilon}\xi_{t,k}^i| \geq \frac{c}{2C_Z} \bar{\epsilon}^{\theta}\right\}\cup \left\{\left\{|R_{t,k}^i(w)| \geq \frac{c}{2} \bar{\epsilon}^{\theta}\right\} \cap \mathbf{E}_{t,k}^i\right\} \cup
\left\{\left\{|R_{t,k}^i(w)| \geq \frac{c}{2} \bar{\epsilon}^{\theta}\right\} \cap \mathbf{E}_{{t,k}^i}^c\right\}
\\{}&\nonumber \nonumber
\subseteq
 \left\{\sup_{t \in [0,S_k]}|\bar{\epsilon}\xi_{t,k}^i| \geq \frac{c}{2C_Z} \bar{\epsilon}^{\theta}\right\}
 \cup
  \left\{\sup_{t \in [0,S_k]}|\bar{\epsilon}\xi_{t,k}^i| \geq \frac{c}{2C_Z\sqrt{C_R}} \bar{\epsilon}^{0.5\theta}\right\}
 \cup \left\{\sup_{t \in [0,S_k]}|\bar{\epsilon}\xi_{t,k}^i| \geq C_E \right\}
 \\{}&\nonumber \nonumber
 \subseteq
 \left\{\sup_{t \in [0,S_k]}|\bar{\epsilon}\xi_{t,k}^i| \geq \frac{c}{2C_Z} \bar{\epsilon}^{\theta}\right\}\;\;\;.
\end{align}
Using Kolmogorov's inequality, for $C_{\theta}>0$:
\begin{align}
\MoveEqLeft[3]
P\left(\sup_{t \in [0,S_k]}|Z_{t,k}^i(w)-Y_{t,k}^i(w)| \geq c \bar{\epsilon}^{\theta}\right)
 \leq
 P\left(\sup_{t \in [0,S_k]}|\bar{\epsilon}\xi_{t,k}^i| \geq \frac{c}{2C_Z} \bar{\epsilon}^{\theta}\right)
 \\{}&\nonumber \nonumber
 \leq
 \frac{4C_Z^2}{c^2\bar{\epsilon}^{2\theta}}\mathop{\mathbb{E}}[\bar{\epsilon}\xi_{t,k}^i]^2
 =\frac{8C_Z^2}{c^2}\bar{\epsilon}^{2-2\theta}\left[\frac{\bar{\epsilon}^{-\rho(1-\alpha_l)}-1}{1-\alpha_l}\right]T\leq
 \frac{8C_Z^2}{c^2}\left[\frac{\bar{\epsilon}^{-\rho(1-\alpha_l)+2-2\theta}}{1-\alpha_l}\right]T \\{}&\nonumber \nonumber = 
 \bar{C}_\theta\bar{\epsilon}^{-\rho(1-\alpha_l)+2-2\theta}T
 \;\;\;.
\end{align}
Final step:
\begin{align}
\MoveEqLeft[3]
P\left(\sup_{t \in [0,T]}|Z_{t,k}^i(w)-Y_{t,k}^i(w)| \geq c \bar{\epsilon}^{\theta}\right) = \int_0^\infty P\left(\sup_{t \in [0,\tau]}|Z_{t,k}^i(w)-Y_{t,k}^i(w)| \geq c \bar{\epsilon}^{\theta}\right)\beta_ie^{-\beta_i\tau}d\tau\\{}&\nonumber \nonumber=\bar{C}_\theta \bar{\epsilon}^{-\rho(1-\alpha_l)+2-2\theta} \int_0^\infty \tau^{1-\rho(1-\alpha_l)+2-2\theta} 
\beta_ie^{-\beta_i\tau}d\tau
\\{}&\nonumber \nonumber= \bar{C}_\theta \bar{\epsilon}^{-\rho(1-\alpha_l)+2-2\theta}\frac{\Gamma(2-\rho(1-\alpha_l)+2-2\theta)}{\beta_i^{2-\rho(1-\alpha_l)+2-2\theta}} = C_\theta \bar{\epsilon}^{-\rho(1-\alpha_l)+2-2\theta} \end{align}\;\;\;. 

\subsection{Proof of Lemma ~\ref{lemma:z_product}}
\label{sec:proof_lemma_z_prod}
In this subsection we will show the full derivation of the approximation of $Z_{t,k}^l$ using stochastic asymptotic expansion, the representation of $Z_t$ in powers of $\bar{\epsilon}=s_t^{\frac{\alpha-1}{\alpha}}\epsilon$:
\begin{equation}
    Z_{t,k}^i=Y_{t,k}^i+\bar{\epsilon}X_{t,k}^i+R_{t,k}^i\;\;\;.
\end{equation}
Where $R_{t,k}^i$ is the error term, we will not discuss this term, for more details see \cite{imkeller2006first}. $X_{t,k}^i$ is the first approximation of $Z_{t,k}^i$ in powers of $\bar{\epsilon}$ and $Y_{t,k}^i$ is the deterministic process. As we show in \ref{sec:theorems}, the relaxation time is much smaller than the interval between the large jumps, hence it's effect on $Z_{t}$ is negligible, thus we will assume: $Z_{t,k} \approx \bar{\epsilon}X_{t,k}$. $X_{t,k}^i$ satisfying the following stochastic differential
equation:
\begin{equation}
    X_{t,k}^i = \int_0^t H(Y_p(w))_{ii}Z_{p,k}^idp+\xi_{p,k}^i\;\;\;.
\end{equation}
The solution to this equation:
\begin{equation}
    X_{t,k}^i = \int_0^t e^{-\int_p^tH(Y_u(w))_{ii}du}d\xi_{p,k}^i\;\;\;.
\end{equation}
Using integration by parts:
\begin{equation}
     X_{t,k}^i = \xi_{t,k}^i-\int_0^t \xi_{p,k}^iH(Y_p(w))_{ii})e^{-\int_p^tH(Y_u(w))_{ii})du}dp
\end{equation}
\begin{align}
\MoveEqLeft[3]
    \mathop{\mathbb{E}}[X_{t,k}^l] =^* \mu_\xi^l t-\int_0^t  \mu_\xi^l tH(Y_p(w)))_{ll}e^{-\int_p^tH(Y_u(w))_{ll})du}dp\\{}&\nonumber \nonumber 
    = \mu_\xi^l t-\int_0^t  \mu_\xi^l ph_{ll}e^{-\int_p^th_{ll}du}dp\\{}&\nonumber \nonumber 
    = \mu_\xi^l t-\int_0^t  \mu_\xi^l ph_{ll}e^{-h_{ll}(t-p)}dp\\{}&\nonumber \nonumber 
    = \mu_\xi^l t- [ \mu_\xi^l h_{ll}[(-\frac{(h_{ll}p+1)}{h_{ll}^2})e^{-h_{ll}(t-p)}]_0^t\\{}&\nonumber \nonumber
    =  \mu_\xi^l t- [ \mu_\xi^l h_{ll}[(-\frac{(h_{ll}t+1)}{h_{ll}^2})+(\frac{1}{h_{ll}^2})e^{-h_{ll}(t)}]\\{}&\nonumber \nonumber=
    \mu_\xi^l t+  \mu_\xi^l \frac{(h_{ll}t+1)}{h_{ll}}-\mu_\xi^l\frac{1}{h_{ll}}e^{-h_{ll}t}
    \\{}&\nonumber \nonumber=
    \mu_\xi^l(2t +  \frac{1}{h_{ll}}-\frac{1}{h_{ll}}e^{-h_{ll}t})\;\;\;.\\{}&\nonumber \nonumber
\end{align}
* using Fubini.\newline
Where $\mu_\xi^l\triangleq \mu_\xi^l(t)$ is the first moment of $\xi_{t,k}^l$:
\begin{align}
\MoveEqLeft[3]
    \mu_\xi^l(t) \triangleq \mathop{\mathbb{E}}[\xi_{t,k}^l]=2t \int_{1}^{\bar{\epsilon}^{-\rho}} \frac{dy}{y^{\alpha_l}}=2t[\frac{1}{1-\alpha_l}y^{1-\alpha_l}]_1^{\bar{\epsilon}^{-\rho}}=2t\left[\frac{\bar{\epsilon}^{-\rho(1-\alpha_l)}-1}{1-\alpha_l}\right]\;\;\;.
\end{align}
We will keep the previous assumptions \cite{imkeller2006levy,imkeller2006first} on the geometry of the potential, that near the basin:$U(w)=h_{ll}\frac{w^2}{2}+o(w^2)$.
Hence we can estimate the expected value of a product of the two processes:
\begin{align}
\MoveEqLeft[3]
    \mathop{\mathbb{E}}[Z_{t,k}^iZ_{t,k}^j] = \mathop{\mathbb{E}}[Y_t^iY_t^j + \bar{\epsilon}_jY_t^iX_{t,k}^j+ \bar{\epsilon}_iY_t^jX_{t,k}^i+\bar{\epsilon}_j\bar{\epsilon}_iX_{t,k}^jX_{t,k}^i]\\{}&\nonumber \nonumber
    \approx^* \mathop{\mathbb{E}}[Y_t^iY_t^j] +  \bar{\epsilon}_jY_t^i\mathop{\mathbb{E}}[X_{t,k}^j]+  \bar{\epsilon}_iY_t^j\mathop{\mathbb{E}}[X_{t,k}^i]    \\{}&\nonumber \nonumber= Y_t^iY_t^j +  \bar{\epsilon}_jY_t^i\mathop{\mathbb{E}}[X_{t,k}^j]+  \bar{\epsilon}_iY_t^j\mathop{\mathbb{E}}[X_{t,k}^i]
    \\{}&\nonumber \nonumber=
    Y_t^iY_t^j +  \bar{\epsilon}_jY_t^i     \mu_\xi^j(2t +  \frac{1}{h_{jj}}-\frac{1}{h_{jj}}e^{-h_{jj}t})+  \bar{\epsilon}_iY_t^j
     \mu_\xi^l(2t +  \frac{1}{h_{ii}}-\frac{1}{h_{ii}}e^{-h_{ii}t})
     \\{}&\nonumber \nonumber\approx
     w_iw_je^{-(h_{ii}+h_{jj})t}+\bar{\epsilon}_jw_ie^{-h_{ii}t}    2t\left[\frac{\bar{\epsilon}^{-\rho(1-\alpha_j)}-1}{1-\alpha_j}\right](2t +  \frac{1}{h_{jj}}-\frac{1}{h_{jj}}e^{-h_{jj}t})\\{}&\nonumber \nonumber
     +\bar{\epsilon}_iw_je^{-h_{jj}t}    2t\left[\frac{\bar{\epsilon}^{-\rho(1-\alpha_i)}-1}{1-\alpha_i}\right](2t +  \frac{1}{h_{ii}}-\frac{1}{h_{ii}}e^{-h_{ii}t})\;\;\;.
\end{align}
*Neglecting terms with order $\bar{\epsilon}^2$.
\subsection{Proof of Proposition \ref{prop:lambda_t_raw_lemma} }
\label{sec:noise_cov_calc}
SGD's covariance:
\begin{equation}
    \Sigma_t = \frac{1}{D} \left[\frac{1}{B}\sum_{i=1}^{Q} \nabla U(W_t)_i \nabla U(W_t)_i^T -\nabla U(W_t) \nabla U(W_t)^T \right]\;\;\;.
\end{equation}
We can approximate the loss landscape near the basin using Taylor expansion:
\begin{align}
\label{eq:taylor_landscape}
\MoveEqLeft[3]
 U(W_t)=U(W^*)+\nabla U(W^*)(W-W^*)+ \frac{1}{2}(W_t-W^*)^T\nabla^2U(W^*)(W_t-W^*)\;\;\;.
\end{align}
Examining SGD's gradient on the $b$-th data point, using the approximation in \ref{eq:taylor_landscape}:
\begin{align}
\MoveEqLeft[3]
\nabla U(W_t)_i\approx\nabla U_d(W^*)+\nabla^2 U_d(W^*)(W_t-W^*)\;\;\;.
\end{align}
The exact gradient (of GD) is:
\begin{align}
\MoveEqLeft[3]
\nabla U(W_t)\approx \nabla^2 U(W^*)(W_t-W^*)\;\;\;.
\end{align}
As a result of empirical evidence in \cite{meng2020dynamic} on the minimum of the covariance curve of SGD, we will drop the first order from the approximation of $\nabla U_d(W)\nabla U_d(W)^T$.
Hence \ref{eq:cov_noise} can be written as:
\begin{align}
\label{eq:cov_noise_new}
\MoveEqLeft[3]
    \lambda(W_t) = \frac{1}{B} \left[\frac{1}{D}\sum_{d=1}^{D} \nabla U_d(W^*)\nabla U_d(W^*)^T +H_d(W^*)W_tW_t^TH_d(W^*)-H(W^*)W_tW_t^TH(W^*) \right]\\{}&\nonumber \nonumber
    \sum_{d=1}^{D}H_d(W^*)W_tW_t^TH_d(W^*) = \frac{1}{D}\sum_{k=1}^N\sum_{p=1}^N\sum_{d=1}^D h_{d,i,k}\tilde{w}_{k,p}h_{d,p,j}
\end{align}
Where $\tilde{w}_{ij}=w_iw_j$
\begin{align}
\MoveEqLeft[3]
    \lambda_{i,j}(t)=\frac{1}{B}\left[\sum_{k=1}^N\sum_{p=1}^N(\frac{1}{D}\sum_{d=1}^D h_{d,i,k}h_{d,p,j}- h_{i,k}h_{p,j})\tilde{w}_{k,p}+\frac{1}{D}\sum_{d=1}^D \nabla  u_{d,i}\nabla u_{d,j}\right]\\{}&\nonumber \nonumber=
    \frac{1}{B}\left[\sum_{k=1}^N\sum_{p=1}^N(\frac{1}{D}\sum_{d=1}^D h_{d,i,k}h_{d,p,j}- h_{i,k}h_{p,j})\tilde{w}_{k,p}+ \tilde{u}_{d,i,j}\right]
\end{align}
$\tilde{u}_{i,j} \triangleq \frac{1}{D}\sum_{d=1}^D \nabla  u_{d,i}\nabla \tilde{u}_{d,j}$, the gradient of all samples in the dataset.
Let us denote: $\bar{h}_{i,k,p,j}\triangleq \frac{1}{D}\sum_{d=1}^D h_{d,i,k}h_{d,p,j}- h_{i,k}h_{p,j}+$
\begin{align}
\MoveEqLeft[3]
    \lambda_{i,j}(t)
    =\frac{1}{B}\left[
        \tilde{u}_{ij}+
    \sum_{k=1}^N\sum_{p=1}^N\bar{h}_{i,k,p,j}W_{t,k}W_{t,p}\right]
    \\{}&\nonumber \nonumber
        \approx \frac{1}{B}\left[
        \tilde{u}_{ij}+
    \sum_{k=1}^N\sum_{p=1}^N\bar{h}_{i,k,p,j}Z_{t,k}Z_{t,p}\right]
    \\{}&\nonumber \nonumber
    \approx \frac{1}{B}
        \tilde{u}_{ij}+
   \sum_{k=1}^N\sum_{p=1}^N\bar{h}_{i,k,p,j} \mathop{\mathbb{E}}[Z_{t,k}Z_{t,p}]
\\{}&\nonumber \nonumber \\{}&\nonumber \nonumber
\lambda_{i,j}(t)\approx 
\frac{1}{BD}
        \tilde{u}_{ij}+\\{}&\nonumber
         \frac{1}{B}\left[
   \sum_{k=1}^N\sum_{p=1}^N\bar{h}_{i,k,p,j}(
     w_kw_pe^{-(h_{kk}+h_{pp})t}+\bar{\epsilon}_pw_ke^{-h_{kk}t} \mu_\xi^p(2t +  \frac{1}{h_{pp}}(1-e^{-h_{pp}t}))
     +\bar{\epsilon}_kw_pe^{-h_{pp}t}    \mu_\xi^k(2t +  \frac{1}{h_{kk}}(1-e^{-h_{kk}t})))\right]
 \end{align}

\subsection{Proof of Lemma~\ref{sec:lemma1}}
\label{sec:det_conv_proof}

We will denote $W^*$ as the optimal point in the basin,
using the differential form, it is known that:
\begin{align}
\MoveEqLeft[3]
\frac{dY_t}{dt}=-\nabla U(Y_t)\;\;\;.
\end{align}
Let us denote:$\zeta(t) = U(Y_t) - U(W^*)$, directly from that notation:
\begin{align}
\MoveEqLeft[3]
d\zeta(t) =  \langle \nabla U(Y_t),dY_t \rangle=-
 \norm{\nabla U(Y_t)}^2 \;\;\;.
\end{align}
Since $U(Y_t)$ is $\mu-$strongly convex near the basin $W^*$:
\begin{align}
\MoveEqLeft[3]
U(Y_t) - U(W^*) \leq \frac{1}{2\mu}\norm{\nabla U(Y_t)}^2
\\{}&\nonumber \nonumber
 -2\mu\zeta(t) \geq d\zeta(t)\;\;\;.
\end{align}
Using Gronwall's lemma \cite{gronwall1919note}::
\begin{align}
\MoveEqLeft[3]
U(Y_t)-U(W^*)\leq(U(w)-U(W^*))e^{-2\mu t}\;\;\;.
\end{align}
Directly from strong convex propriety $U(Y_t)-U(W^*)\geq \frac{\mu}{2}\norm{Y_t-W^*}^2$ , we can achieve:
\begin{align}
\MoveEqLeft[3]
\norm{Y_t-W^*}^2\leq \frac{2(U(w)-U(W^*))}{\mu}e^{-2\mu t}=\frac{2\zeta(t)}{\mu}e^{-2\mu t}\;\;\;.
\end{align}
\newpage
\section{Additional Theorems}
\everypar{\looseness=-1}
\setlength{\abovedisplayskip}{1pt}
\setlength{\belowdisplayskip}{1pt}

\subsection{Mean Escape Time-Multistep scheduler}
Next theorem deals with the mean transition time of a popular scheduler, the multi-step scheduler.
Before stating the theorem, let us define few constants first,  $\nu_{nn} \triangleq \frac{\gamma_p^{\alpha_\nu-1}\bar{m}_\nu}{\bar{\beta}_\nu}\Phi_\nu$ this term express the global attributes of the DNN. Next constant
 $\tilde{C}_{l,\nu,p}=\frac{\frac{\bar{\beta}_l}{\bar{\beta}_S}\gamma_p^{\rho(\alpha_l-\alpha_\nu)}(\gamma_p^{\alpha_l-1}\bar{m}_l \Phi_l - \gamma_p^{\rho(\alpha_l-1)}\bar{\beta}_l)}{\left[\nu_{nn} + \frac{\bar{\beta}_l}{\bar{\beta}_S}\gamma_p^{\rho(\alpha_l-\alpha_\nu)}(1 -\nu_{nn})\right]}$
 utters global and single parameters attributes
 , last: $\bar{C}_{l,p}\triangleq \bar{\beta}_l\gamma_p^{\rho(\alpha_l-1)}$. 
 \begin{theorem}
 Let $s_t$ be a multi-step scheduler . Further, let us notate  $C_{l,\nu,p}\triangleq \bar{C}_{l,\nu,p} +\tilde{C}_{l,\nu,p}$, and $E_p \triangleq e^{-C_{l,\nu,p}T_p}T_p$. The mean transition time with a multi-step scheduler satisfies:
\begin{align}
\MoveEqLeft[1]
\mathop{\mathbb{E}}[\sigma_{\mathcal{G}}]
\approx \nonumber
\sum_{l=0}^N \sum_{p=0}^{P}\frac{\bar{\beta_l}}{\bar{\beta}_SC_{l,\nu,p}}  \gamma_p^{\alpha_l(1-\rho)-1+\rho\alpha_\nu}\bar{m}_l \Phi_l
A_{l,\nu}^{-1}
(E_p-E_{p+1})\;\;\;.
\end{align}
\end{theorem}
Proof:
\begin{align}
\MoveEqLeft[3]
\mathop{\mathbb{E}}[\sigma_{\mathcal{G}}]\leq \sum_{k=1}^{\infty}\mathop{\mathbb{E}}[\tau_k^*\mathbbm{I}\{\sigma_{\mathcal{G}}=\tau_k^*\}]\\{}&\nonumber \nonumber
=\sum_{k=1}^{\infty}\mathop{\mathbb{E}}[\tau_k^*\mathbbm{I}\{\sum_{l=0}^Ns_t\epsilon\mathbf{1}^T\lambda_l(t)J_1^l\mathbbm{I}\{\tau_1^l=\tau_1^*\} \in \mathcal{G},\sum_{l=0}^Ns_t\epsilon\mathbf{1}^T\lambda_l(t)J_2^l\mathbbm{I}\{\tau_2^l=\tau_2^*\} \in \mathcal{G},..,\sum_{l=0}^Ns_t\epsilon\mathbf{1}^T\lambda_l(t)J_k^l\mathbbm{I}\{\tau_k^l=\tau_k^*\} \notin \mathcal{G}\}]
\\{}&\nonumber \nonumber
=
\sum_{k=1}^{\infty}\mathop{\mathbb{E}}[\tau_k^*\mathbbm{I}\{J_1^* \in \mathcal{G},J_2^* \in \mathcal{G},..,J_k^* \notin \mathcal{G}\}]\\{}&\nonumber \nonumber
\leq
\sum_{k=1}^{\infty}\mathop{\mathbb{E}}[\tau_k^*(1- \mathbbm{I}\{J^{*}_k \notin \mathcal{G}\})^{k-1} \mathbbm{I}\{J^{*}_k \notin \mathcal{G}\}]
  \\{}&\nonumber \nonumber=
  \sum_{k=1}^{\infty} \sum_{l=1}^{N}\mathop{\mathbb{E}}[\tau_k^l(1- \mathbbm{I}\{J^{l}_k \notin \mathcal{G}\})^{k-1} \mathbbm{I}\{J^{l}_k \notin \mathcal{G}\}\mathbbm{I}\{\tau_k^l=\tau_k^*\}]
  \\{}&\nonumber \nonumber\leq
 \sum_{k=1}^{\infty} \sum_{l=1}^{N}\sum_{w=1}^{k}\mathop{\mathbb{E}}[\tau_w^l(1- \mathbbm{I}\{s_t^{\frac{\alpha_l-1}{\alpha_l}} \epsilon\mathbf{1}^T\lambda_l(t)J^{l}_w \notin \mathcal{G}\})^{w-1}(1- \mathbbm{I}\{s_t^{\frac{\alpha_\nu-1}{\alpha_\nu}} \epsilon\mathbf{1}^T\lambda_\nu(t)J^{m}_w \notin \mathcal{G}\})^{k-w}
 \\{}&\nonumber \nonumber
 \mathbbm{I}\{s_t^{\frac{\alpha_l-1}{\alpha_l}} \epsilon\mathbf{1}^T\lambda_i(t)J^{l}_w \notin \mathcal{G}\}\mathbbm{I}\{\tau_{w}^l=\tau_k^*\}]
\\{}&\nonumber \nonumber=
\sum_{k=1}^\infty  \sum_{l=0}^N \sum_{w=1}^{k} \int_0^\infty \frac{\beta_l(t)}{\beta_S(t)}\frac{(k-1)!}{(w-1)!(k-w)!}\left(\frac{\beta_l(t)}{\beta_S(t)}\right)^{w-1}\left(1-\frac{\beta_l(t)}{\beta_S(t)}\right)^{k-w}\beta_l(t)t\\{}&\nonumber \nonumber e^{-\beta_l(t)t}\frac{(\beta_l(t)t)^{w-1}}{(w-1)!} [1-\frac{s_t^{\alpha_l-1}m_l(t)}{\beta_l(t)} \Phi_l]^{w-1}
[1-\frac{s_t^{\alpha_\nu-1}m_\nu(t)}{\beta_\nu(t)} \Phi_\nu]^{k-w}\frac{s_t^{\alpha_l-1}m_l(t)}{\beta_l(t)} \Phi_ldt
\\{}&\nonumber \nonumber= 
\sum_{k=1}^\infty  \sum_{l=0}^N \int_0^\infty \frac{\beta_l(t)}{\beta_S(t)}t e^{-\beta_l(t)t} s_t^{\alpha_l-1}m_l(t) \Phi_l
\\{}&\nonumber \nonumber
\sum_{w=1}^{k}\frac{[\beta_l(t)t-s_t^{\alpha_l-1}m_l(t) \Phi_lt]^{w-1}}{(w-1)!}\frac{(k-1)!}{(w-1)!(k-w)!}\left(\frac{\beta_l(t)}{\beta_S(t)}\right)^{w-1}\left[\left(1-\frac{s_t^{\alpha_\nu-1}m_\nu(t)}{\beta_\nu(t)} \Phi_\nu\right)\left(1-\frac{\beta_l(t)}{\beta_S(t)}\right)\right]^{k-w}dt
\\{}&\nonumber \nonumber=
\sum_{k=1}^\infty  \sum_{l=0}^N \int_0^\infty \frac{\beta_l(t)}{\beta_S(t)}t e^{-\beta_l(t)t} s_t^{\alpha_l-1}m_l(t) \Phi_l
\\{}&\nonumber \nonumber
\left[\left(1-\frac{s_t^{\alpha_\nu-1}m_\nu(t)}{\beta_\nu(t)} \Phi_\nu\right)\left(1-\frac{\beta_l(t)}{\beta_S(t)}\right)\right]^{k-1}L_{k-1}\left(\frac{\frac{\beta_l(t)}{\beta_S(t)}(s_t^{\alpha_l-1}m_l(t) \Phi_lt-\beta_l(t)t)}{\left[\left(1-\frac{s_t^{\alpha_\nu-1}m_\nu(t)}{\beta_\nu(t)} \Phi_\nu\right)\left(1-\frac{\beta_l(t)}{\beta_S(t)}\right)\right]}\right)
\end{align}

\begin{align}
\MoveEqLeft[1]
\sum_{k=1}^\infty  \sum_{l=0}^N \int_0^\infty \frac{\beta_l(t)}{\beta_S(t)}t e^{-\beta_l(t)t} s_t^{\alpha_l-1}m_l(t) \Phi_l
\\{}&\nonumber \nonumber
\left[\left(1-\frac{s_t^{\alpha_\nu-1}m_\nu(t)}{\beta_\nu(t)} \Phi_\nu\right)\left(1-\frac{\beta_l(t)}{\beta_S(t)}\right)\right]^{k-1}L_{k-1}\left(\frac{\frac{\beta_l(t)}{\beta_S(t)}(s_t^{\alpha_l-1}m_l(t) \Phi_lt-\beta_l(t)t)}{\left[\left(1-\frac{s_t^{\alpha_\nu-1}m_\nu(t)}{\beta_\nu(t)} \Phi_\nu\right)\left(1-\frac{\beta_l(t)}{\beta_S(t)}\right)\right]}\right)
\\{}&\nonumber \nonumber=
 \sum_{l=0}^N \int_0^\infty \frac{\beta_l(t)}{\beta_S(t)}t e^{-\beta_l(t)t} s_t^{\alpha_l-1}m_l(t) \Phi_l
 \\{}&\nonumber \nonumber
\left[\left(1-\frac{s_t^{\alpha_\nu-1}m_\nu(t)}{\beta_\nu(t)} \Phi_\nu\right)\left(1-\frac{\beta_l(t)}{\beta_S(t)}\right)\right]^{-1}
e^{-\frac{\frac{\beta_l(t)}{\beta_S(t)}(s_t^{\alpha_l-1}m_l(t) \Phi_lt - \beta_l(t)t)}{\left[1-\left(1-\frac{s_t^{\alpha_\nu-1}m_\nu(t)}{\beta_\nu(t)} \Phi_\nu\right)\left(1-\frac{\beta_l(t)}{\beta_S(t)}\right)\right]}}
dt 
\\{}&\nonumber \nonumber=
\sum_{l=0}^N \int_0^\infty \frac{\beta_l(t)}{\beta_S(t)}t e^{-\beta_l(t)t} s_t^{\alpha_l-1}m_l(t) \Phi_l
\\{}&\nonumber \nonumber
\left[\left(1-\frac{s_t^{\alpha_\nu-1}m_\nu(t)}{\beta_\nu(t)} \Phi_\nu\right)\left(1-\frac{\beta_l(t)}{\beta_S(t)}\right)\right]^{-1}
e^{-\frac{\frac{\beta_l(t)}{\beta_S(t)}(s_t^{\alpha_l-1}m_l(t) \Phi_lt - \beta_l(t)t)}{\left[\frac{s_t^{\alpha_\nu-1}m_\nu(t)}{\beta_\nu(t)}\Phi_\nu + \frac{\beta_l(t)}{\beta_S(t)} -\frac{s_t^{\alpha_\nu-1}m_\nu(t)\Phi_\nu}{\beta_\nu(t)}\frac{\beta_l(t)}{\beta_S(t)}\right]}}
dt 
\end{align}

\begin{align}
\MoveEqLeft[1]
\sum_{l=0}^N \int_0^\infty \frac{\beta_l(t)}{\beta_S(t)}t e^{-\beta_l(t)t} s_t^{\alpha_l-1}m_l(t) \Phi_l
\\{}&\nonumber \nonumber
\left[\left(1-\frac{s_t^{\alpha_\nu-1}m_\nu(t)}{\beta_\nu(t)} \Phi_\nu\right)\left(1-\frac{\beta_l(t)}{\beta_S(t)}\right)\right]^{-1}
e^{-\frac{\frac{\beta_l(t)}{\beta_S(t)}(s_t^{\alpha_l-1}m_l(t) \Phi_lt - \beta_l(t)t)}{\left[\frac{s_t^{\alpha_\nu-1}m_\nu(t)}{\beta_\nu(t)}\Phi_\nu + \frac{\beta_l(t)}{\beta_S(t)} -\frac{s_t^{\alpha_\nu-1}m_\nu(t)\Phi_\nu}{\beta_\nu(t)}\frac{\beta_l(t)}{\beta_S(t)}\right]}}
dt 
\\{}&\nonumber \nonumber=
\sum_{l=0}^N \sum_{p=0}^{P}\int_{T_p}^{T_{p+1}}\frac{\bar{\beta_l}}{\bar{\beta}_S}  \gamma_p^{(\alpha_l-1+\rho(\alpha_l-\alpha_\nu)}\bar{m}_l \Phi_l
\left[\left(1-\frac{\bar{m}_\nu}{\bar{\beta}_\nu} \Phi_\nu\right)\left(1-\frac{\bar{\beta}_l}{\bar{\beta}_S}\right)\right]^{-1}t
\\{}&\nonumber \nonumber
e^{\left(-\frac{\frac{\bar{\beta}_l}{\bar{\beta}_S}\gamma_p^{\rho(\alpha_l-\alpha_\nu)}(\gamma_p^{\alpha_l-1}\bar{m}_l \Phi_l - \gamma_p^{\rho(\alpha_l-1)}\bar{\beta}_l)}{\left[\frac{\gamma_p^{\alpha_\nu-1}\bar{m}_\nu}{\bar{\beta}_\nu}\Phi_\nu + \frac{\bar{\beta}_l}{\bar{\beta}_S}\gamma_p^{\rho(\alpha_l-\alpha_\nu)} -\frac{\gamma_p^{\alpha_\nu-1}\bar{m}_\nu\Phi_\nu}{\beta_\nu(t)}\frac{\bar{\beta}_l}{\bar{\beta}_S}\gamma_p^{\rho(\alpha_l-\alpha_\nu)}\right]}-\bar{\beta}_l\gamma_p^{\rho(\alpha_l-1)}\right)t}
dt 
\end{align}
Let us notate:  $C_{l,\nu,p}\triangleq \frac{\frac{\bar{\beta}_l}{\bar{\beta}_S}\gamma_p^{\rho(\alpha_l-\alpha_\nu)}(\gamma_p^{\alpha_l-1}\bar{m}_l \Phi_l - \gamma_p^{\rho(\alpha_l-1)}\bar{\beta}_l)}{\left[\frac{\gamma_p^{\alpha_\nu-1}\bar{m}_\nu}{\bar{\beta}_nu}\Phi_\nu + \frac{\bar{\beta}_l}{\bar{\beta}_S}\gamma_p^{\rho(\alpha_l-\alpha_\nu)} +\frac{\gamma_p^{\alpha_\nu-1}\bar{m}_\nu\Phi_\nu}{\beta_\nu(t)}\frac{\bar{\beta}_l}{\bar{\beta}_S}\gamma_p^{\rho(\alpha_l-\alpha_\nu)}\right]}+\bar{\beta}_l\gamma_p^{\rho(\alpha_l-1)}$,$A_{l,\nu}\triangleq\left[\left(1-\frac{\bar{m}_\nu}{\bar{\beta}_\nu} \Phi_\nu\right)\left(1-\frac{\bar{\beta}_l}{\bar{\beta}_S}\right)\right]$
\begin{align}
\MoveEqLeft[3]
\sum_{l=0}^N \sum_{p=0}^{P}\int_{T_p}^{T_{p+1}}\frac{\bar{\beta_l}}{\bar{\beta}_S}  \gamma_p^{(\alpha_l-1+\rho(\alpha_l-\alpha_\nu)}\bar{m}_l \Phi_l
A_{l,\nu}^{-1}t
e^{-C_{l,\nu,p}t}
\\{}&\nonumber \nonumber=
\sum_{l=0}^N \sum_{p=0}^{P}\frac{\bar{\beta_l}}{\bar{\beta}_SC_{l,\nu,p}^2}  \gamma_p^{\alpha_l-1+\rho(\alpha_l-\alpha_\nu)}\bar{m}_l \Phi_l
A_{l,\nu}^{-1}  
(e^{-C_{l,\nu,p}T_p}(C_{l,\nu,p}T_p+1)-e^{-C_{l,\nu,p}T_{p+1}}(C_{l,\nu,p}T_{p+1}+1))
\\{}&\nonumber \nonumber\approx
\sum_{l=0}^N \sum_{p=0}^{P}\frac{\bar{\beta_l}}{\bar{\beta}_SC_{l,\nu,p}}  \gamma_p^{\alpha_l-1+\rho(\alpha_l-\alpha_\nu)}\bar{m}_l \Phi_l
A_{l,\nu}^{-1}
(e^{-C_{l,\nu,p}T_p}T_p-e^{-C_{l,\nu,p}T_{p+1}}T_{p+1})
\end{align}
\clearpage
\subsection{Trapping probability}
\begin{theorem}
Let $s_t$ be an exponential scheduler $s_t=t^{\gamma-1}$,  $\gamma$ is the cooling rate.
The probability of the  process to be trapped in the domain $\mathcal{G}$ is upper bounded by:
\begin{align}
    P(\sigma<\infty) \leq  \sum_{l=0}^N\frac{\bar{m}_l \Phi_l}{\bar{\beta}_S} 
\left[\frac{\bar{\beta}_l}{\bar{\beta}_S}-\frac{\bar{\beta}_l}{\bar{\beta}_S}\frac{\bar{m}_\nu}{\bar{\beta}_\nu} \Phi_\nu\right]^{-1}
  \\{}&\nonumber \nonumber 
  \frac{(2\beta_l-m_l \Phi_l)^{\frac{(\gamma-1)(\alpha_l-1+\rho(\alpha_{\nu}-1))+1}{(\gamma-1)(\rho(\alpha_l-1))+1}}}{{(\gamma-1)(\rho(\alpha_l-1))+1}}
\Gamma({\frac{(\gamma-1)(\alpha_l-1+\rho(\alpha_{\nu}-1))+1}{(\gamma-1)(\rho(\alpha_l-1))+1}})
\end{align}
\end{theorem}
Proof:
\begin{align}
\MoveEqLeft[3]  
P(\sigma<\infty) = \sum_{k=0}^{\infty} P(\sigma = \tau_k^*) \leq
\sum_{k=0}^{\infty} P(J^{*}_1 \in \mathcal{G},J^{*}_2 \in \mathcal{G},..,J^{*}_k \notin \mathcal{G}) \\{}&\nonumber \nonumber  = 
\sum_{k=1}^{\infty}\prod_{j=1}^{k-1}P(J_j^*\in \mathcal{G})P(J^{*}_k \notin \mathcal{G})
=
\sum_{k=1}^{\infty}(1-P(J_k^*\notin \mathcal{G}))^{k-1}P(J^{*}_k \notin \mathcal{G})
\\{}&\nonumber \nonumber=
\sum_{k=1}^{\infty}\sum_{l=1}^{N}
\sum_{w=1}^{k-1}(1-P(s_t^{\frac{\alpha_l-1}{\alpha_l}}\epsilon\mathbf{1}^t\lambda_l(t)J_w^l\notin \mathcal{G}))^{w-1}(1-P(s_t^{\frac{\alpha_\nu-1}{\alpha_\nu}}\epsilon\mathbf{1}^t\lambda_\nu(t)J_w^v\notin \mathcal{G}))^{k-w}
\\{}&\nonumber \nonumber
P(s_t^{\frac{\alpha_l-1}{\alpha_l}}\epsilon\mathbf{1}^t\lambda_l(t)J_w^l\notin \mathcal{G})P(\tau_w^l=\tau_k^*)
\\{}&\nonumber \nonumber 
=
\sum_{k=1}^\infty  \sum_{l=0}^N \sum_{w=1}^{k} \int_0^\infty \frac{\beta_l(t)}{\beta_S(t)}\frac{(k-1)!}{(w-1)!(k-w)!}\left(\frac{\beta_l(t)}{\beta_S(t)}\right)^{w-1}\left(1-\frac{\beta_l(t)}{\beta_S(t)}\right)^{k-w}\\{}&\nonumber \nonumber e^{-\beta_l(t)t}\frac{(\beta_l(t)t)^{w-1}}{(w-1)!} [1-\frac{s_t^{\alpha_l-1}m_l(t)}{\beta_l(t)} \Phi_l]^{w-1}
[1-\frac{s_t^{\alpha_\nu-1}m_\nu(t)}{\beta_\nu(t)} \Phi_\nu]^{k-w}s_t^{\alpha_l-1}m_l(t) \Phi_ldt
\\{}&\nonumber \nonumber= 
\sum_{k=1}^\infty  \sum_{l=0}^N \int_0^\infty \frac{\beta_l(t)}{\beta_S(t)} e^{-\beta_l(t)t} s_t^{\alpha_l-1}m_l(t) \Phi_l
\\{}&\nonumber \nonumber
\sum_{w=1}^{k}\frac{[\beta_l(t)t-s_t^{\alpha_l-1}m_l(t) \Phi_lt]^{w-1}}{(w-1)!}\frac{(k-1)!}{(w-1)!(k-w)!}\left(\frac{\beta_l(t)}{\beta_S(t)}\right)^{w-1}\left[\left(1-\frac{s_t^{\alpha_\nu-1}m_\nu(t)}{\beta_\nu(t)} \Phi_\nu\right)\left(1-\frac{\beta_l(t)}{\beta_S(t)}\right)\right]^{k-w}dt
\\{}&\nonumber \nonumber=
\sum_{k=1}^\infty  \sum_{l=0}^N \int_0^\infty \frac{m_l(t) \Phi_l}{\beta_S(t)} e^{-\beta_l(t)t} s_t^{\alpha_l-1}
\\{}&\nonumber \nonumber
\left[\left(1-\frac{s_t^{\alpha_\nu-1}m_\nu(t)}{\beta_\nu(t)} \Phi_\nu\right)\left(1-\frac{\beta_l(t)}{\beta_S(t)}\right)\right]^{k-1}L_{k-1}\left(\frac{\frac{\beta_l(t)}{\beta_S(t)}(s_t^{\alpha_l-1}m_l(t) \Phi_lt-\beta_l(t)t)}{\left[\left(1-\frac{s_t^{\alpha_\nu-1}m_\nu(t)}{\beta_\nu(t)} \Phi_\nu\right)\left(1-\frac{\beta_l(t)}{\beta_S(t)}\right)\right]}\right)
\\{}&\nonumber \nonumber=
 \sum_{l=0}^N \int_0^\infty \frac{m_l(t) \Phi_l}{\beta_S(t)} e^{-\beta_l(t)t} s_t^{\alpha_l-1}
 \\{}&\nonumber \nonumber
\left[1-\left(1-\frac{s_t^{\alpha_\nu-1}m_\nu(t)}{\beta_\nu(t)} \Phi_\nu\right)\left(1-\frac{\beta_l(t)}{\beta_S(t)}\right)\right]^{-1}
e^{-\frac{\left[\left(1-\frac{s_t^{\alpha_\nu-1}m_\nu(t)}{\beta_\nu(t)} \Phi_\nu\right)\left(1-\frac{\beta_l(t)}{\beta_S(t)}\right)\right]\frac{\beta_l(t)}{\beta_S(t)}(\beta_l(t)t-s_t^{\alpha_l-1}m_l(t) \Phi_lt)}{\left[1-\left(1-\frac{s_t^{\alpha_\nu-1}m_\nu(t)}{\beta_\nu(t)} \Phi_\nu\right)\left(1-\frac{\beta_l(t)}{\beta_S(t)}\right)\right]}}
dt 
\\{}&\nonumber \nonumber=
 \sum_{l=0}^N \int_0^\infty \frac{m_l(t) \Phi_l}{\beta_S(t)} e^{-\beta_l(t)t} s_t^{\alpha_l-1}
 \left[1-\left(1-\frac{s_t^{\alpha_\nu-1}m_\nu(t)}{\beta_\nu(t)} \Phi_\nu\right)\left(1-\frac{\beta_l(t)}{\beta_S(t)}\right)\right]^{-1}
 \\{}&\nonumber \nonumber
e^{-\frac{\frac{\beta_l(t)}{\beta_S(t)}(\beta_l(t)t -s_t^{\alpha_l-1}m_l(t) \Phi_lt)}{\left[\frac{s_t^{\alpha_\nu-1}m_\nu(t)}{\beta_\nu(t)} \Phi_\nu+\frac{\beta_l(t)}{\beta_S(t)}-\frac{s_t^{\alpha_\nu-1}m_\nu(t)}{\beta_\nu(t)} \Phi_\nu\frac{\beta_l(t)}{\beta_S(t)}\right]}-
\frac{\beta_l(t)}{\beta_S(t)}(\beta_l(t)t-s_t^{\alpha_l-1}m_l(t) \Phi_lt)}
dt \\{}&\nonumber \nonumber=
 \sum_{l=0}^N \int_0^\infty \frac{m_l(t) \Phi_l}{\beta_S(t)} e^{-\beta_l(t)t} s_t^{\alpha_l-1}
 \left[1-\left(1-\frac{s_t^{\alpha_\nu-1}m_\nu(t)}{\beta_\nu(t)} \Phi_\nu\right)\left(1-\frac{\beta_l(t)}{\beta_S(t)}\right)\right]^{-1}
 \\{}&\nonumber \nonumber 
 e^{-
\frac{\beta_l(t)}{\beta_S(t)}(\beta_l(t)t-s_t^{\alpha_l-1}m_l(t) \Phi_lt)\left[\frac{1}{\frac{s_t^{\alpha_\nu-1}m_\nu(t)}{\beta_\nu(t)} \Phi_\nu+\frac{\beta_l(t)}{\beta_S(t)}-\frac{s_t^{\alpha_\nu-1}m_\nu(t)}{\beta_\nu(t)} \Phi_\nu\frac{\beta_l(t)}{\beta_S(t)}}-1\right]}
dt
\end{align}
\begin{align}
\MoveEqLeft[3]  
 \sum_{l=0}^N \int_0^\infty \frac{m_l(t) \Phi_l}{\beta_S(t)} e^{-\beta_l(t)t} s_t^{\alpha_l-1}
 \left[1-\left(1-\frac{s_t^{\alpha_\nu-1}m_\nu(t)}{\beta_\nu(t)} \Phi_\nu\right)\left(1-\frac{\beta_l(t)}{\beta_S(t)}\right)\right]^{-1}
 \\{}&\nonumber \nonumber 
 e^{-
\frac{\beta_l(t)}{\beta_S(t)}(\beta_l(t)t-s_t^{\alpha_l-1}m_l(t) \Phi_lt)\left[\frac{1}{\frac{s_t^{\alpha_\nu-1}m_\nu(t)}{\beta_\nu(t)} \Phi_\nu+\frac{\beta_l(t)}{\beta_S(t)}-\frac{s_t^{\alpha_\nu-1}m_\nu(t)}{\beta_\nu(t)} \Phi_\nu\frac{\beta_l(t)}{\beta_S(t)}}-1\right]}
\\{}&\nonumber \nonumber\leq
 \sum_{l=0}^N\int_0^\infty  \frac{\bar{m}_l \Phi_l}{\bar{\beta}_S}  s_t^{\alpha_l-1+\rho(\alpha_{\nu}-1)}
 \\{}&\nonumber \nonumber 
\left[\frac{\bar{\beta}_l}{\bar{\beta}_S}-\frac{\bar{\beta}_l}{\bar{\beta}_S}\frac{\bar{m}_\nu}{\bar{\beta}_\nu} \Phi_\nu\right]^{-1}
e^{-(2\beta_l-m_l \Phi_l)s_t^{\rho(\alpha_l-1)}t}
dt
\\{}&\nonumber \nonumber = 
\frac{\bar{m}_l \Phi_l}{\bar{\beta}_S} \left[\frac{\bar{\beta}_l}{\bar{\beta}_S}-\frac{\bar{\beta}_l}{\bar{\beta}_S}\frac{\bar{m}_\nu}{\bar{\beta}_\nu} \Phi_\nu\right]^{-1}
  \\{}&\nonumber \nonumber 
  \int_0^\infty   
  t^{(\gamma-1)(\alpha_l-1+\rho(\alpha_{\nu}-1))}
e^{-(2\beta_l-m_l \Phi_l)t^{(\gamma-1)(\rho(\alpha_l-1))+1}}
dt\\{}&\nonumber \nonumber=
  \sum_{l=0}^N\frac{\bar{m}_l \Phi_l}{\bar{\beta}_S} 
\left[\frac{\bar{\beta}_l}{\bar{\beta}_S}-\frac{\bar{\beta}_l}{\bar{\beta}_S}\frac{\bar{m}_\nu}{\bar{\beta}_\nu} \Phi_\nu\right]^{-1}
  \\{}&\nonumber \nonumber 
  \frac{(2\beta_l-m_l \Phi_l)^{\frac{(\gamma-1)(\alpha_l-1+\rho(\alpha_{\nu}-1))+1}{(\gamma-1)(\rho(\alpha_l-1))+1}}}{{(\gamma-1)(\rho(\alpha_l-1))+1}}
\Gamma({\frac{(\gamma-1)(\alpha_l-1+\rho(\alpha_{\nu}-1))+1}{(\gamma-1)(\rho(\alpha_l-1))+1}}) \;\;\;.
\end{align}
\subsection{Trapping probability - Multi step scheduler}
\begin{theorem}
Let $s_t$ be an multistep scheduler,  $\gamma$ is the cooling rate. 
The probability of the training process to be trapped in the domain $\mathcal{G}$ is upper bounded by:
\begin{equation}
    P(\sigma<\infty) \leq  \sum_{l=0}^N \sum_{p=0}^{P}\left[\frac{\bar{\beta}_l}{\bar{\beta}_S}(1-\frac{\bar{m}_\nu}{\bar{\beta}_\nu} \Phi_\nu)\right]^{-1} \frac{\bar{m}_l \Phi_l}{\bar{\beta}_S}\gamma_p^{\rho(1-\alpha_{\nu})+\alpha_l-1-\rho(\alpha_l-1)} 
\frac{e^{-(2\bar{\beta}_l-\bar{m}_l \Phi_l)\gamma_p^{\rho(\alpha_l-1)}Tp}(1-e^{-(2\bar{\beta}_l-\bar{m}_l \Phi_l)\gamma_p^{\rho(\alpha_l-1)T}})}{(2\bar{\beta}_l-\bar{m}_l \Phi_l)} \;\;\;.
\end{equation}
\end{theorem}
Proof:
\begin{align}
\MoveEqLeft[3]
P(\sigma<\infty) \approx \sum_{k=0}^{\infty} P(\sigma = \tau_k^*) \leq
\sum_{k=0}^{\infty} P(J^{*}_1 \in \mathcal{G},J^{*}_2 \in \mathcal{G},..,J^{*}_k \notin \mathcal{G}) \\{}&\nonumber \nonumber  = \sum_{k=1}^{\infty}\prod_{j=1}^{k-1}P(J_j^*\in \mathcal{G})P(J_k^*\in
\Omega_i^+)
=
\sum_{k=1}^{\infty}(1-P(J_k^*\notin \mathcal{G}))^{k-1}P(J^{*}_k \notin \mathcal{G})
\\{}&\nonumber \nonumber=
\sum_{k=1}^{\infty}\sum_{l=1}^{N}
\sum_{w=1}^{k-1}(1-P(s_t^{\frac{\alpha_l-1}{\alpha_l}}\epsilon\mathbf{1}^t\lambda_l(t)J_w^l\notin \mathcal{G}))^{w-1}(1-P(s_t^{\frac{\alpha_\nu-1}{\alpha_\nu}}\epsilon\mathbf{1}^t\lambda_\nu(t)J_w^v\notin \mathcal{G}))^{k-w}
\\{}&\nonumber \nonumber
P(s_t^{\frac{\alpha_l-1}{\alpha_l}}\epsilon\mathbf{1}^t\lambda_l(t)J_w^l\notin \mathcal{G})P(\tau_w^l=\tau_k^*)
\\{}&\nonumber \nonumber 
=
\sum_{k=1}^\infty  \sum_{l=0}^N \sum_{w=1}^{k} \int_0^\infty \frac{\beta_l(t)}{\beta_S(t)}\frac{(k-1)!}{(w-1)!(k-w)!}\left(\frac{\beta_l(t)}{\beta_S(t)}\right)^{w-1}\left(1-\frac{\beta_l(t)}{\beta_S(t)}\right)^{k-w}\\{}&\nonumber \nonumber e^{-\beta_l(t)t}\frac{(\beta_l(t)t)^{w-1}}{(w-1)!} [1-\frac{s_t^{\alpha_l-1}m_l(t)}{\beta_l(t)} \Phi_l]^{w-1}
[1-\frac{s_t^{\alpha_\nu-1}m_\nu(t)}{\beta_\nu(t)} \Phi_\nu]^{k-w}\frac{s_t^{\alpha_l-1}m_l(t)}{\beta_l(t)} \Phi_ldt
\\{}&\nonumber \nonumber= 
\sum_{k=1}^\infty  \sum_{l=0}^N \int_0^\infty \frac{\beta_l(t)}{\beta_S(t)} e^{-\beta_l(t)t} s_t^{\alpha_l-1}m_l(t) \Phi_l
\\{}&\nonumber \nonumber
\sum_{w=1}^{k}\frac{[\beta_l(t)t-s_t^{\alpha_l-1}m_l(t) \Phi_lt]^{w-1}}{(w-1)!}\frac{(k-1)!}{(w-1)!(k-w)!}\left(\frac{\beta_l(t)}{\beta_S(t)}\right)^{w-1}\left[\left(1-\frac{s_t^{\alpha_\nu-1}m_\nu(t)}{\beta_\nu(t)} \Phi_\nu\right)\left(1-\frac{\beta_l(t)}{\beta_S(t)}\right)\right]^{k-w}dt
\\{}&\nonumber \nonumber=
\sum_{k=1}^\infty  \sum_{l=0}^N \int_0^\infty \frac{m_l(t) \Phi_l}{\beta_S(t)} e^{-\beta_l(t)t} s_t^{\alpha_l-1}
\\{}&\nonumber \nonumber
\left[\left(1-\frac{s_t^{\alpha_\nu-1}m_\nu(t)}{\beta_\nu(t)} \Phi_\nu\right)\left(1-\frac{\beta_l(t)}{\beta_S(t)}\right)\right]^{k-1}L_{k-1}\left(\frac{\frac{\beta_l(t)}{\beta_S(t)}(s_t^{\alpha_l-1}m_l(t) \Phi_lt-\beta_l(t)t)}{\left[\left(1-\frac{s_t^{\alpha_\nu-1}m_\nu(t)}{\beta_\nu(t)} \Phi_\nu\right)\left(1-\frac{\beta_l(t)}{\beta_S(t)}\right)\right]}\right)
\\{}&\nonumber \nonumber=
 \sum_{l=0}^N \int_0^\infty \frac{m_l(t) \Phi_l}{\beta_S(t)} e^{-\beta_l(t)t} s_t^{\alpha_l-1}
 \\{}&\nonumber \nonumber
\left[1-\left(1-\frac{s_t^{\alpha_\nu-1}m_\nu(t)}{\beta_\nu(t)} \Phi_\nu\right)\left(1-\frac{\beta_l(t)}{\beta_S(t)}\right)\right]^{-1}
e^{-\frac{\left[\left(1-\frac{s_t^{\alpha_\nu-1}m_\nu(t)}{\beta_\nu(t)} \Phi_\nu\right)\left(1-\frac{\beta_l(t)}{\beta_S(t)}\right)\right]\frac{\beta_l(t)}{\beta_S(t)}(\beta_l(t)t-s_t^{\alpha_l-1}m_l(t) \Phi_lt)}{\left[1-\left(1-\frac{s_t^{\alpha_\nu-1}m_\nu(t)}{\beta_\nu(t)} \Phi_\nu\right)\left(1-\frac{\beta_l(t)}{\beta_S(t)}\right)\right]}}
dt 
\\{}&\nonumber \nonumber=
 \sum_{l=0}^N \int_0^\infty \frac{m_l(t) \Phi_l}{\beta_S(t)} e^{-\beta_l(t)t} s_t^{\alpha_l-1}
 \left[1-\left(1-\frac{s_t^{\alpha_\nu-1}m_\nu(t)}{\beta_\nu(t)} \Phi_\nu\right)\left(1-\frac{\beta_l(t)}{\beta_S(t)}\right)\right]^{-1}\\{}&\nonumber \nonumber
e^{-\frac{\frac{\beta_l(t)}{\beta_S(t)}(\beta_l(t)t -s_t^{\alpha_l-1}m_l(t) \Phi_lt)}{\left[\frac{s_t^{\alpha_\nu-1}m_\nu(t)}{\beta_\nu(t)} \Phi_\nu+\frac{\beta_l(t)}{\beta_S(t)}-\frac{s_t^{\alpha_\nu-1}m_\nu(t)}{\beta_\nu(t)} \Phi_\nu\frac{\beta_l(t)}{\beta_S(t)}\right]}-
\frac{\beta_l(t)}{\beta_S(t)}(\beta_l(t)t-s_t^{\alpha_l-1}m_l(t) \Phi_lt)}
dt 
\end{align}

\begin{align}
\MoveEqLeft[3]
 \sum_{l=0}^N \int_0^\infty \frac{m_l(t) \Phi_l}{\beta_S(t)} e^{-\beta_l(t)t} s_t^{\alpha_l-1}
\left[1-\left(1-\frac{s_t^{\alpha_\nu-1}m_\nu(t)}{\beta_\nu(t)} \Phi_\nu\right)\left(1-\frac{\beta_l(t)}{\beta_S(t)}\right)\right]^{-1} \\{}&\nonumber \nonumber
e^{+\frac{\frac{\beta_l(t)}{\beta_S(t)}(\beta_l(t)t -s_t^{\alpha_l-1}m_l(t) \Phi_lt)}{\left[\frac{s_t^{\alpha_\nu-1}m_\nu(t)}{\beta_\nu(t)} \Phi_\nu+\frac{\beta_l(t)}{\beta_S(t)}-\frac{s_t^{\alpha_\nu-1}m_\nu(t)}{\beta_\nu(t)} \Phi_\nu\frac{\beta_l(t)}{\beta_S(t)}\right]}-
\frac{\beta_l(t)}{\beta_S(t)}(\beta_l(t)t-s_t^{\alpha_l-1}m_l(t) \Phi_lt)}
dt 
\\{}&\nonumber \nonumber=
 \sum_{l=0}^N \int_0^\infty \frac{m_l(t) \Phi_l}{\beta_S(t)} e^{-\beta_l(t)t} s_t^{\alpha_l-1}
\left[1-\left(1-\frac{s_t^{\alpha_\nu-1}m_\nu(t)}{\beta_\nu(t)} \Phi_\nu\right)\left(1-\frac{\beta_l(t)}{\beta_S(t)}\right)\right]^{-1} \\{}&\nonumber \nonumber 
 e^{-
\frac{\beta_l(t)}{\beta_S(t)}(\beta_l(t)t-s_t^{\alpha_l-1}m_l(t) \Phi_lt)\left[\frac{1}{\frac{s_t^{\alpha_\nu-1}m_\nu(t)}{\beta_\nu(t)} \Phi_\nu+\frac{\beta_l(t)}{\beta_S(t)}-\frac{s_t^{\alpha_\nu-1}m_\nu(t)}{\beta_\nu(t)} \Phi_\nu\frac{\beta_l(t)}{\beta_S(t)}}-1\right]}
dt
\\{}&\nonumber \nonumber\leq
 \sum_{l=0}^N\int_0^\infty  \frac{\bar{m}_l \Phi_l}{\bar{\beta}_S}  s_t^{\alpha_l-1+\rho(\alpha_{\nu}-1)}
\left[\frac{\bar{\beta}_l}{\bar{\beta}_S}(1-\frac{\bar{m}_\nu}{\bar{\beta}_\nu} \Phi_\nu)\right]^{-1}
e^{-(2\beta_l-m_l \Phi_l)s_t^{\rho(\alpha_l-1)}t}
dt
\\{}&\nonumber \nonumber=
\sum_{l=0}^N \sum_{p=0}^{P}\left[\frac{\bar{\beta}_l}{\bar{\beta}_S}(1-\frac{\bar{m}_\nu}{\bar{\beta}_\nu} \Phi_\nu)\right]^{-1} \frac{\bar{m}_l \Phi_l}{\bar{\beta}_S}\gamma_p^{\rho(1-\alpha_{\nu})+\alpha_l-1} \int_{T_{p}}^{T_{p+1}}
e^{-(2\bar{\beta}_l-\bar{m}_l \Phi_l)\gamma_p^{\rho(\alpha_l-1)}t}
dt
\\{}&\nonumber \nonumber=
\sum_{l=0}^N \sum_{p=0}^{P}\left[\frac{\bar{\beta}_l}{\bar{\beta}_S}(1-\frac{\bar{m}_\nu}{\bar{\beta}_\nu} \Phi_\nu)\right]^{-1} \frac{\bar{m}_l \Phi_l}{\bar{\beta}_S}\gamma_p^{\rho(1-\alpha_{\nu})+\alpha_l-1} 
\frac{e^{-(2\bar{\beta}_l-\bar{m}_l \Phi_l)\gamma_p^{\rho(\alpha_l-1)}T_{p}}-e^{-(2\bar{\beta}_l-\bar{m}_l \Phi_l)\gamma_p^{\rho(\alpha_l-1)}T_{p+1}}}{(2\bar{\beta}_l-\bar{m}_l \Phi_l)\gamma_p^{\rho(\alpha_l-1)}}
\\{}&\nonumber \nonumber=
\sum_{l=0}^N \sum_{p=0}^{P}\left[\frac{\bar{\beta}_l}{\bar{\beta}_S}(1-\frac{\bar{m}_\nu}{\bar{\beta}_\nu} \Phi_\nu)\right]^{-1} \frac{\bar{m}_l \Phi_l}{\bar{\beta}_S}\gamma_p^{\rho(1-\alpha_{\nu})+\alpha_l-1} 
\frac{e^{-(2\bar{\beta}_l-\bar{m}_l \Phi_l)\gamma_p^{\rho(\alpha_l-1)}Tp}-e^{-(2\bar{\beta}_l-\bar{m}_l \Phi_l)\gamma_p^{\rho(\alpha_l-1)}T(p+1)}}{(2\bar{\beta}_l-\bar{m}_l \Phi_l)\gamma_p^{\rho(\alpha_l-1)}}
\\{}&\nonumber \nonumber=
\sum_{l=0}^N \sum_{p=0}^{P}\left[\frac{\bar{\beta}_l}{\bar{\beta}_S}(1-\frac{\bar{m}_\nu}{\bar{\beta}_\nu} \Phi_\nu)\right]^{-1} \frac{\bar{m}_l \Phi_l}{\bar{\beta}_S}\gamma_p^{\rho(1-\alpha_{\nu})+\alpha_l-1-\rho(\alpha_l-1)} 
\frac{e^{-(2\bar{\beta}_l-\bar{m}_l \Phi_l)\gamma_p^{\rho(\alpha_l-1)}Tp}(1-e^{-(2\bar{\beta}_l-\bar{m}_l \Phi_l)\gamma_p^{\rho(\alpha_l-1)T}})}{(2\bar{\beta}_l-\bar{m}_l \Phi_l)}\;\;\;.
\end{align}
\subsection{Probability of escaping after time $u$}
\label{ssec:proof_u}
We further investigate the probability of exiting before time $u$:
\begin{theorem}
\label{th:ecape_uA}Let $s_t = t^{\gamma-1}$, where $\gamma$ is the cooling rate,let us denote two constants that express the effect of the scheduler: $\gamma_l\triangleq 1+(\gamma-1)(\alpha_l-1)$
and $\kappa \triangleq \frac{1+(\gamma-1)(\alpha_l-1+\rho(\alpha_l-\alpha_\nu))}{\gamma_l}$,
for $u>0$:
\begin{align}
\MoveEqLeft[3]P(\sigma>u) \leq
\sum_{l=0}^N  A_{l,\nu}^{-1}\frac{\bar{\beta}_l\bar{m}_l \Phi_l}{\bar{\beta}_S\gamma_l(\bar{m}_l\Phi_l)^{\kappa}}   \Gamma\left(\kappa,
\bar{m}_l\Phi_lu^{\gamma_l}\right)\;\;\;.
 \end{align}
 \end{theorem}

\noindent
In order to further investigate this expression, let us temporally neglect the cooling effect.
\begin{corollary}
Using Thm.~\ref{th:exit_dir_exp},  for $\gamma \rightarrow 1$:
\begin{align}
\MoveEqLeft[3]P(\sigma>u) \leq
\sum_{l=0}^N  A_{l,\nu}^{-1}\frac{\bar{\beta}_l}{\bar{\beta}_S} e^{-\bar{m}_l\Phi_lu}\;\;\;.
 \end{align}
\end{corollary}
As in $1d$ \cite{imkeller2006first,imkeller2006levy} it can be seen that for small $\epsilon$, the probability is exponentially depends on the time $u$.

\begin{align}
\MoveEqLeft[3]  
P(\sigma>u) = \sum_{k=0}^{\infty} P(\tau_k^*>u) P(\sigma = \tau_k^*) \leq
\sum_{k=0}^{\infty}  P(\tau_k^*>u) P(J^{*}_1 \in \mathcal{G},J^{*}_2 \in \mathcal{G},..,J^{*}_k \notin \mathcal{G}) \\{}&\nonumber \nonumber  = 
\sum_{k=1}^{\infty}\ P(\tau_k^*>u) \prod_{j=1}^{k-1}P(J_j^*\in \mathcal{G})P(J^{*}_k \notin \mathcal{G})\\{}&\nonumber \nonumber
=
\sum_{k=1}^{\infty} P(\tau_k^*>u)(1-P(J_k^*\notin \mathcal{G}))^{k-1}P(J^{*}_k \notin \mathcal{G})
\\{}&\nonumber \nonumber\approx
\sum_{k=1}^{\infty}\sum_{l=1}^{N}
\sum_{w=1}^{k-1}(1-P(s_t^{\frac{\alpha_l-1}{\alpha_l}}\epsilon\mathbf{1}^t\lambda_l(t)J_w^l\notin \mathcal{G}))^{w-1}(1-P(s_t^{\frac{\alpha_\nu-1}{\alpha_\nu}}\epsilon\mathbf{1}^t\lambda_\nu(t)J_w^v\notin \mathcal{G}))^{k-w}
\\{}&\nonumber \nonumber
P(s_t^{\frac{\alpha_l-1}{\alpha_l}}\epsilon\mathbf{1}^t\lambda_l(t)J_w^l\notin \mathcal{G})P(\tau_w^l=\tau_k^*) P(\tau_w^l>u)
\\{}&\nonumber \nonumber 
=
\sum_{k=1}^\infty  \sum_{l=0}^N \sum_{w=1}^{k} \int_u^\infty \frac{\beta_l(t)}{\beta_S(t)}\frac{(k-1)!}{(w-1)!(k-w)!}\left(\frac{\beta_l(t)}{\beta_S(t)}\right)^{w-1}\left(1-\frac{\beta_l(t)}{\beta_S(t)}\right)^{k-w}\\{}&\nonumber \nonumber e^{-\beta_l(t)t}\frac{(\beta_l(t)t)^{w-1}}{(w-1)!} [1-\frac{s_t^{\alpha_l-1}m_l(t)}{\beta_l(t)} \Phi_l]^{w-1}
[1-\frac{s_t^{\alpha_\nu-1}m_\nu(t)}{\beta_\nu(t)} \Phi_\nu]^{k-w}\frac{s_t^{\alpha_l-1}m_l(t)}{\beta_l(t)} \Phi_ldt
\\{}&\nonumber \nonumber= 
\sum_{k=1}^\infty  \sum_{l=0}^N \int_0^\infty \frac{\beta_l(t)}{\beta_S(t)} e^{-\beta_l(t)t} s_t^{\alpha_l-1}m_l(t) \Phi_l
\\{}&\nonumber \nonumber
\sum_{w=1}^{k}\frac{[\beta_l(t)t-s_t^{\alpha_l-1}m_l(t) \Phi_lt]^{w-1}}{(w-1)!}\frac{(k-1)!}{(w-1)!(k-w)!}\left(\frac{\beta_l(t)}{\beta_S(t)}\right)^{w-1}\left[\left(1-\frac{s_t^{\alpha_\nu-1}m_\nu(t)}{\beta_\nu(t)} \Phi_\nu\right)\left(1-\frac{\beta_l(t)}{\beta_S(t)}\right)\right]^{k-w}dt
\\{}&\nonumber \nonumber=
\sum_{k=1}^\infty  \sum_{l=0}^N \int_u^\infty \frac{\beta_l(t)m_l(t) \Phi_l}{\beta_S(t)} e^{-\beta_l(t)t} s_t^{\alpha_l-1}
\\{}&\nonumber \nonumber
\left[\left(1-\frac{s_t^{\alpha_\nu-1}m_\nu(t)}{\beta_\nu(t)} \Phi_\nu\right)\left(1-\frac{\beta_l(t)}{\beta_S(t)}\right)\right]^{k-1}L_{k-1}\left(\frac{\frac{\beta_l(t)}{\beta_S(t)}(s_t^{\alpha_l-1}m_l(t) \Phi_lt-\beta_l(t)t)}{\left[\left(1-\frac{s_t^{\alpha_\nu-1}m_\nu(t)}{\beta_\nu(t)} \Phi_\nu\right)\left(1-\frac{\beta_l(t)}{\beta_S(t)}\right)\right]}\right)
\\{}&\nonumber \nonumber=
 \sum_{l=0}^N \int_u^\infty \frac{\beta_l(t)m_l(t) \Phi_l}{\beta_S(t)} e^{-\beta_l(t)t} s_t^{\alpha_l-1}
 \\{}&\nonumber \nonumber
\left[\left(1-\frac{s_t^{\alpha_\nu-1}m_\nu(t)}{\beta_\nu(t)} \Phi_\nu\right)\left(1-\frac{\beta_l(t)}{\beta_S(t)}\right)\right]^{-1}
e^{-\frac{\frac{\beta_l(t)}{\beta_S(t)}(s_t^{\alpha_l-1}m_l(t) \Phi_lt - \beta_l(t)t)}{\left[1-\left(1-\frac{s_t^{\alpha_\nu-1}m_\nu(t)}{\beta_\nu(t)} \Phi_\nu\right)\left(1-\frac{\beta_l(t)}{\beta_S(t)}\right)\right]}}
dt 
\\{}&\nonumber \nonumber\approx
 \sum_{l=0}^N  \left[\left(1-\frac{\bar{m}_\nu}{\bar{\beta}_\nu} \Phi_\nu\right)\left(1-\frac{\bar{\beta}_l}{\bar{\beta}_S}\right)\right]^{-1} \int_u^\infty \frac{\bar{\beta}_l\bar{m}_l \Phi_l}{\bar{\beta}_S}  s_t^{\alpha_l-1+\rho(\alpha_l-\alpha_\nu)}
e^{-s_t^{\alpha_l-1}\bar{m}_l \Phi_lt}
dt 
\end{align}
Exponential scheduler
\begin{align}
\MoveEqLeft[3]
 \sum_{l=0}^N  \left[\left(1-\frac{\bar{m}_\nu}{\bar{\beta}_\nu} \Phi_\nu\right)\left(1-\frac{\bar{\beta}_l}{\bar{\beta}_S}\right)\right]^{-1} \int_u^\infty \frac{\bar{\beta}_l\bar{m}_l \Phi_l}{\bar{\beta}_S}  s_t^{\alpha_l-1+\rho(\alpha_l-\alpha_\nu)}
e^{-s_t^{\alpha_l-1}\bar{m}_l \Phi_lt}
dt \\{}&\nonumber \nonumber=
 \sum_{l=0}^N  \left[\left(1-\frac{\bar{m}_\nu}{\bar{\beta}_\nu} \Phi_\nu\right)\left(1-\frac{\bar{\beta}_l}{\bar{\beta}_S}\right)\right]^{-1} \int_u^\infty \frac{\bar{\beta}_l\bar{m}_l \Phi_l}{\bar{\beta}_S}  t^{(\gamma-1)(\alpha_l-1+\rho(\alpha_l-\alpha_\nu))}
e^{-t^{1+(\gamma-1)\rho(\alpha_l-1)}\bar{m}_l \Phi_l}
dt \\{}&\nonumber \nonumber=
\sum_{l=0}^N  \left[\left(1-\frac{\bar{m}_\nu}{\bar{\beta}_\nu} \Phi_\nu\right)\left(1-\frac{\bar{\beta}_l}{\bar{\beta}_S}\right)\right]^{-1}\frac{\bar{\beta}_l\bar{m}_l \Phi_l}{\bar{\beta}_S} \\{}&\nonumber \nonumber \left[ \frac{(\bar{m}_l\Phi_l)^{-\frac{1+(\gamma-1)(\alpha_l-1+\rho(\alpha_l-\alpha_\nu))}{1+(\gamma-1)\rho(\alpha_l-1)}}\Gamma\left({\frac{1+(\gamma-1)(\alpha_l-1+\rho(\alpha_l-\alpha_\nu))}{1+(\gamma-1)\rho(\alpha_l-1)}},
\bar{m}_l\Phi_lu^{1+(\gamma-1)\rho(\alpha_l-1)}\right)}{1+(\gamma-1)\rho(\alpha_l-1)}\right]
\end{align}
For $\gamma \rightarrow 1$:
\begin{align}
\MoveEqLeft[3] 
\sum_{l=0}^N  \left[\left(1-\frac{\bar{m}_\nu}{\bar{\beta}_\nu} \Phi_\nu\right)\left(1-\frac{\bar{\beta}_l}{\bar{\beta}_S}\right)\right]^{-1}\frac{\bar{\beta}_l}{\bar{\beta}_S} \Gamma\left(1,
\bar{m}_l\Phi_lu\right)\\{}&\nonumber \nonumber=
\sum_{l=0}^N  \left[\left(1-\frac{\bar{m}_\nu}{\bar{\beta}_\nu} \Phi_\nu\right)\left(1-\frac{\bar{\beta}_l}{\bar{\beta}_S}\right)\right]^{-1}\frac{\bar{\beta}_l}{\bar{\beta}_S} e^{-\bar{m}_l\Phi_lu}\;\;\;.
\end{align}
\section{Extras}
\lemma
$\forall T\in[S_j,S_{j+1}
]$,  $\forall j\in \mathbb{N}$,and  $\forall w \in [d_i^-,d_i^+]$ there exist a finite $C_Z$ s.t:
\begin{equation}
    \sup_{T} |X_{t}^i(w)| \leq C_z^I \sup_T |\xi_t^{i}|\;\;\;.
\end{equation}
Using stochastic asymptotic expansion: 
\begin{equation}
    |X_{t}^i(w)| \leq \sup_{t \in [0,T]}|\xi_{t,l}|\left(1+\sup_{t \in [0,T]}\int_0^t H(Y_p(w))_{ii}e^{-\int_p^tH(Y_u(w))_{ii}du}dp
\right)\;\;\;.
\end{equation}
For some $\delta>0$, the inequality :$m_{1}^i \leq \sup_{|w|\leq \delta}H(Y_p(w))\leq  \inf_{|w| \leq\delta}H(Y_p(w))\leq m_{2}^i$.\newline
Let us denote:
\begin{equation}
    C_1 = \max_{w\in \mathcal{G}} \int_0^{\hat{T}} H(Y_p(w))_{ii}e^{-\int_p^tH(Y_u(w))_{ii}du}dp\;\;\;.
\end{equation}
For arbitrary $\hat{T}\leq t$:
\begin{align}
\MoveEqLeft[3]
\int_0^t H(Y_p(w))_{ii}e^{-\int_p^tH(Y_u(w))_{ii}du}dp =
\\{}&\nonumber \nonumber \int_0^{\hat{T}} H(Y_p(w))_{ii}e^{-\int_p^tH(Y_u(w))_{ii}du}dp+\int_{\hat{T}}^t H(Y_p(w))_{ii}e^{-\int_p^tH(Y_u(w))_{ii}du}dp 
\end{align}
The estimate for the first term:
\begin{align}
\MoveEqLeft[3]
 \int_0^{\hat{T}} H(Y_p(w))_{ii}e^{-\int_{p}^tH(Y_u(w))_{ii}du}dp=e^{-\int_{\hat{T}}^tH(Y_u(w))_{ii}du}\int_0^{\hat{T}} H(Y_p(w))_{ii}e^{-\int_p^{\hat{T}}H(Y_u(w))_{ii}du}dp \\{}&\nonumber \nonumber \leq 
 e^{-m_{1}^i(t-\hat{T})} C_1\leq C_1 \;\;\;.
\end{align}
The second sum:
\begin{align}
\MoveEqLeft[3]
\int_{\hat{T}}^t H(Y_p(w))_{ii}e^{-\int_p^tH(Y_u(w))_{ii}du}dp\leq
\int_{\hat{T}}^t m_{2}^ie^{-m_{1}^i(t-p)}dp\leq \frac{m_{2}^i}{m_{1}^i}\;\;\;.
\end{align}
And:
$ C_Z^l = C_1 +\frac{m_2^l}{m_1^l}$.

\section{Framework properties and notations}
\label{sec:assumptions}
Let us first make few assumptions on the geometry of $\mathcal{G}$ and notations:
\begin{enumerate}
    \item Near the basin $W^*$, $\nabla U :\bar{\mathcal{G}}\rightarrow \mathbb{R}^d$.
     \item  $U$ is $\mu-$strongly convex  .
\item The boundary of our domain is denoted as $\partial  \mathcal{G}$, which is a $C^1$ manifold, so that the vector field of the outer normals on the boundary exists. This means that $\nabla U$ “points into $\mathcal{G}$”, hence:
\begin{equation}
    \langle\nabla U(w),n(w)\rangle
     <-\frac{1}{C}\;\;\;,
\end{equation}
for any $w\in \partial \mathcal{G}$
\item Zero is an attractor of the domain (i.e. $\nabla U(0)=0$, and for every starting value $w \in \mathcal{G}$,the deterministic solution vanishes asymptotically:
\begin{equation}
    lim_{t\rightarrow \infty}Y_t(w)\rightarrow 0\;\;\;.
\end{equation}
\item Let us define the inner part of $\mathcal{G}$ as $\mathcal{G}_\delta=\{y \in \mathcal{G}:dict(w,\partial\mathcal{G})\geq \delta \}$\;\;\;,
\end{enumerate}
where $C>1$.
\newline
Let us define $\delta_0>0$ as the point which if $\norm{w}<\delta_0$ then $w\in \mathcal{G}$ and $\forall \delta \in(0,\delta)$.
The following is valid:
\begin{itemize}
    \item From the exponential stability of 0, $\norm{Y_t}<Ce^{-\frac{1}{C}t}\norm{w}$.
    \item For $\norm{w}<\delta_0$, and $g^i_{w,+}=w+tr_i$, $g^i_{w,-}=w-tr_i$, we shall define the distance  to the boundary as:
    \begin{equation}
        d_i^+(w) \triangleq inf\{t>0 :g^i_{w,+}(t)\in \partial \mathcal{G}\}\;\;\;.
    \end{equation}
    \item We will define  $\delta$-tubes as $\Omega_i^+(\delta)\triangleq\{w\in\mathbb{R}^d:\norm{\langle w,r_i\rangle r_i}< \delta,\langle w,r_i\rangle >0   \}\cap \mathcal{G}^c$
    and $\Omega_i^-(\delta)\triangleq\{w\in\mathbb{R}^d:\norm{\langle w,r_i\rangle r_i}< \delta,\langle w,r_i\rangle <0   \}\cap \mathcal{G}^c$.
    \item $\mathcal{G}_\delta$ with the dynamic process $Y_t$ and the initial point $w\in \mathcal{G}_\delta$ is a Positively invariant set \label{item:positively_invariant} \cite{amann2011gewohnliche}  .
\end{itemize}
\section{Constructing the SDE}
The stochastic gradient descent update rule we follow is:
\begin{equation}
    w_{n+1}=w_n-s_{n+1}\eta\nabla U(w_n)+s_{n+1}\eta\lambda_{n+1} S_{n+1}
\end{equation}
Where $\eta$ is the learning rate, $s_k$ is the learning rate scheduler, $\lambda\in \mathbb{R}^N$ is the noise covariance matrix of $S_k$, which is a $N$-dimensional vector where each element is distributed as a symmetric alpha stable random variable.
\newline
Let us denote $\bar{\eta}_n=s_n\eta$, hence:
\begin{equation}
        w_{n+1}=w_n-\bar{\eta}_{n+1}\nabla U(w_n)+\bar{\eta}_{n+1}\lambda_{n+1}S_{n+1} \;\;\; .
\end{equation}
\newline
The Euler-Maruyama discretization of a L\'evy $\mathcal{S\alpha S}$, which is given as follows\cite{2005math......9712P,2017arXiv170603649S,duan2015introduction}
 \begin{equation}
     W_{n+1} = W_{n}-\bar{\eta}_{n+1}\nabla U(W_n)+\bar{\eta}_{n+1}^{\frac{\alpha-1}{\alpha}}\lambda_{n+1}\Delta  L_{n+1}^\alpha  \;\;\; .
 \end{equation}
Using the Euler-Maruyama discretization and using different representation, we obtain the following SDE:
\begin{equation}
    W_t = w+  \int_{0}^{t} \nabla U(W_p)\,dp +\sum_{i=1}^{N}s_t^{\frac{\alpha_l-1}{\alpha_l}} \epsilon\mathbf{1}^T\lambda_i(t) r_i L_t^i   \;\;\; .
\end{equation}
Can also be written as:
\begin{equation}
    W_t = w+  \int_{0}^{t}\nabla U(W_p)\,dp +\int_{0}^{t} \sum_{i=1}^{N} s_t^{\frac{\alpha_l-1}{\alpha_l}} \epsilon\mathbf{1}^T\lambda_i(t) r_i dL_t^i  \;\;\; .
\end{equation}
$L_t^i$ are independent $1d$ mean-zero Lévy processes
with symmetric $\alpha_i$-stable components,
where $\alpha_i\in(0,2)$.
using this definition of $\lambda_i$, our framework is able to express anisotropic and  state dependent noise (even though the noise matrix is diagonal), $\epsilon$ is a function of the learning rate, $\epsilon(\alpha_i) = \epsilon=\eta^{\frac{\alpha_i-1}{\alpha_i}}$ where, $\eta$ is the learning rate. $s_t$  is the scheduler that decrease the learning rate during the optimization process, hence we can remark that $s_{t_1}\epsilon \geq s_{t_2}\epsilon, \forall  t_1 \leq t_2 $.
\newline 
Notice that $N$ does not have to be equal to $d$, which means some neurons may have the same noise distribution.
\newline
The convergence of the Euler-Maruyama discritization please see \cite{jacod2005approximate,protter1997euler,bally1996law}.

\section{Experimental Section}
\subsection{LRdecay plot}
\begin{figure*}[!h]
\begin{tabular}{cc}
\hspace{-0.1in}\includegraphics[width=.4875\linewidth]{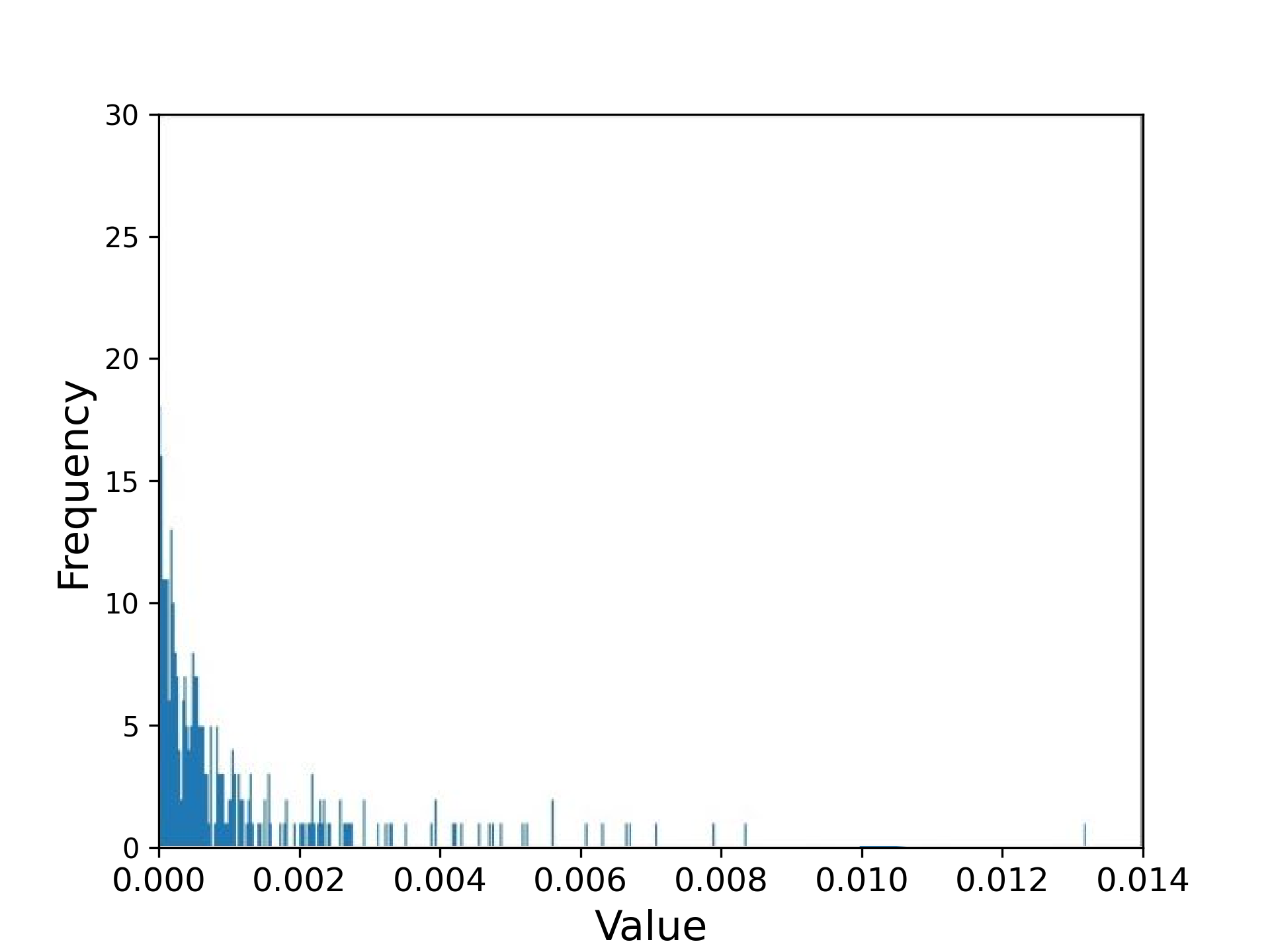}
  \hspace{-0.1in}
 \includegraphics[width=.4875\linewidth]{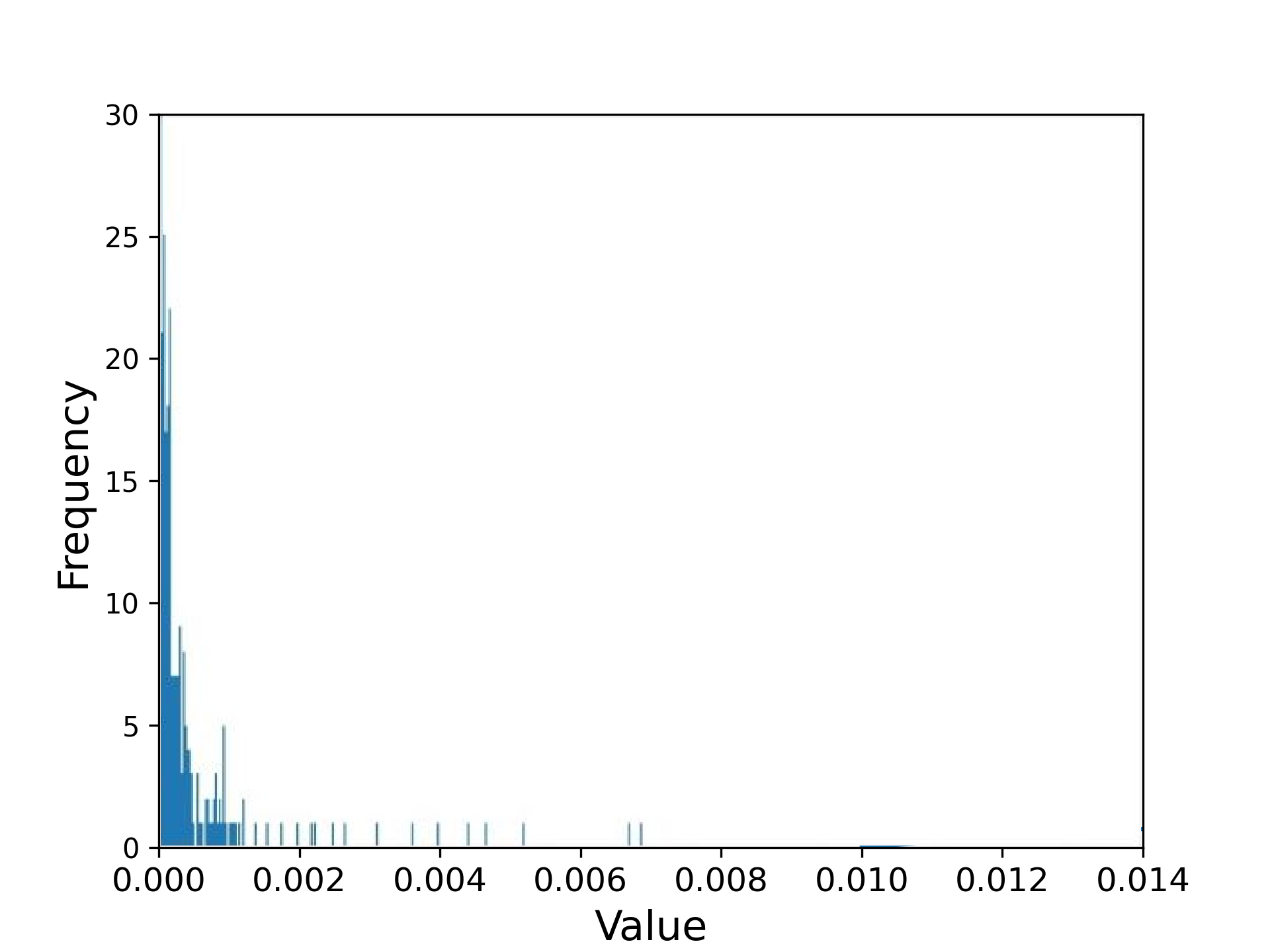}\\
 \hspace{-3.3in}(a)&\hspace{-1.85in}(b)\\
\hspace{-0.1in}\includegraphics[width=.4875\linewidth]{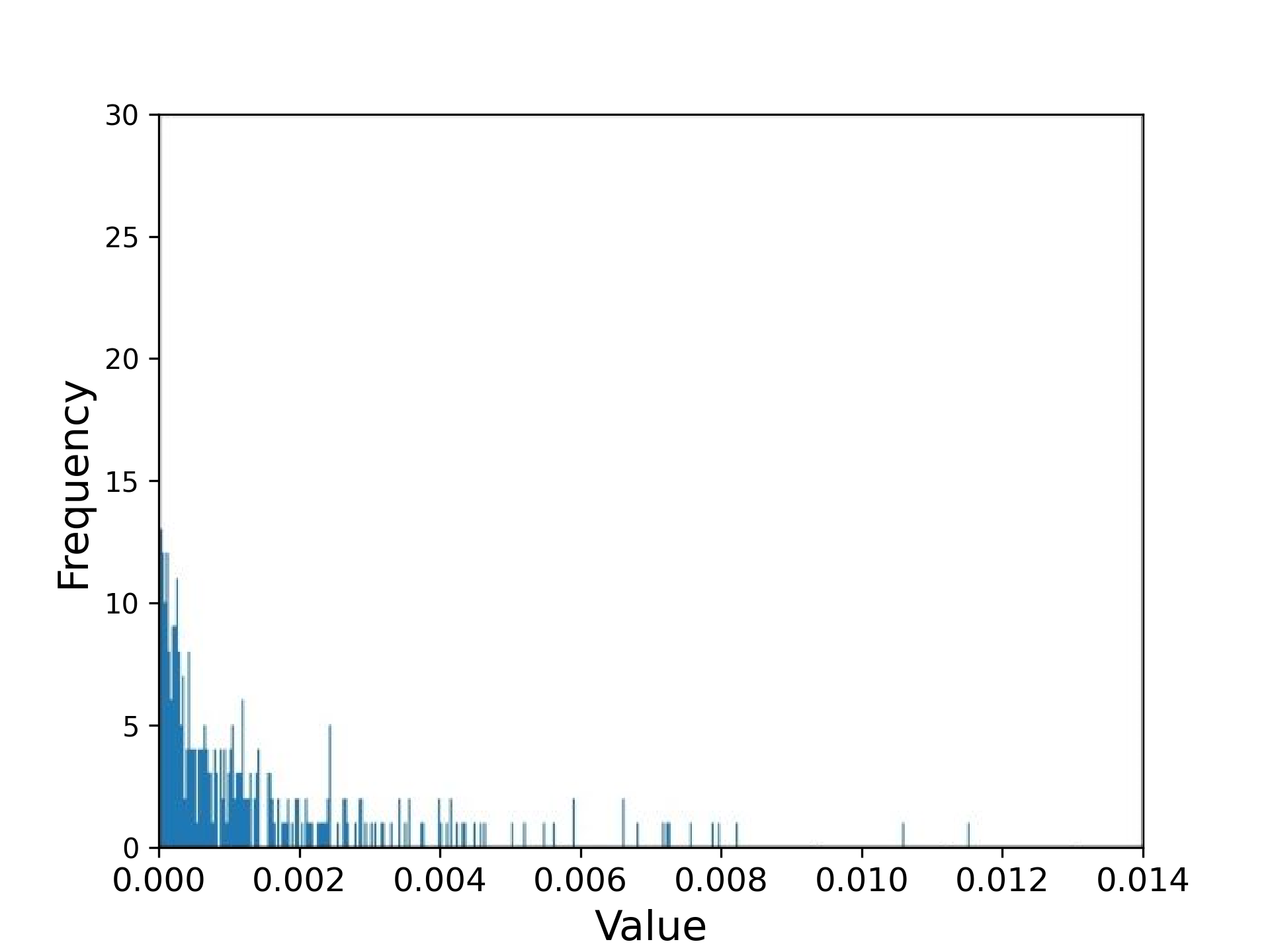}
\hspace{-0.1in}
 \includegraphics[width=.4875\linewidth]{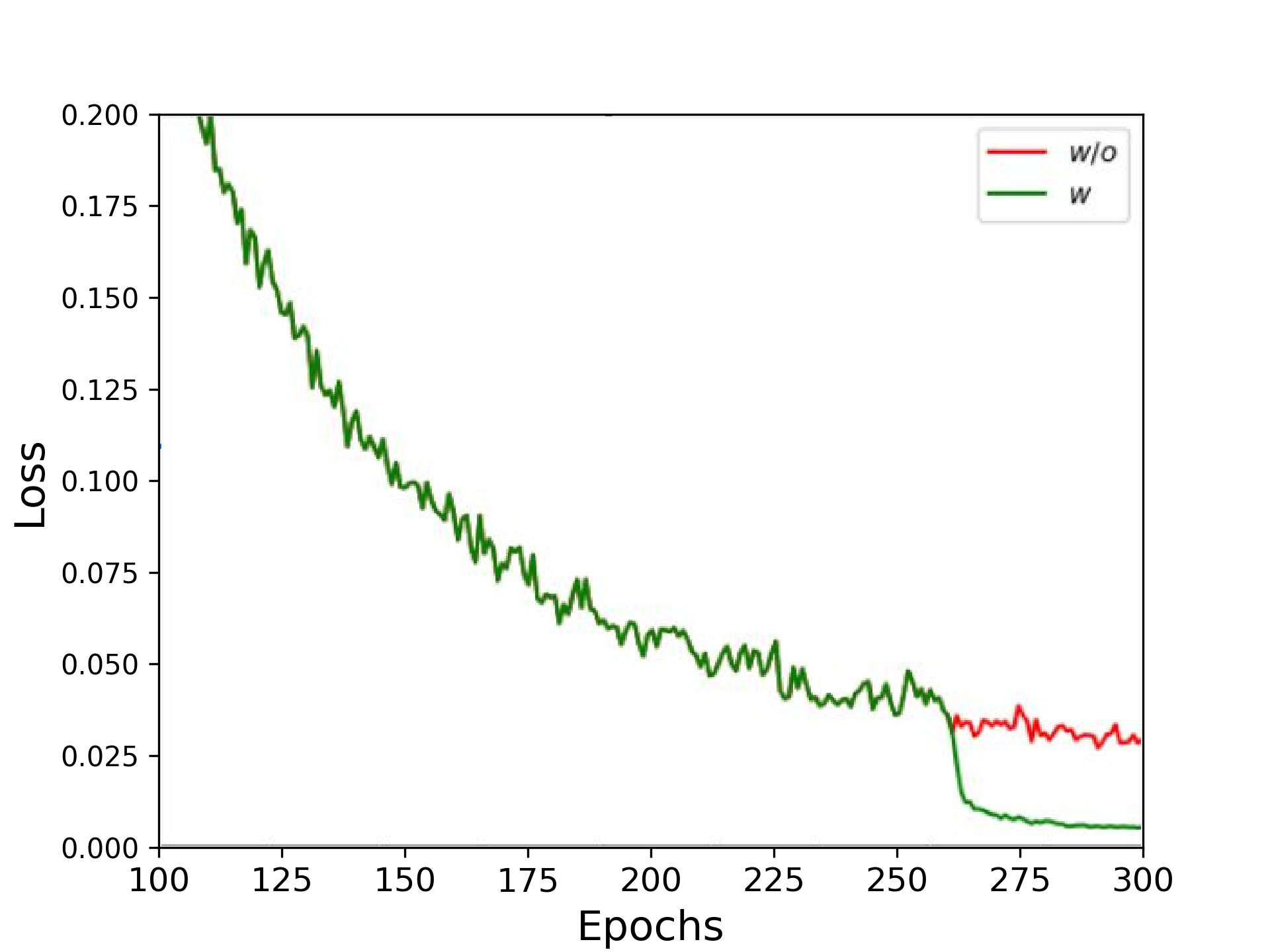}\\
\hspace{-3.55in}(a)&\hspace{-2.9in}(b)\\
\end{tabular}
\caption{ The stochastic gradient noise of a single parameter in ResNet110 \cite{2015arXiv151203385H}. (a) Before applying learning rate decay, at epoch 279. (b) After applying learning rate decay, at epoch 281. (c) Without learning rate decay, at epoch 280. (d) The training loss with and without learning rate decay applied at epoch 280. }
\label{fig:lrdecplots}
 \end{figure*}
 \subsection{SGN another visual examples}
\label{sec:more_visual_histo}
 \begin{figure}[H]
\begin{tabular}{cc}
\includegraphics[width=.5475\linewidth]{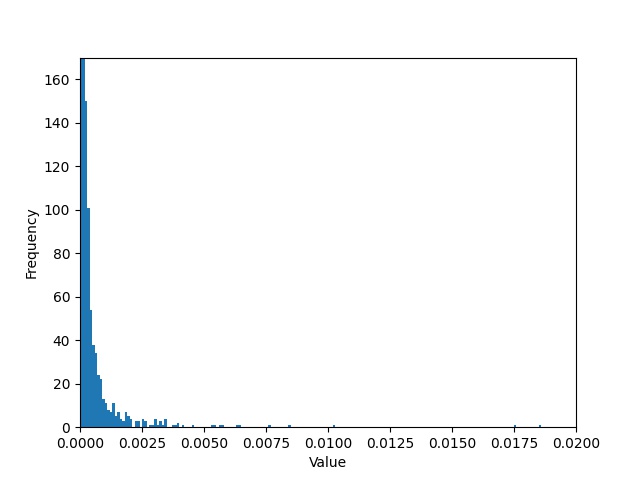}
  \hspace{-0.1in}
 \includegraphics[width=.5475\linewidth]{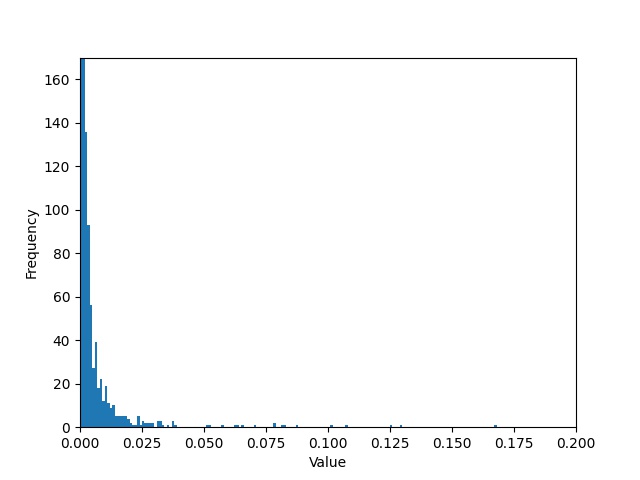}\\
 \hspace{-3.55in}(a)&\hspace{-2.9in}(b)\\
\includegraphics[width=.5475\linewidth]{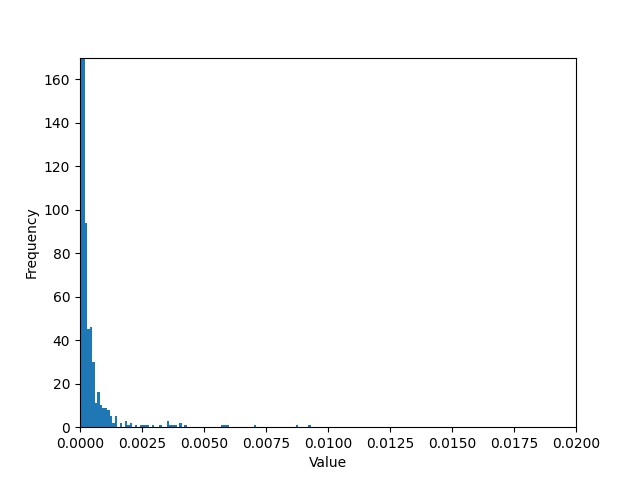}
  \hspace{-0.1in}
 \includegraphics[width=.5475\linewidth]{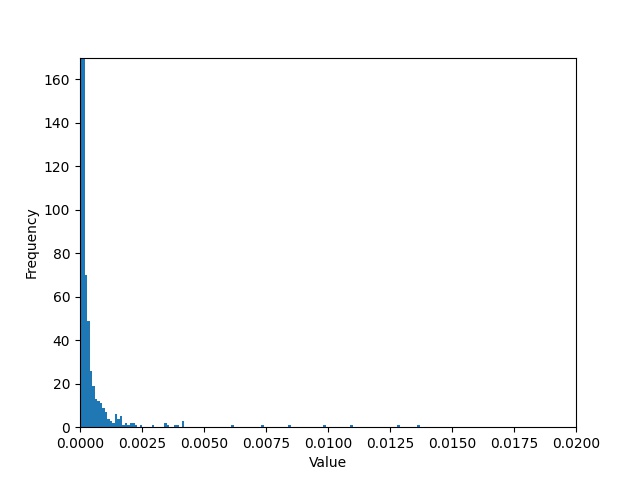}\\
\hspace{-3.55in}(c)&\hspace{-2.9in}(d)\\
\end{tabular}
\caption{ The stochastic gradient noise of a ResNet50 trained on CIFAR100  for four randomly sampled parameters, please zoom in in order to see the long tail behaviour. }
\label{fig:cifar100_r50}
\end{figure}
\begin{figure}[H]
\begin{tabular}{cc}
\includegraphics[width=.5475\linewidth]{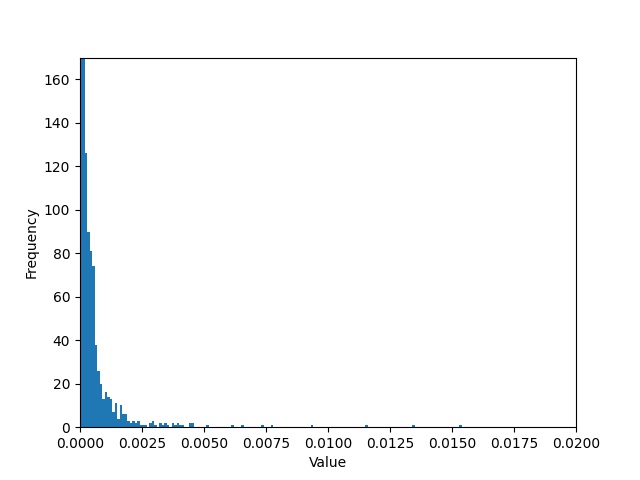}
\includegraphics[width=.5475\linewidth]{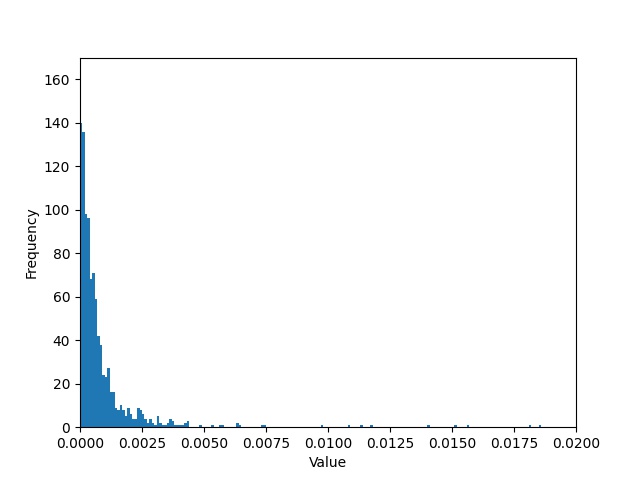}\\
 \hspace{-3.55in}(a)&\hspace{-2.9in}(b)\\
\includegraphics[width=.5475\linewidth]{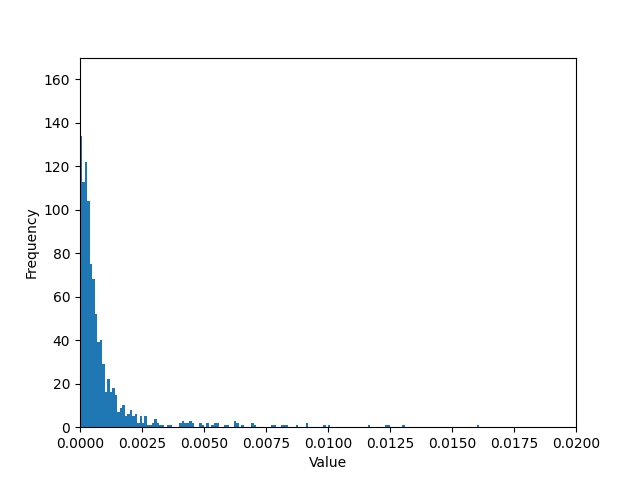}
\includegraphics[width=.5475\linewidth]{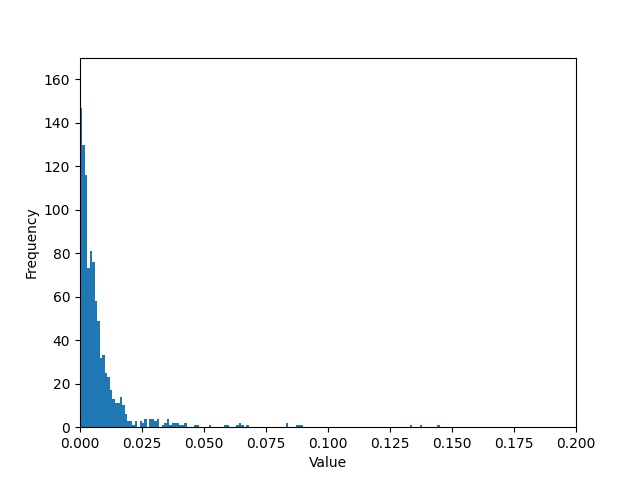}\\
\hspace{-3.55in}(c)&\hspace{-2.9in}(d)\\
\end{tabular}
\caption{ The stochastic gradient noise of a ResNet18 trained on CIFAR100  for four randomly sampled parameters, please zoom in in order to see the long tail behaviour. }
\label{fig:cifar100_r18}
\end{figure}

\subsection{More Escape time experiments}
\label{sec:escapetime_moreexp}
 \begin{figure}[H]
\begin{tabular}{ccc}
 \includegraphics[width=.3225\linewidth]{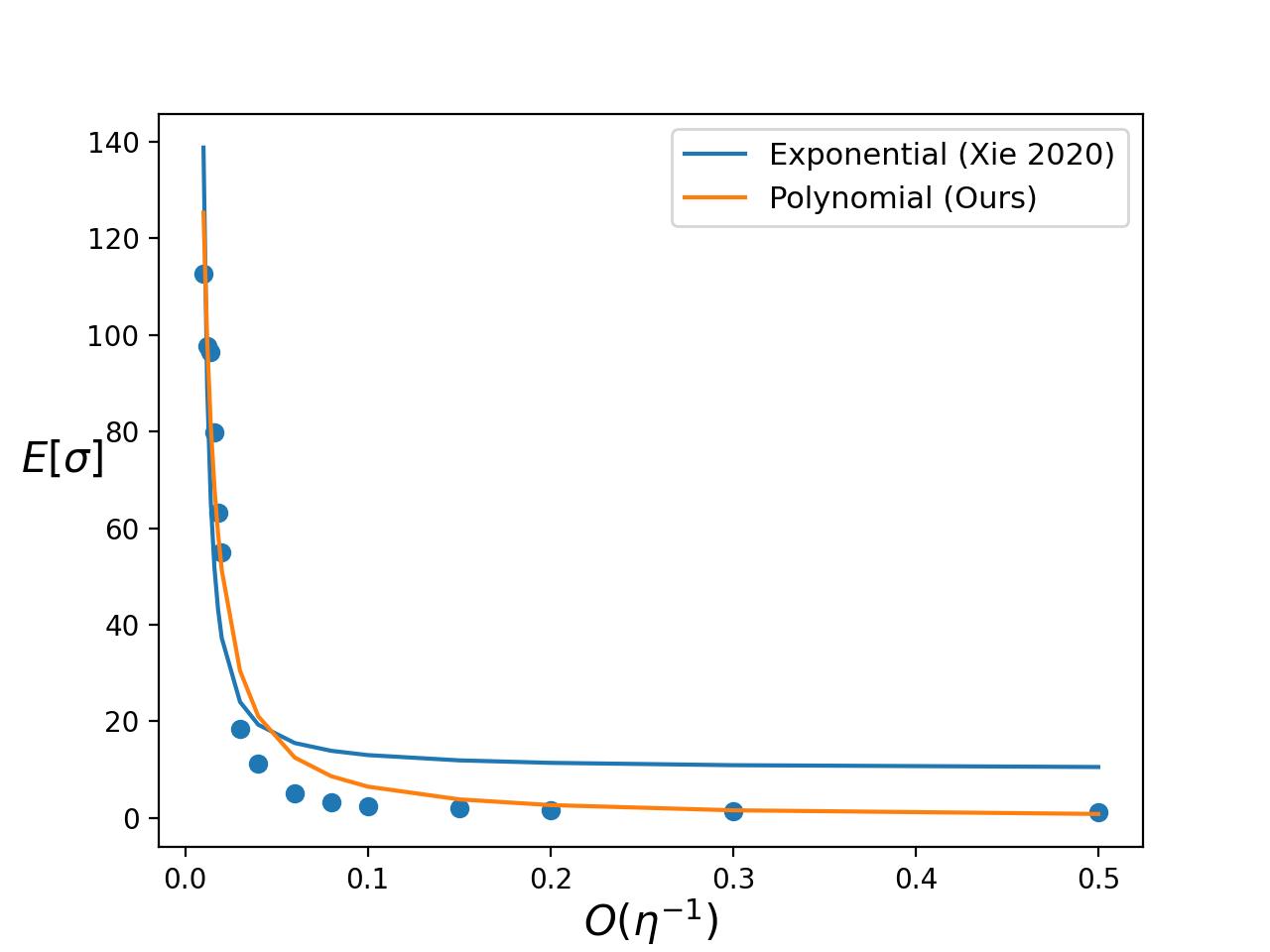}
 \includegraphics[width=.3225\linewidth]{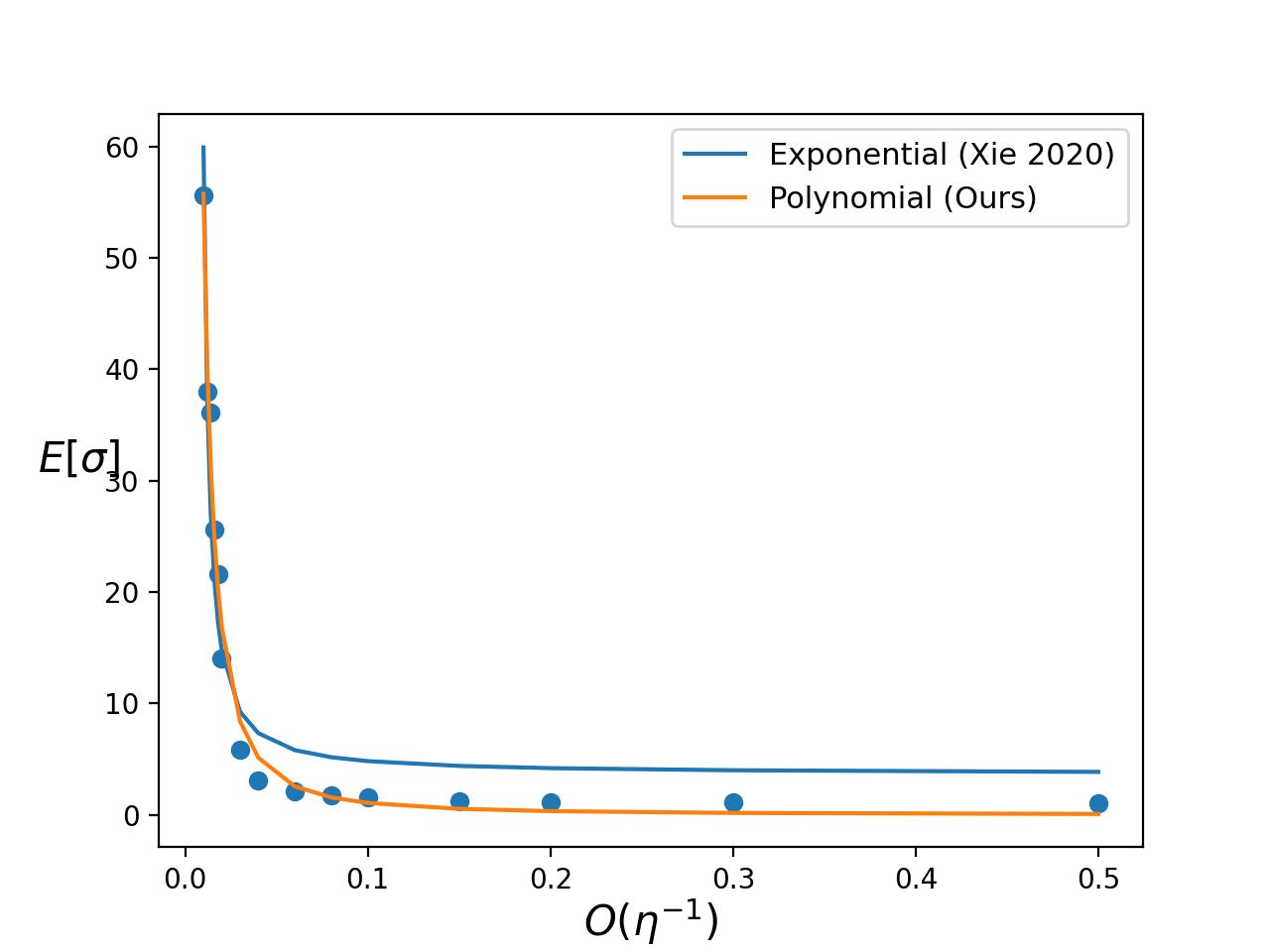}
 \includegraphics[width=.3225\linewidth]{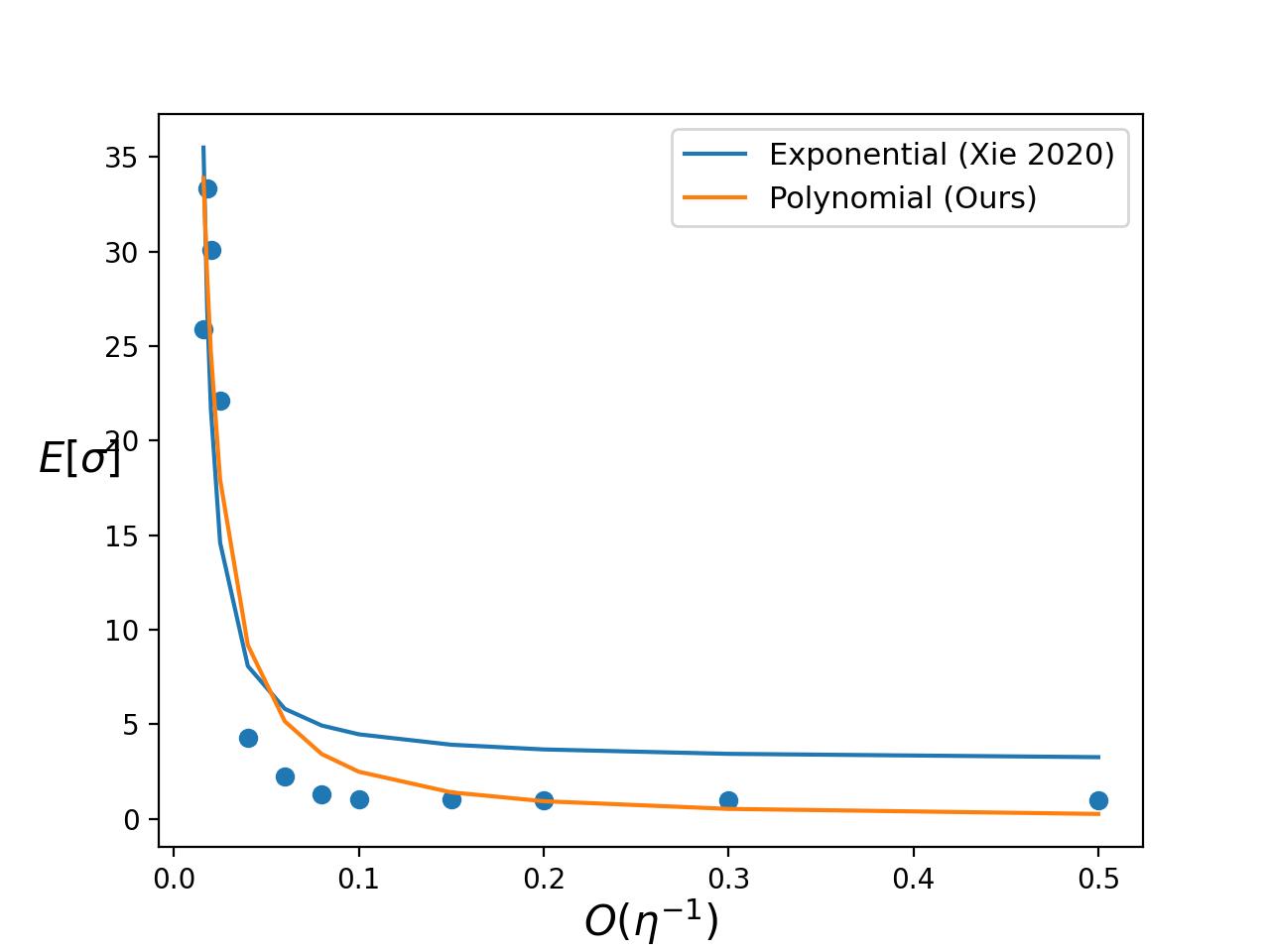} 
 \\
 \includegraphics[width=.3225\linewidth]{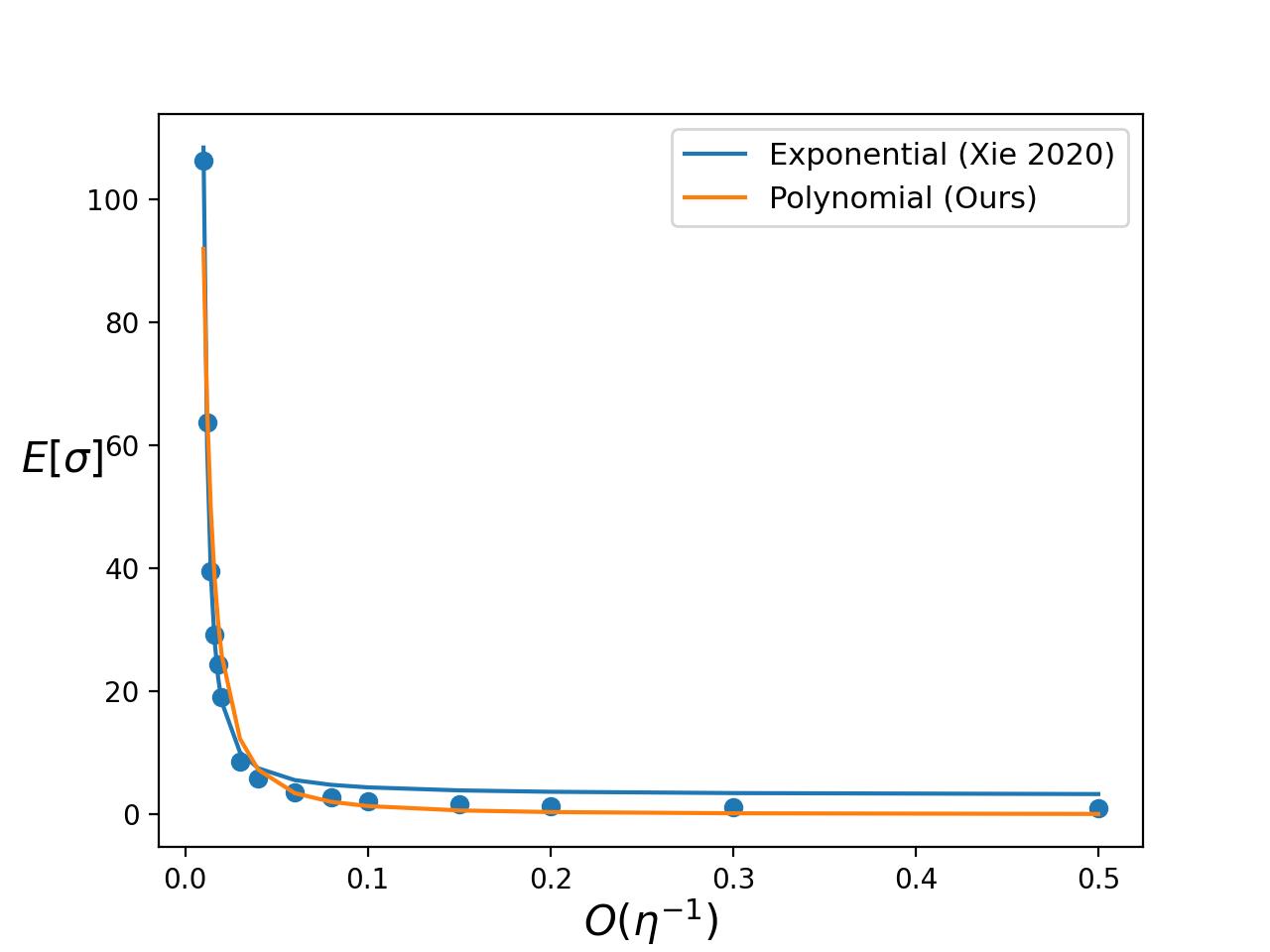}
 \includegraphics[width=.3225\linewidth]{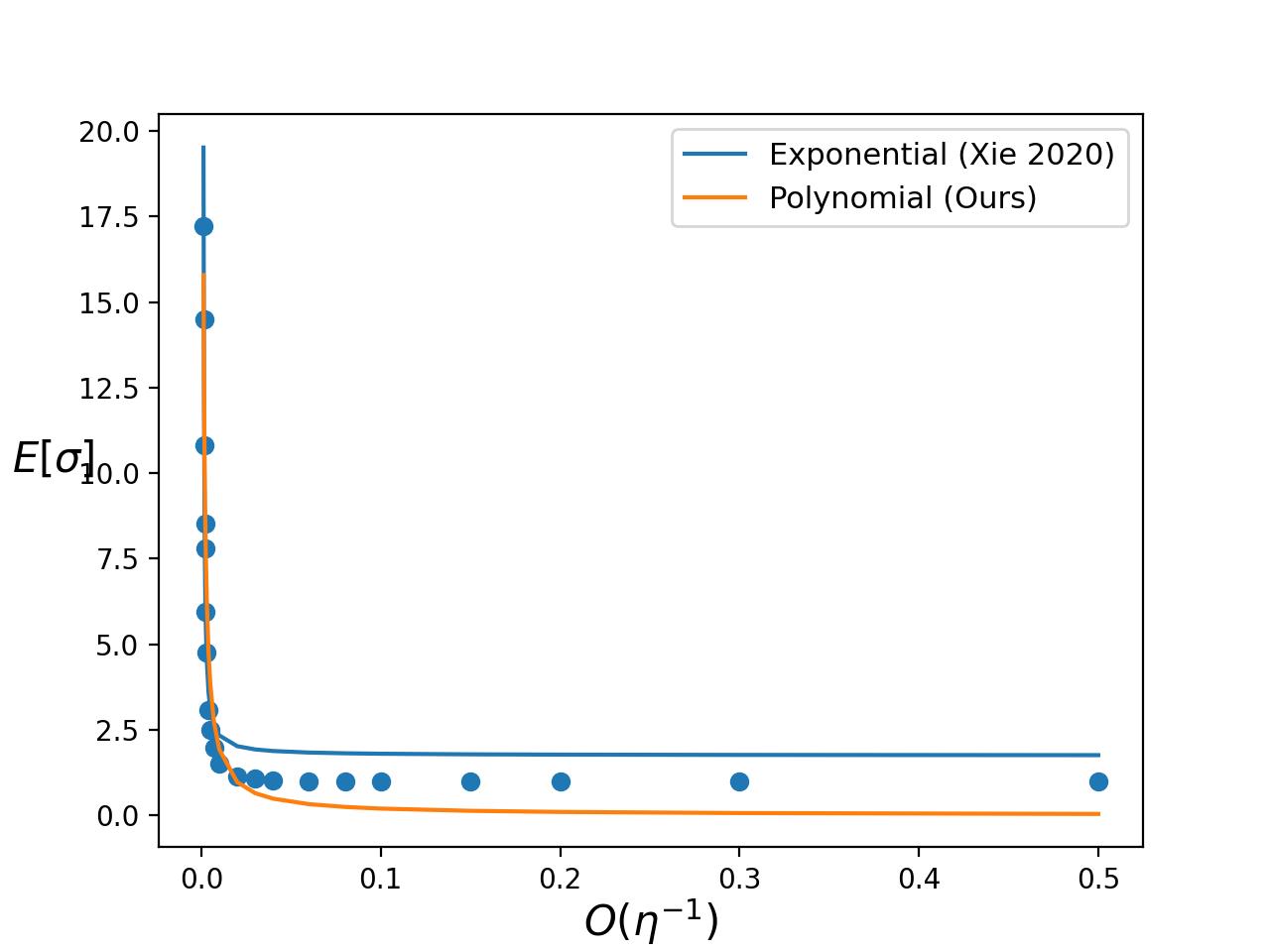}
 \includegraphics[width=.3225\linewidth]{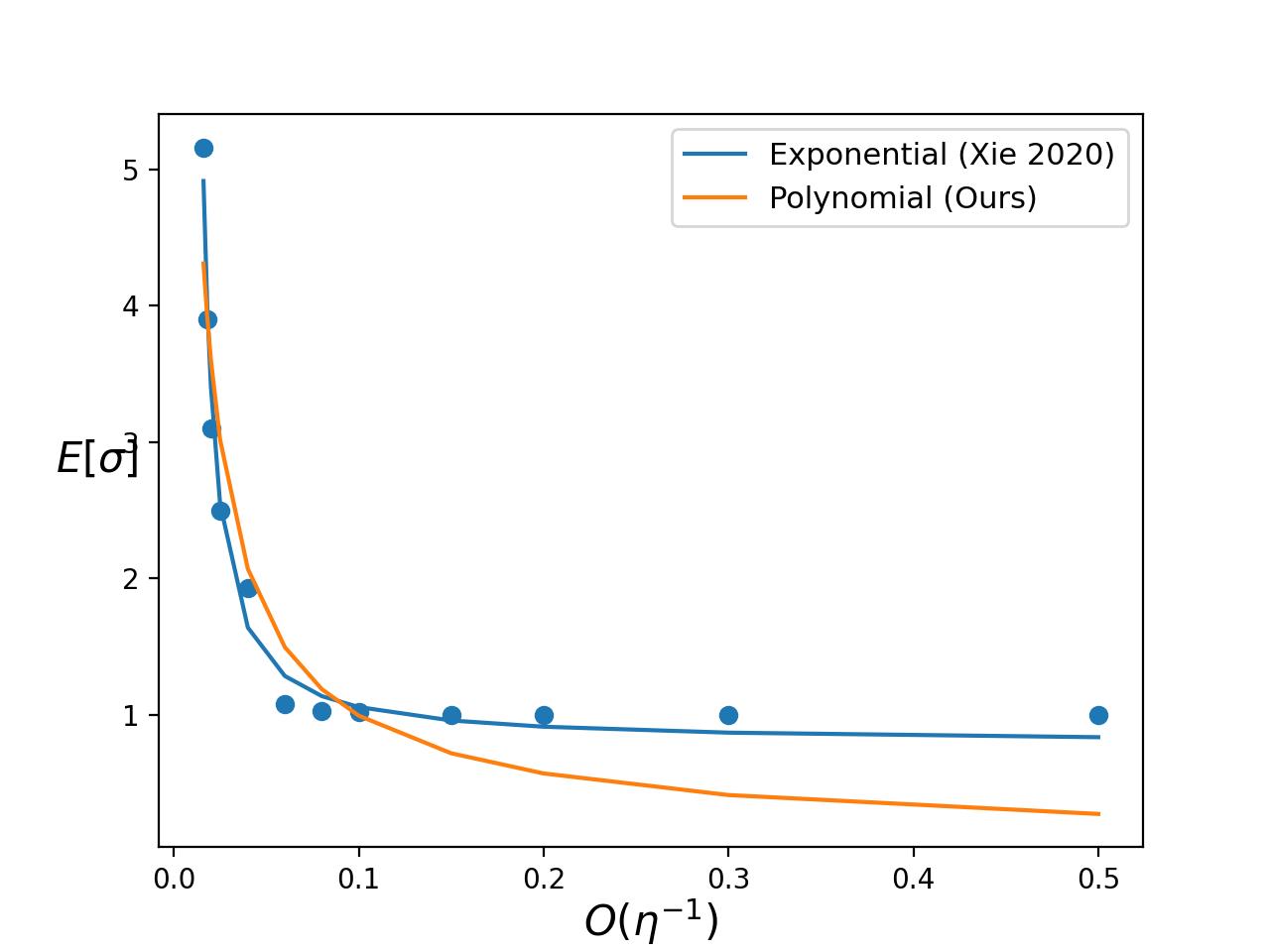} 

\end{tabular}
\caption{The mean escape time of SGD on Breastw (left), Cardio (middle), and Satellite (right) datasets. The plots show the fitting base on two  methods: ours and \cite{xie2020diffusion}, on the upper row shows escaping with batch size 32, while the bottom row is with batch size 8. Each dot represents the mean escape time for a sweep of learning rates. For each learning rate, the dot is an average of over $100$ random seeds. One can observe that the empiric results are better explained by our theory for batch size of 32 in all three datasets examined. On the contrary using batch size 8, our theory overshoot when predicting escape time for Satellite dataset, competitive on Cardio and better on BreastW dataset. }
\end{figure}
\subsection{$\alpha_i$ Variability}
\begin{figure}[H]
\centering
\includegraphics[width=.525\linewidth]{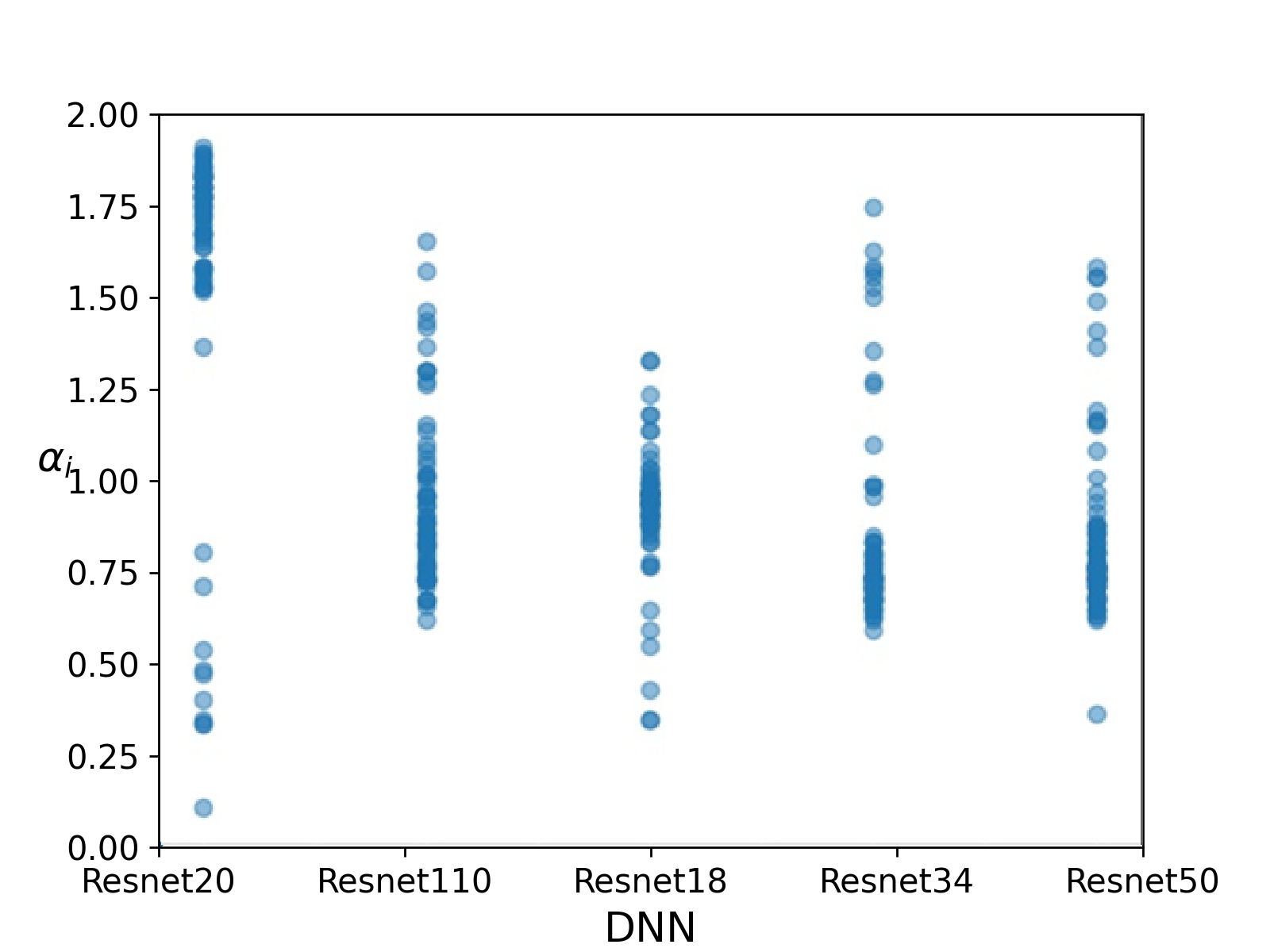}
\caption{
Each dot represents the distribution parameter $\alpha_i$ of a single weight in the DNN. Values on the x-axis represent five different DNNs, left to right: ResNet20/110/18/34/50 \cite{2015arXiv151203385H}; this plot confirms that distinct weights in a DNN lead to different noise distributions during training. }\label{fig:mean_and_std}
\end{figure}

\subsection{Escape axis plot}
\begin{figure*}[!h]
\centering
\includegraphics[width=.525\linewidth]{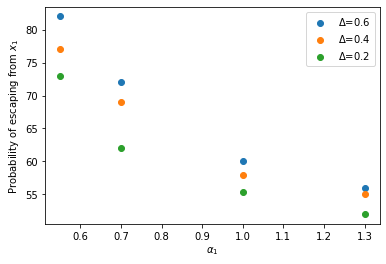}
\caption{
Four different values of $\alpha_1$ and three values of $\Delta$ are selected, and the y-axis shows the probability of escaping from $x_1$, which is the axis with lower $\alpha$. For example, the top-left most dot (blue) shows that when $\alpha_1=0.55$ and $\alpha_2=1.05$ the probability of the process to escape from axis $x_1$ is $\sim 82\%$. }\label{fig:eacape_axis}
\end{figure*}
\end{document}